%% file: example_paper.tex
\documentclass{article}

\usepackage{microtype}
\usepackage{graphicx}
\usepackage{subfigure}
\usepackage{booktabs} 

\usepackage{hyperref}

\usepackage[accepted]{icml2023}

\usepackage{amsmath}
\usepackage{amssymb}
\usepackage{mathtools}
\usepackage{amsthm}

\usepackage[capitalize,noabbrev]{cleveref}

\theoremstyle{plain}
\newtheorem{theorem}{Theorem}[section]
\newtheorem{proposition}[theorem]{Proposition}

\theoremstyle{definition}

\theoremstyle{remark}

\usepackage[textsize=tiny]{todonotes}

\usepackage{amsmath}

\usepackage{bbold}

\usepackage{colortbl}

\usepackage{graphicx} 
\usepackage{wrapfig}

\usepackage{pgffor}

\usepackage{adjustbox}

\usepackage{comment}

\newcolumntype{P}[1]{>{\centering\arraybackslash}p{#1}}
\newcolumntype{M}[1]{>{\centering\arraybackslash}m{#1}}

\usepackage{tikz}
\usetikzlibrary{math}

\newcommand\styleAuthor{1}
\usepackage{ifthen}
\ifthenelse{\equal{\styleAuthor}{\string 1}}
{} 
{\usepackage{authblk}} 

\usepackage[nomessages]{fp}

\usepackage{microtype}
\usepackage{caption}

\usepackage[T1]{fontenc}

\usepackage{xcolor}
\definecolor{tab:blue}{HTML}{1F77B4}
\definecolor{tab:red}{HTML}{D62728}
\definecolor{defcolor1}{HTML}{1F77B4}
\definecolor{defcolor2}{HTML}{FF7F0E}

\usepackage{graphicx}
\usepackage{lipsum}

\usepackage[normalem]{ulem}

\usepackage{tikz}

\newcommand{\udensdash}[1]{%
    \tikz[baseline=(todotted.base)]{
        \node[inner sep=1pt,outer sep=0pt] (todotted) {#1};
        \draw[densely dashed] (todotted.south west) -- (todotted.south east);
    }%
}%

\icmltitlerunning{Predicting Grokking Long Before it Happens}

\begin{document}

\twocolumn[


\icmltitle{Predicting Grokking Long Before it Happens: \\A look into the loss landscape of models which grok}



\icmlsetsymbol{equal}{*}

\begin{icmlauthorlist}
\icmlauthor{Pascal Jr. Tikeng Notsawo}{yyy,sch}
\icmlauthor{Hattie Zhou}{yyy,sch}
\icmlauthor{Mohammad Pezeshki}{meta}
\icmlauthor{Irina Rish}{yyy,sch}
\icmlauthor{Guillaume Dumas}{sch,comp}
\end{icmlauthorlist}

\icmlaffiliation{yyy}{Mila, Montréal, Quebec, Canada}
\icmlaffiliation{comp}{CHU Sainte-Justine Research Center, Montréal, Quebec, Canada}
\icmlaffiliation{sch}{Université de Montréal, Montréal, Quebec, Canada}
\icmlaffiliation{meta}{Meta AI Research}

\icmlcorrespondingauthor{Pascal Tikeng Notsawo}{pascal.tikeng@mila.quebec}

\icmlkeywords{Machine Learning, ICML}

\vskip 0.3in
]



\printAffiliationsAndNotice{}  

\begin{abstract}

This paper focuses on predicting the occurrence of grokking in neural networks, a phenomenon in which perfect generalization emerges long after signs of overfitting or memorization are observed. It has been reported that grokking can only be observed with certain hyper-parameters. This makes it critical to identify the parameters that lead to grokking. However, since grokking occurs after a large number of epochs, searching for the hyper-parameters that lead to it is time-consuming.  In this paper, we propose a low-cost method to predict grokking without training for a large number of epochs. In essence, by studying the learning curve of the first few epochs, we show that one can predict whether grokking will occur later on. Specifically, if certain oscillations occur in the early epochs, one can expect grokking to occur if the model is trained for a much longer period of time. We propose using the \emph{spectral signature} of a learning curve derived by applying the Fourier transform to quantify the amplitude of low-frequency components to detect the presence of such oscillations. We also present additional experiments to explain the cause of these oscillations and characterize the loss landscape.

\end{abstract}

\section{Introduction}
\label{introduction}
\vspace{-0.01in}
Despite the recent growth of theoretical studies and empirical successes of neural networks \citep{DBLP:journals/corr/abs-1303-5778,DBLP:journals/corr/HeZRS15,NIPS2012_c399862d,44806}, understanding why such networks 
find generalizable solutions
in over-parameterized regimes, where the number of learnable parameters is much larger than the number of training samples, remains an open question. Indeed, one of the major challenges of deep learning is that, in practice, neural networks are highly overparameterized \citep{AllenZhu2019LearningAG,Zhang2017UnderstandingDL}, and in some cases, it is observed that overparameterization empirically improves optimization and generalization, which seems to contradict traditional learning theory. For example, \citet{DBLP:journals/corr/LivniSS14} observed that in synthetic data generated from a target network, the learned network converges faster when it has more parameters than the target network. \citet{DBLP:journals/corr/abs-1802-05296} also found that, in practice, trained overparameterized networks can often be compressed into simpler networks with far fewer parameters, without affecting their generalizability. \citet{DBLP:conf/iclr/NakkiranKBYBS20} experiment a double-descent phenomenon where, as the model size increases, performance first gets worse and then gets better.  Recently, \citet{power2022grokking} have shown through a phenomenon they named \textit{grokking} that long after severe overfitting, 
validation accuracy sometimes suddenly begins to increase from chance level to perfect generalization, and that the amount of optimization required for this generalization increases rapidly as the size of the data set decreases. 

The grokking phenomenon opens the way to new studies concerning the structure of the minimum found by Stochastic Gradient Descent (SGD), and how networks behave in the neighbourhood of SGD training convergence. Indeed, neural activity is often characterized by an exploratory early phase of rapid learning with a rapid and sometimes abrupt decrease in the loss function \citep{DBLP:conf/iclr/NakkiranKBYBS20,doi:10.1073/pnas.2015617118}. This phase is followed by a second phase, often called the diffusion phase \citep{shwartz-ziv2017opening}, when the learning error reaches its minimum value and the global loss decreases again, but much more slowly and gradually. This second phase, often chaotic \citep{herrmann2022chaotic,cohen2021gradient,Thilak2022TheSM}, is characteristic of the structure of the minimum found by the optimization algorithm \citep{DBLP:journals/corr/GoodfellowV14,im2016empirical,smith2017exploring,DBLP:conf/iclr/KeskarMNST17,DBLP:conf/nips/Li0TSG18,jastrzębski2017three,doi:10.1073/pnas.2015617118}. 
Optimization hyper-parameters and initialization strongly affect training dynamics during the above phases, the convergence to a specific region in parameter space, and the geometries and generalization properties of solutions found by SGD \citep{DBLP:conf/iclr/KeskarMNST17,jastrzębski2017three}. For example, \citet{DBLP:conf/iclr/KeskarMNST17} empirically showed that a larger batch size correlates with sharper or isolated  minima and worse generalization performance, while small batch methods correlates with flater minima. \citet{jastrzębski2017three} find that the critical control parameter for SGD is not the batch size alone, but the ratio of the learning rate and the batch size, and that higher values for this ratio result in convergence to wider minima. 

Recent work has shown that grokking is observed only with a certain range of hyperparameters \citep{power2022grokking,Liu2022TowardsUG}. Others have studied microscopic phenomena that coincide or come in tandem with delayed generalization, such as the emergence of structure in embedding space \citep{Liu2022TowardsUG} and the slingshot effect \citep{Thilak2022TheSM}. 
But observing grokking often requires training the model for a very long time, making it difficult to construct a phase diagram of generalization covering all the hyperparameters.
In this work, we propose a low-cost method that can predict grokking long before it occurs. Our main contributions are:
\begin{itemize}
    \item We study the learning curves of a transformer network \citep{DBLP:journals/corr/VaswaniSPUJGKP17} trained on arithmetic data in settings with and without grokking (section \ref{sec:preliminaries}).
    \item We propose \textit{spectral signature} to quantify the oscillations of the loss in the early phases of training, and show that empirically it can be used to predict a potential generalization 
    (section \ref{sec:predit_grokking}).
    \item 
     To understand the origin of the oscillations and the shape of the minimizer, we analyze the model's loss landscape along the training trajectory, and present evidence linking the shape of this landscape to the different phenomena related to grokking such as the slingshot mechanism \citep{Thilak2022TheSM} (section \ref{sec:landscape}).
\end{itemize}


\section{Preliminaries}
\vspace{-0.01in}
\label{sec:preliminaries}

\subsection{Notations}
\vspace{-0.01in}
The index $t$ is used to characterize the training steps ($\theta_t$, etc.).  $L_t = L(\theta_t)$ denotes the loss at $\theta_t$ (the parameter update at time $t$ given the optimization algorithm of choice), $G_t = \nabla L(\theta_t)$ the gradient of the loss function at  $\theta_t$  and $\mathcal{H}_t = \nabla^2 L(\theta_t)$ the local Hessian matrix of the loss at $\theta_t$ ($\theta(t)$, $L(t)$, $G(t)$ and $\mathcal{H}(t)$ in the continuous case). We define $\dot{x} =  \partial x (t) /  \partial t$ for a time function $x = x(t)$.

\subsection{
Task definition and training scheme
}

Let $\circ$ be a binary mathematical operator (such as addition or multiplication), and let $p$ and $q$ be two strictly positive integers. The task is to predict $(a \circ b) \ mod \ q$ for any pair of numbers $(a, b) \in [p]^2$, with $[p] = \{0, \dots, p-1\}$. The dataset $\mathcal{D}$ that we can thus constitute has a size of $p(p+1)/2$ if $\circ$ is symmetric (and we consider $a \circ b$ and $b \circ a$ as the same operation), and $p^2$ otherwise. $\mathcal{D}$ is randomly partitioned into two disjoint and non-empty sets $\mathcal{D}_{train}$ and $\mathcal{D}_{val}$, the training and the validation dataset respectively. The training data fraction $r = |\mathcal{D}_{train}| / |\mathcal{D}|$ is a  hyperparameter.

This problem can be solved with an auto-regressive approach. For $a, b \in [p]$, let $s = s_1 \dots s_5 = \langle a \rangle \langle \circ \rangle \langle b \rangle \langle = \rangle \langle (a \circ b) \ mod \ q \rangle$ where $s_i = \langle x \rangle$ 
stands for the token corresponding to the element $x$. 
The training is performed by maximizing the likelihood under the direct autoregressive factorization, and the loss (as well as the accuracy) is calculated only on the answer part $s_5$ of the equation :
$$
\max_{\theta} \log \ p_{\theta} (s_5 | s_{<5}) 
= 
\log \mbox{ } \frac {\exp(h_{\theta}(s_{<5})^T e(s_5))} 
{\sum_{x \in [q]} \exp(h_{\theta}(s_{<5})^T e(\langle x \rangle ))}
$$

where $h_{\theta}(s_{<5}) \in \mathbb{R}^{m}$ is a context representation produced by neural models parameterized by $\theta$ for $s_{<5} = s_1 \dots s_4$; 
and $e(s_i)$ the trainable embedding  vector of $s_i$ (the embedding weights are used as weights for the output linear classification layer).


We focus on the following binary operations, for a prime number $p = 97$, and $q = p$ : modular addition ($+$), $a \circ b = a + b \ (mod \ p)$ for $a, b \in [p]$; 
and  
multiplication in the permutation group $S_5$, $a \circ b = a \cdot b$ for $a, b \in S_5$.

For all experiments, we used a transformer with 2 layers, 128 embeddings, and 4 attention heads, with a total of roughly $4 \cdot 10^5$ non-embedding parameters. We use the following hyperparameters unless stated otherwise: AdamW optimizer with a learning rate of $10^{-4}$, weight decay of $1$ and ($\beta_1, \beta_2) = (0.9, 0.98)$, linear learning rate warm up over the first $10$ updates, and minibatch size of $|\mathcal{D}_{train}|$.

\subsection{
Grokking and Non-grokking
}
\label{subsec:grokking}

\begin{figure}[htp]
\centering
\includegraphics[width=1.\linewidth]{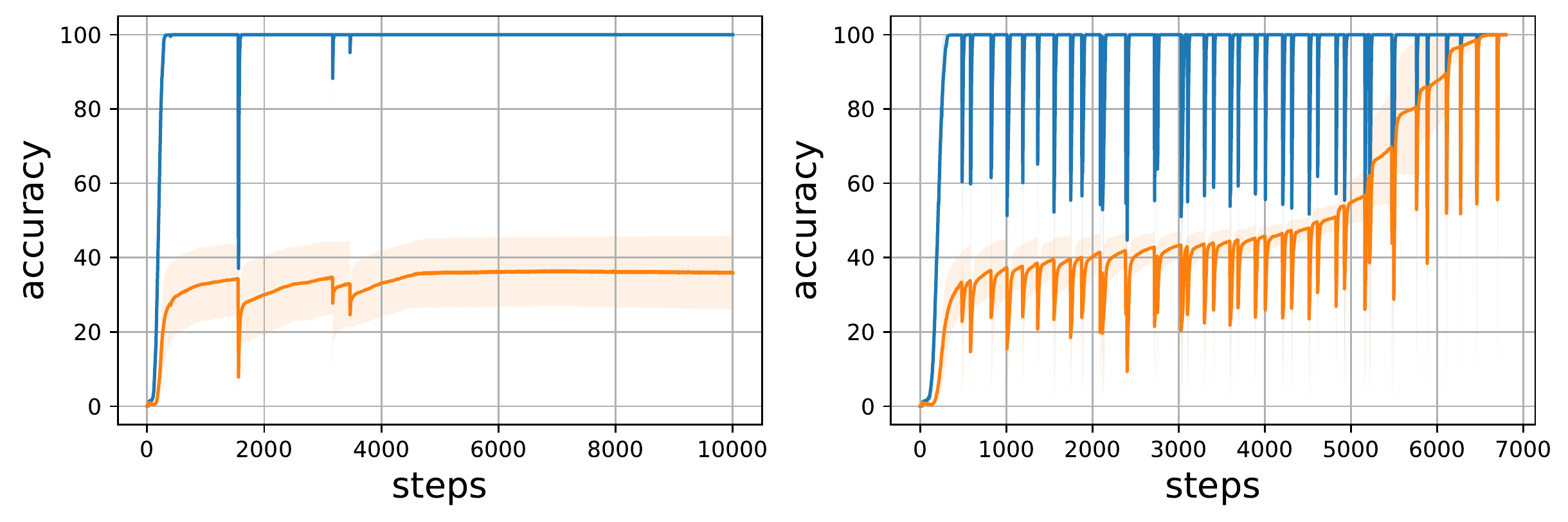}
\hfill
\caption{ 
Oscillation in \textcolor{defcolor1}{training} and \textcolor{defcolor2}{validation} accuracies. We train on modular addition with $r=0.5$.  
The left curve shows a case where the model did not grok after 10k steps of training. The right curve shows the generalization after overfitting. Training accuracy becomes close to perfect at $t_2 < 300$ optimization steps, but it takes close to $t_4 \approx 7k$ steps for validation accuracy to reach that level.}
\label{fig:phases1}
\end{figure}

The phenomenon of grokking is characterized by three phases in order: 1) the initial learning phase where both training and validation performance are poor, 2) the second phase where training performance is near perfect while validation performance is low, 3) and the generalization phase where validation performance improves to match training performance. The grokking phenomenon is marked by a long memorization phase (figure ~\ref{fig:phases1}). More specifically, let $t_1$ be the transition step from phase 1 to phase 2 (when the training accuracy reaches a value strictly greater than 0\% for the first time), $t_2$ the step at which the training data is completely memorized (when the training accuracy reaches 100\% for the first time), $t_3$ the transition step from memorization to comprehension (when the validation accuracy reaches a value strictly greater than 0\% for the first time), and $t_4$ the step at which the model generalizes (when the validation accuracy reaches 100\% for the first time). It is the delay between memorization and generalization $t_4 - t_2$
that allow, in general, to name the learning process grokking or simple generalization. But in our case, for the sake of simplicity, we will characterize all situations with generalization as grokking. 

Although it is easy to identify grokking, it is very difficult to give a formal definition of its opposite, since in practice we cannot train a model for an infinite number of steps. We proceeded empirically by training the models for a large range of hyperparameters, and we fit a function that predicts $t_4$, the generalization step, for each training data fraction $r$. That is, if for other hyper-parameters, and for a given training data fraction $r$, we train a model for $t_4(r)+\epsilon$ (we used $\epsilon = 1k$) steps and have no generalization, we can stop the training. In general, more data leads to faster grokking (i.e. $t_2(r)$ and $t_4(r)$ are decreasing functions of $r$). Empirically, $t_4$ follows a power law of the form $t_4(r) = a r ^{-\gamma} + b$. This was first predicted by \citet{phase2022grokking}.  We give more details in the appendix (\ref{subsec:grokking_appendix}).

\section{Predicting grokking}
\label{sec:predit_grokking}

A starting observation is that the learning curves of models that grok exhibit oscillatory behaviors 
(Figure~\ref{fig:phases1}).
A related phenomenon, the slingshot effect,  was observed and named recently by \citet{Thilak2022TheSM}. Cyclic transitions between a stable and unstable training regime characterize this phenomenon. 
\citet{Thilak2022TheSM} characterized this phenomenon as the complete cycle starting with the norm growth phase and ending with the norm plateau phase, and found that it is ubiquitous and can be easily replicated in multiple scenarios, encompassing a variety of models and data sets. They observe that slingshots and grokking tend to come in tandem, that is grokking almost exclusively happens at the onset of slingshots, and is absent without it. 
As pointed by \citet{Thilak2022TheSM}, this type of transition is reminiscent of the \textit{catapult phenomenon} \citep{lewkowycz2020large}, where the loss initially increases and begins to decrease once the model catapults to a region of lower curvature early in the training, for a sufficiently large training step. 

Based on this observation, we conjecture that \textit{the spectral signature of training loss in early epochs can hint us about the existence of an upcoming grokking}. We first try to comparatively quantify the oscillations in the training loss when the model groks and when it does not grok, the ideal being to stop the training if the model does not seem to be able to grok, in order to save computational resources.  
From the gradient flow equation $\dot{\theta} = - G(t)$, it holds that $\dot{L} \approx - \| G(t) \|^2$ 
and 
$\ddot{L} \approx 2 G(t)^T H(t) G(t) = 2 \sum_{i} \lambda_i(t) \langle G(t), v_i(t) \rangle^2
$ 
with 
$\{ \lambda_i(t) \}_i$ the spectrum of $H(t)$, and $\{ v_i(t) \}_i$ the associated eigenvectors. From this, it becomes clear that the evolution of $L$ over time depends on the norm of the gradient, and how fast it changes depends on the curvature of its landscape. Any signal that can be represented as a variable that varies in time has a corresponding frequency spectrum. 
We considered the training loss $L$ over the training steps (and in the early stages of training) as signals and analyzed its spectral signature. By spectral signature of the loss, we mean any measure or set of measures that can quantify the oscillations in the loss, such as the spectral energy or the Hjorth parameters \citep{HJORTH1970306} - activity, mobility and complexity. The Hjorth activity, which is the variance of the signal in the time domain, is equal to the spectral energy if the signal $L$ has zero mean. The latter condition is obtained 
by passing $L$ through a sufficiently low-pass filter in the frequency domain, which removes its non-oscillatory components 
\footnote{A non-oscillatory component of a signal is any component that does not vary rapidly with time, such as a constant value, a linear trend, or a smooth curve. These components are generally considered to be low-frequency signals, and can be removed by passing the signal through a low-pass filter. 
}
that are not necessary for the quantification of the oscillations. In this case, the Hjorth parameters are directly related to the loss landscape as illustrated below.

\begin{figure}[tbh]
\includegraphics[width=1.0\linewidth]{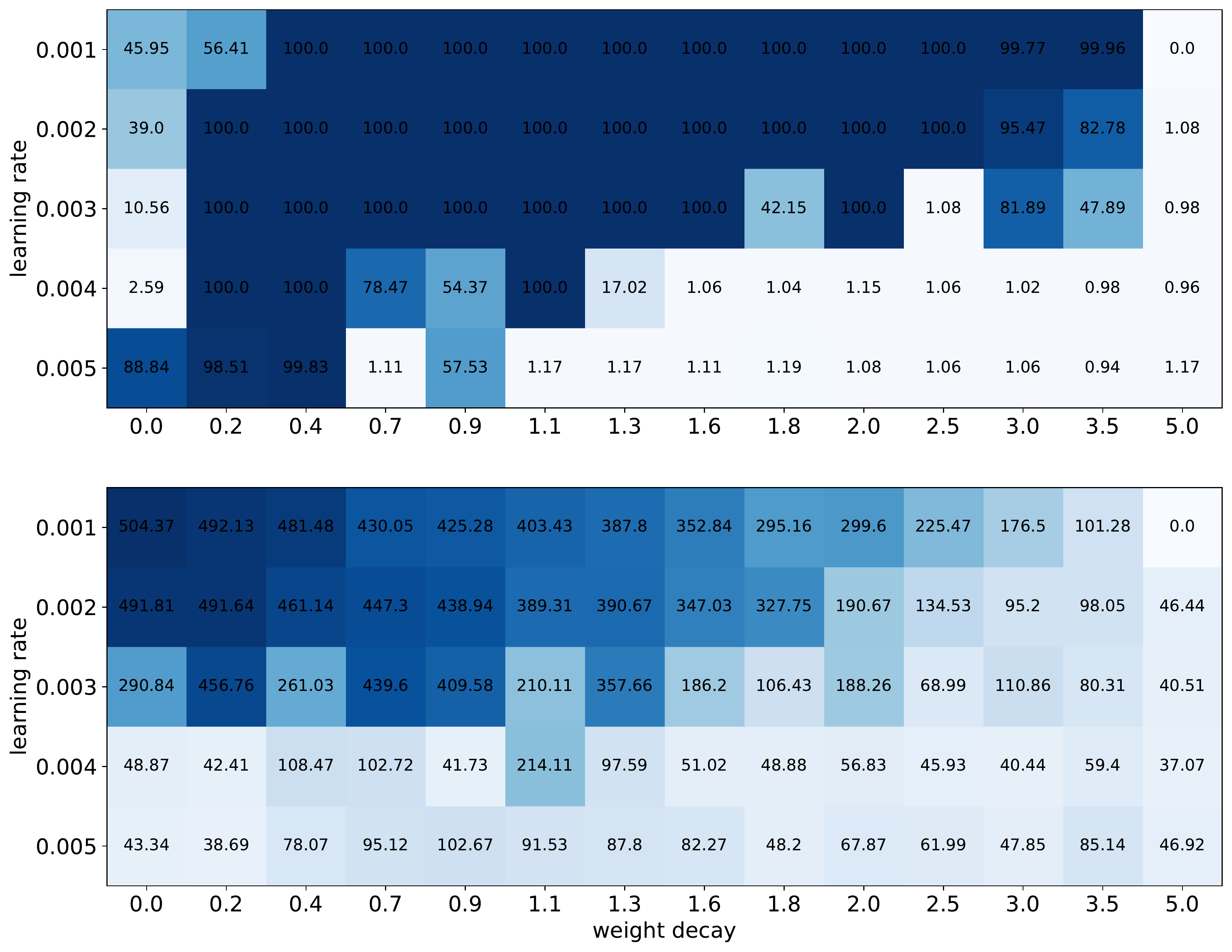}
\hfill
\vspace{-0.1in}
\caption{The first figure (top) represents the validation accuracy (\%) at the end of the training ($10k$ steps), and the second figure (bottom) represents the spectral energy (activity) in the training loss for the first 400 training steps ($r = 0.5$). On the x-axis we have the weight decay strength, and on the y-axis we have the learning rate. A similarity is observed between the oscillation patterns in the training loss during the initial stages of training and the validation accuracy. This suggests that the spectral signature can be used as an indicator or proxy for the upcoming grokking phenomenon. 
The highest degree of generalization is typically observed when using small learning rates and small weight decay. While large learning rates may increase oscillations, this does not directly lead to grokking and is not necessarily evident in the early stages of training. Instead, such effects become more noticeable near the basin of attraction of the minimum.
}
\label{fig:spect_diagram}
\end{figure}

Let $\mathcal{F}(L)$ denote the Fourier transform of $L(t)$ and $m_n(L) = \int \omega^n  |\mathcal{F}(L)(\omega)|^2 d \omega$, the $n^{th}$ moment of $\mathcal{F}^2 (L)$, 
with $|\mathcal{F}(L)(\omega)|^2$ the energy spectral density present in the 
pulse $\omega$.  
The Hjorth activity represents the signal power, the surface of the power spectrum in the frequency domain. It is given by $m_0(L)$, which is equal to $\int L^2 (t) dt$ by the parseval's theorem \footnote{
The Hjorth mobility is the mean frequency or the proportion of standard deviation of the power spectrum, and is given by $\sqrt{m_2 (L) / m_0(L)}$ with 
$m_2(L) = \int  | \omega \mathcal{F}(L)(\omega)|^2 d \omega
= \int \dot{L}^2 (t) dt \approx  m_0 (\dot{L})$ the activity of the gradient norm $\| G(t) \|^2$. In a similar way,
The Hjorth complexity, which indicates how the shape of a signal is similar to a pure sine wave, is given by $\sqrt{m_4 (L) / m_2(L)}$ with 
$m_4(L) = \int  | \omega^2 \mathcal{F}(L)(\omega)|^2 d \omega = \int \ddot{L}^2 (t) dt \approx m_0 (\ddot{L})$ the activity of the hessian spectrum. 
}. Figure \ref{fig:spect_diagram} shows a similarity between the oscillation in the training loss in the early phases of training and the validation accuracy for $r=0.5$ (more results in the appendix, \ref{subsec:grokking_appendix}), suggesting that the spectral signature can serve as a proxy to upcoming grokking phenomenon. The generalization is most observed for small learning rate and small weight decay. Although large learning rates have the effect of increasing the oscillations, this does not directly result in grokking and is not necessarily visible in the early steps, but more near the basin of attraction of the minimum. Importantly, the spectral signature of the loss is not an explicit capacity measure so either positive or negative correlation with generalization could potentially be informative. Our observations is related to the empirical findings of \citet{jiang2019fantastic}. They investigate more than 40 complexity measures derived from both theoretical and empirical studies and train a variety of models by systematically varying commonly used hyperparameters. Their results suggest that the difficulty of optimization during the initial phase of the optimization benefits the final generalization, but the evolution of the loss when it reaches a certain value is not correlated to the generalization of the final solution. 

\section{Grokking loss landscape}
\label{sec:landscape}



What can be a possible explanation of the oscillations observed in the early phases and the slingshot phenomenon? Many hypotheses can in fact be put forward to reconcile oscillations with delayed generalization. \textit{Does the model, during the confusion, memorization and comprehension phases, oscillate around a local minimum, cross a very flat region, or circumvent a large obstacle?} 
One initial intuitive explanation for delayed generalization could be that the model gets stuck in local solutions during the memorization phase. The ease of escaping these local solutions depends on factors such as initialization or hyperparameters like the amount of training data, and the model achieves grokking when it successfully breaks free from the basin of attraction of such solutions.
This explanation is similar to the working hypothesis of \citet{DBLP:conf/uai/DziugaiteR17} that SGD finds good solutions only if they are surrounded by a relatively large volume of solutions that are nearly as good. 
A second attempt to explain grokking landscape is that the model crosses an ill-conditioned surface, potentially a valley with almost no curvature in the majority of directions and very high curvature in some directions.
This results in a weak progression in the directions of low curvature, and a lot of back and forth in the directions of high curvature. 
A third attempt is to consider the first two hypotheses together. That is, the model goes through several ill-conditioned local minima during training. 
We present in the following the loss landscape of grokking, a discussion on how they are linked to our hypothesis.

Since neural loss functions live in a very high-dimensional space, visualizations are only possible using low-dimensional 1D (line) or 2D (surface) plots. In this work, we consider the approach of \citet{DBLP:conf/nips/Li0TSG18}. Let $\theta$ be the point near which we want to observe the loss landscape.  We plot the loss and the accuracy as a function of $A \subseteq \mathbb{R}$, $f(\alpha) = L(\theta + \alpha \vec{\delta})$, where $\vec{\delta}$ is a direction vector carefully chosen in $\vec{\Theta}$ \footnote{
In 2D, the loss is plot as a function of $A \times B \subseteq \mathbb{R}^2$, $f(\alpha, \beta) = L(\theta + \alpha \vec{\delta} + \beta \vec{\eta})$, where $\vec{\delta}$ and $\vec{\eta}$ are two carefully chosen direction vectors in $\vec{\Theta}$. 
$\vec{\delta}$ and $\vec{\eta}$ can be randomly chosen or defined by $\vec{\delta} = \theta^{'} - \theta$ and $\vec{\eta} = \theta^{''} - \theta$, with $\theta^{'}$ and $\theta^{''}$ another points whose choice will be specified.
}. Due to scale invariance in network weights, this approach may fail to capture the intrinsic geometry of loss surfaces. To remove this scaling effect, we plot the loss functions using an adaptation of the filter-wise normalized directions \citep{DBLP:conf/nips/Li0TSG18}, that is 
$
w_k(\vec{\delta}) \leftarrow \frac{w_k(\vec{\delta})}{\| w_k(\vec{\delta}) \|} \| w_k(\theta) \|
\text{ and } 
b_k(\vec{\delta}) \leftarrow \frac{b_k(\vec{\delta})}{| b_k(\vec{\delta}) |} | b_k(\theta) |
$ for each weight $W = [\dots, w_k, \dots]^T$ ($w_k$ is a vector) and bias $b = [\dots, b_k, \dots]$ ($b_k$ is a scalar) in each layer of $\theta$ and $\vec{\delta}$ \footnote{
The same as for $\vec{\eta}$ when applied
}. This loss-landscape visualization approach, although simple, has the advantage of allowing to visualize the potential local convexity of the loss 
in the chosen direction \footnote{
A function $f: \mathbb{R}^n \to \mathbb{R}$ is convex if and only if $
g_{(x, y)} \colon 
\alpha \in [0, 1] \mapsto 
g_{(x, y)}(\alpha) = 
f\big(x + \alpha (y-x)\big)$ is convex for all $x, y \in \mathbb{R}^n$.
}.
For $r = 0.3$, figure \ref{fig:cover} shows the 1D projection of the grokking loss surface for a single epoch of training (just after the grokking step), while figures \ref{fig:30_t_T} ($\vec{\delta}_t \propto \theta^* - \theta_t$), \ref{fig:30_t_t+1} ($\vec{\delta}_t \propto \theta_{t+1} - \theta_t$), \ref{fig:t0_30} ($\vec{\delta}_t \propto \theta_0 - \theta_t$) and \ref{fig:rand_30} (random $\vec{\delta}_t$) show it for different training epochs (more experiments in the Appendix \ref{sec:landscape_appendix}). We can see that the 1-D subspace from initial to final parameters and from one minimizer to another contains many difficult and exotic structures. 

\def\cmap{plasma}



\def\sizefig{.49}

\begin{figure}[htp]
\hfill
\subfigure[multiplication ($S_5$)]{\includegraphics[width=\sizefig\linewidth]{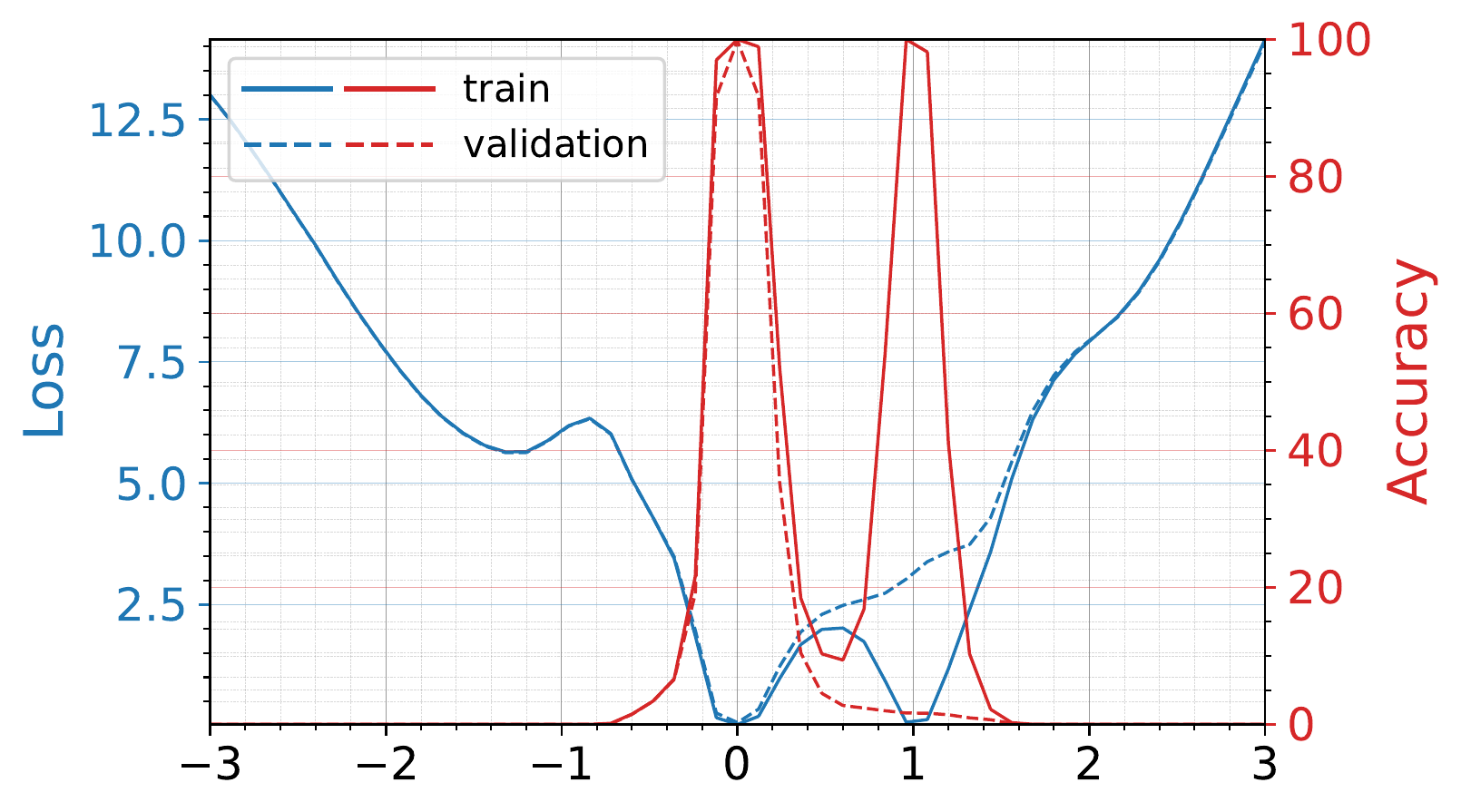}}
\hfill
\subfigure[addition ($+$)]{\includegraphics[width=\sizefig\linewidth]{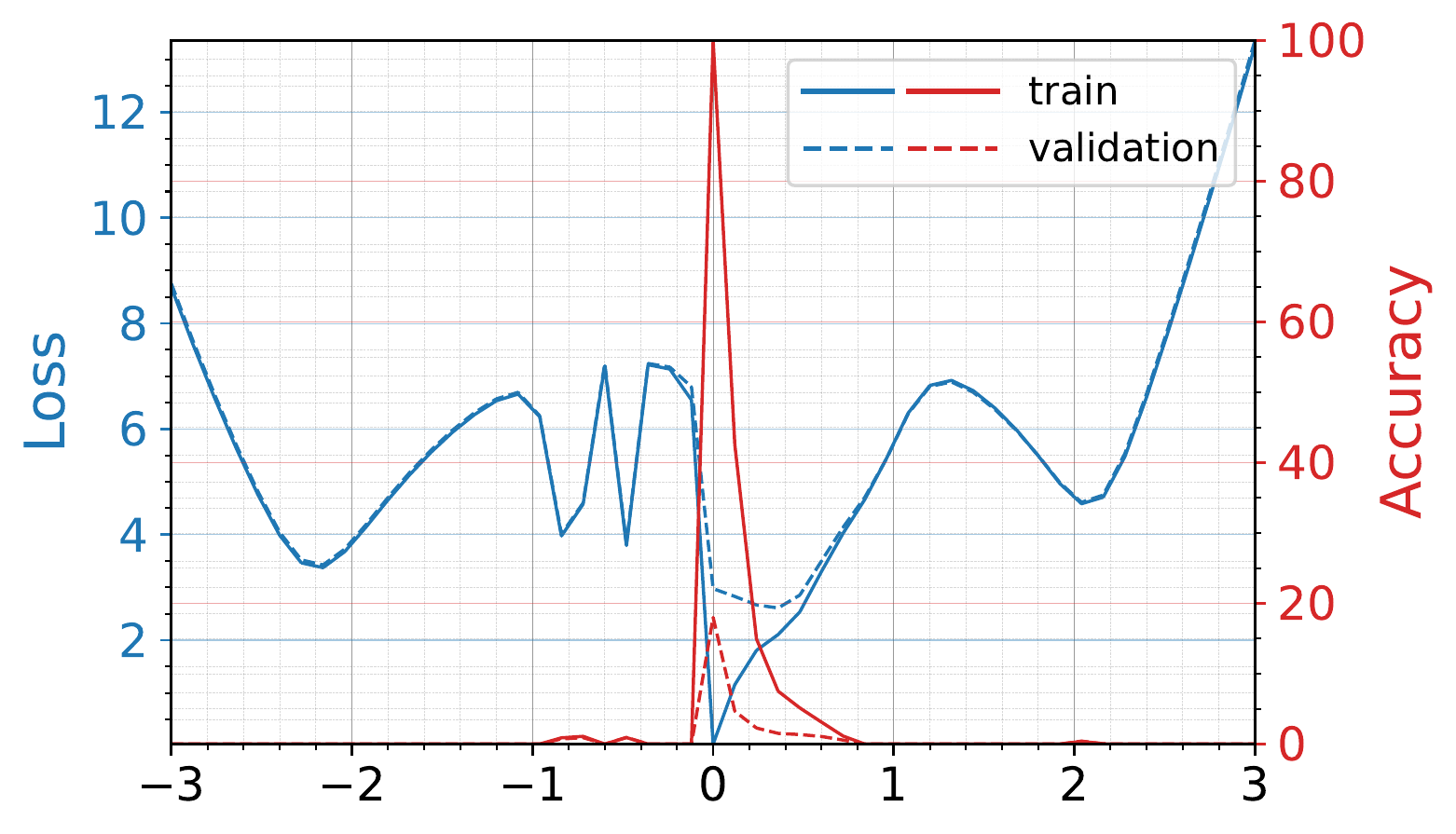}}
\hfill
\caption{1D projection of the grokking loss and accuracy surface ($r=0.3$).
The x-axis is $\alpha \in [-3, 3]$, and the y-axis are the \textcolor{tab:blue}{loss} (left axis) and \textcolor{tab:red}{accuracy} (right axis) at $\theta^* + \alpha \delta$ with $\delta \propto \theta_0 - \theta^*$ (with an adaptation of the  filter-wise normalization \citep{DBLP:conf/nips/Li0TSG18}), where $\theta_0$ is the initial parameter and $\theta^*$ the parameter just after the model has grokked (a) and before grokking (b).
We can see that the 1-D subspace from initialization to grokking contains many difficult and exotic structures. In (a), we have two minimizers of the training loss (\textcolor{tab:blue}{\underline{solid line}}), but only one of them also minimizes the validation loss (\textcolor{tab:blue}{\udensdash{dashed line}}). 
}
\vspace{-0.01in}
\label{fig:cover}
\end{figure}

\begin{figure}[htp]
\hfill
\subfigure[Loss surface : $f_t(\alpha) = Loss(\theta_t + \alpha \vec{\delta}_t)$ for each epoch $t$]{
\includegraphics[width=\sizefig\linewidth]{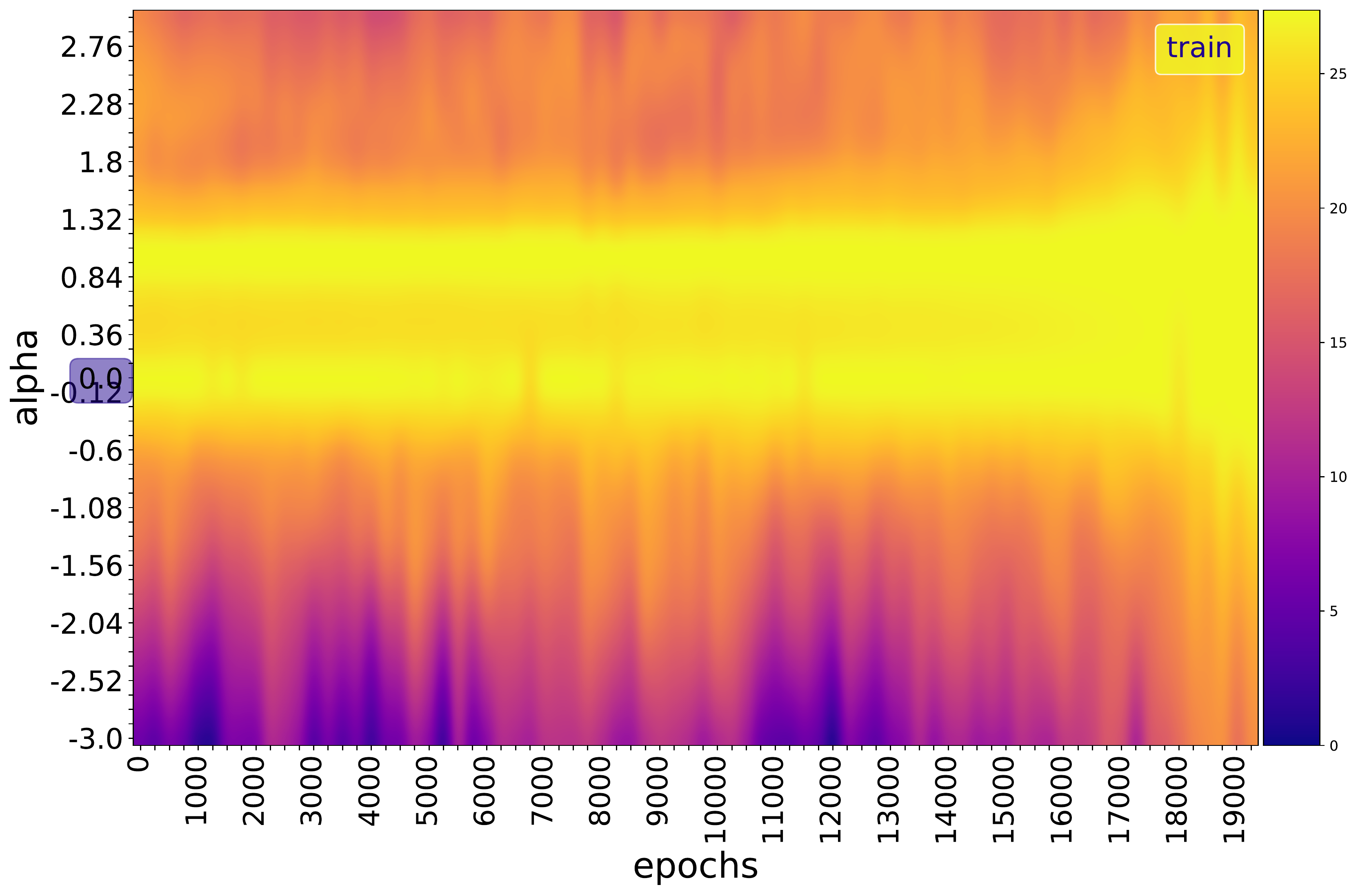}
\includegraphics[width=\sizefig\linewidth]{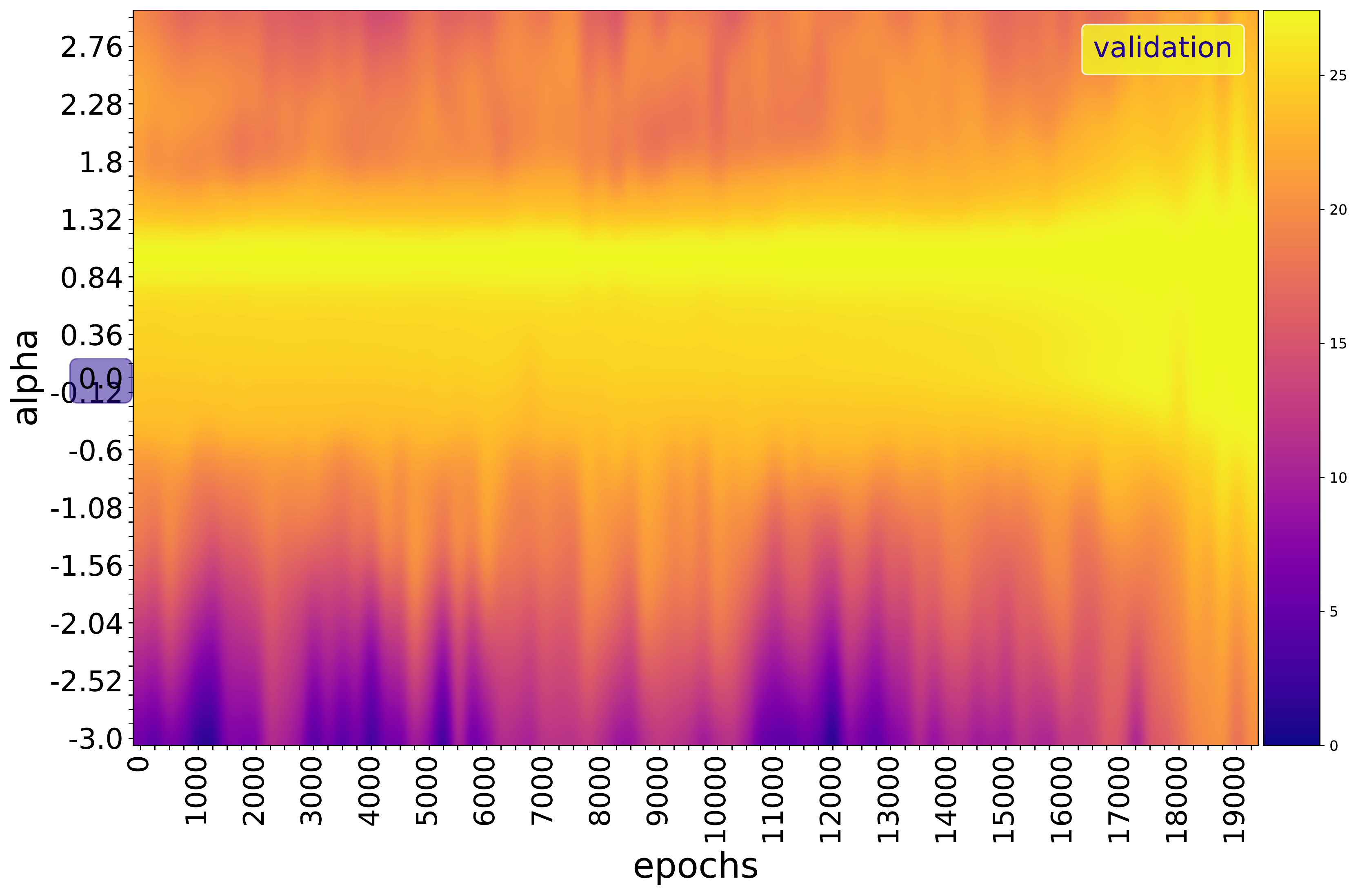}
}
\hfill
\subfigure[Accuracy surface : $f_t(\alpha) = Acc(\theta_t + \alpha \vec{\delta}_t)$ for each epoch $t$]{
\includegraphics[width=\sizefig\linewidth]{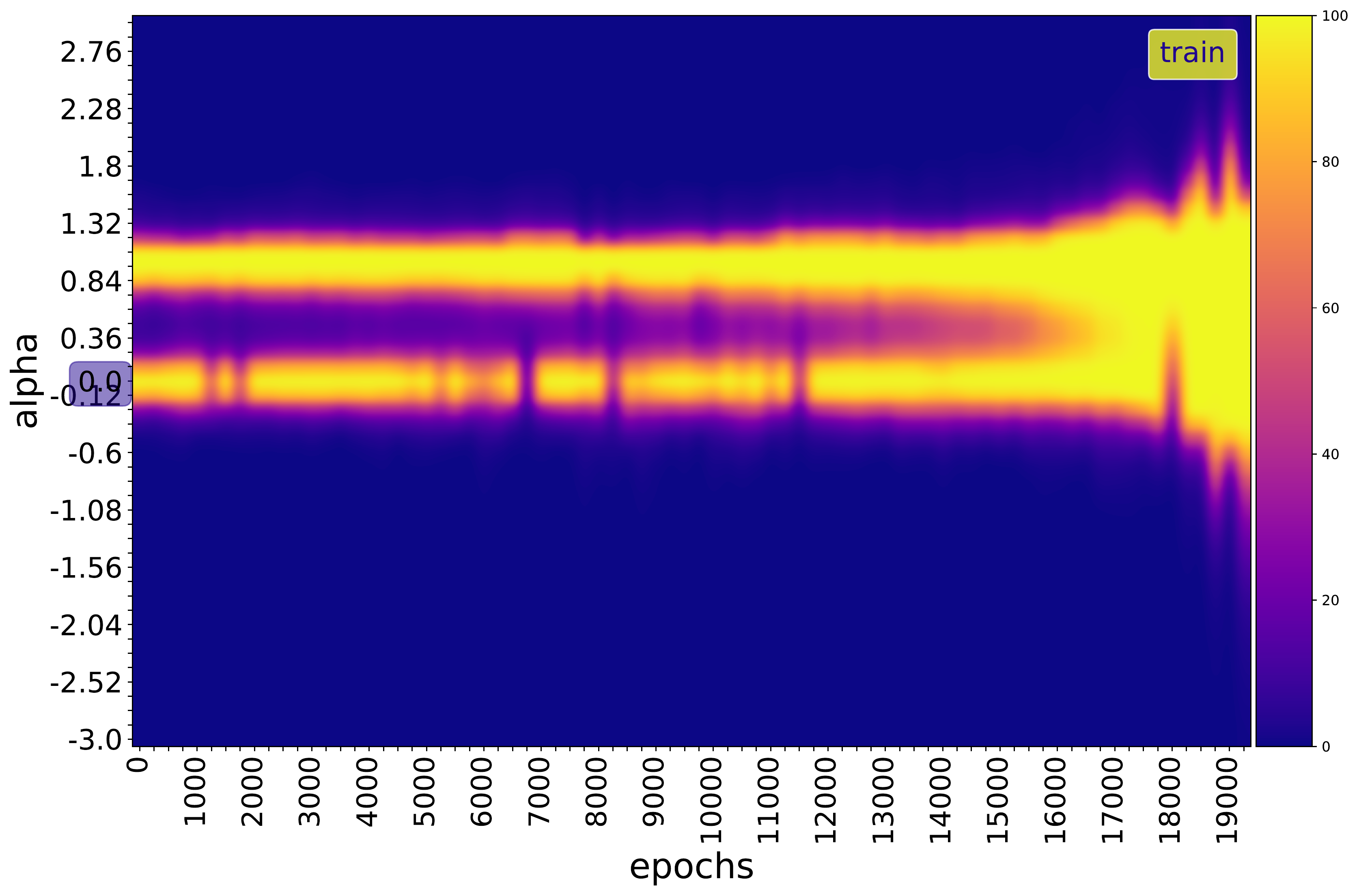}
\includegraphics[width=\sizefig\linewidth]{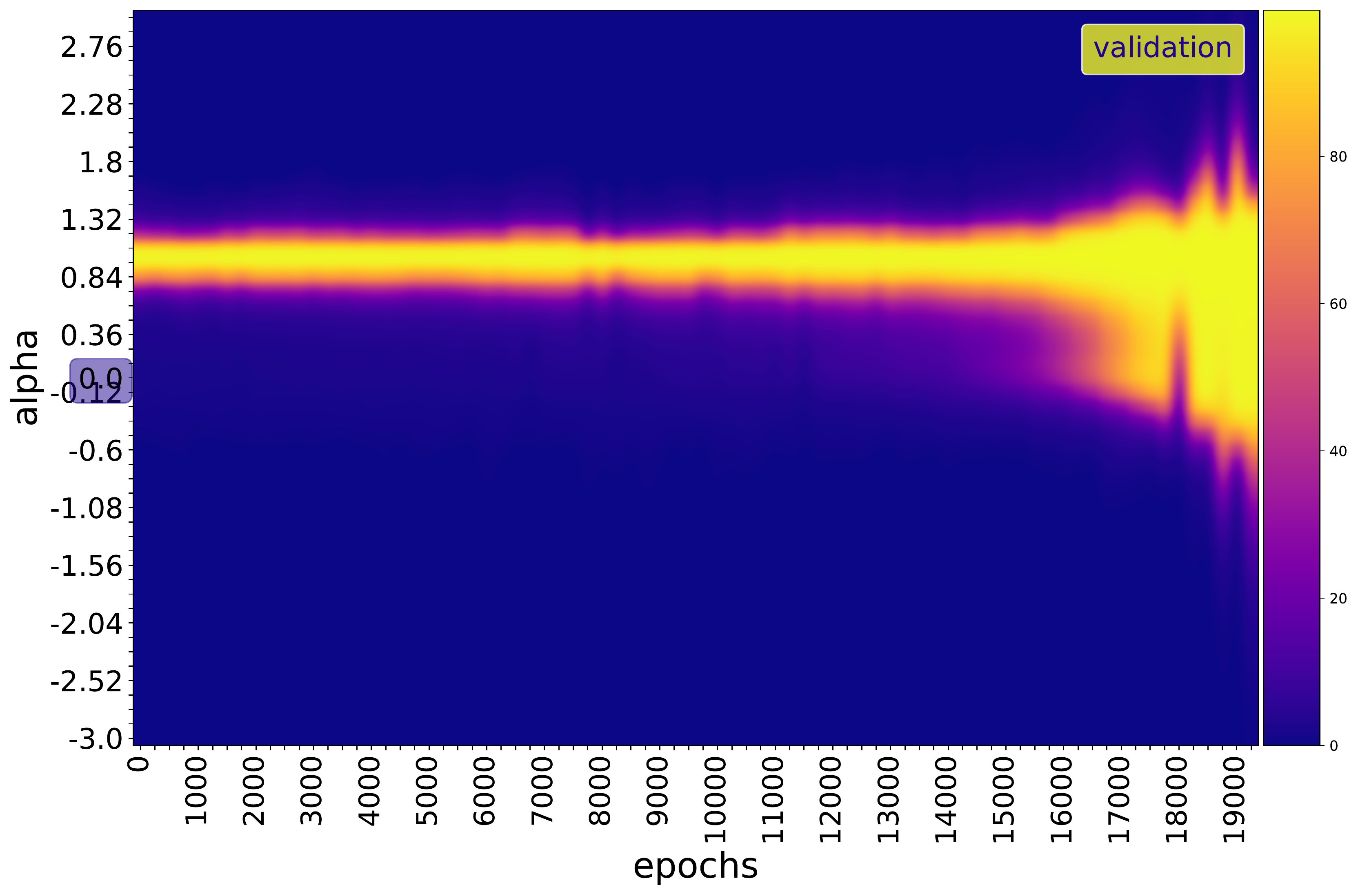}
}
\hfill
\caption{1D projection of the grokking loss and accuracy surface. This corresponds to figure \ref{fig:cover}.a, but for several training epochs ($r=0.3$). The direction $\vec{\delta}_t$ used for each training epoch $t$ is the unit vector of  $\theta^* - \theta_t$ (filter-wise normalization \citep{DBLP:conf/nips/Li0TSG18}), the direction from the parameter at epoch $t$ to the minimum. 
Here, the structure is more exotic. We can clearly see two minimizers of the training loss, but only one minimizes the validation loss: during memorization, the model is in this local minimum, and it achieves
grokking when it successfully breaks free from this local solution.
}
\vspace{-0.01in}
\label{fig:30_t_T}
\end{figure}

\begin{figure}[htp]
\hfill
\subfigure[Loss surface : $f_t(\alpha) = Loss(\theta_t + \alpha \vec{\delta}_t)$ for each epoch $t$]{
\includegraphics[width=\sizefig\linewidth]{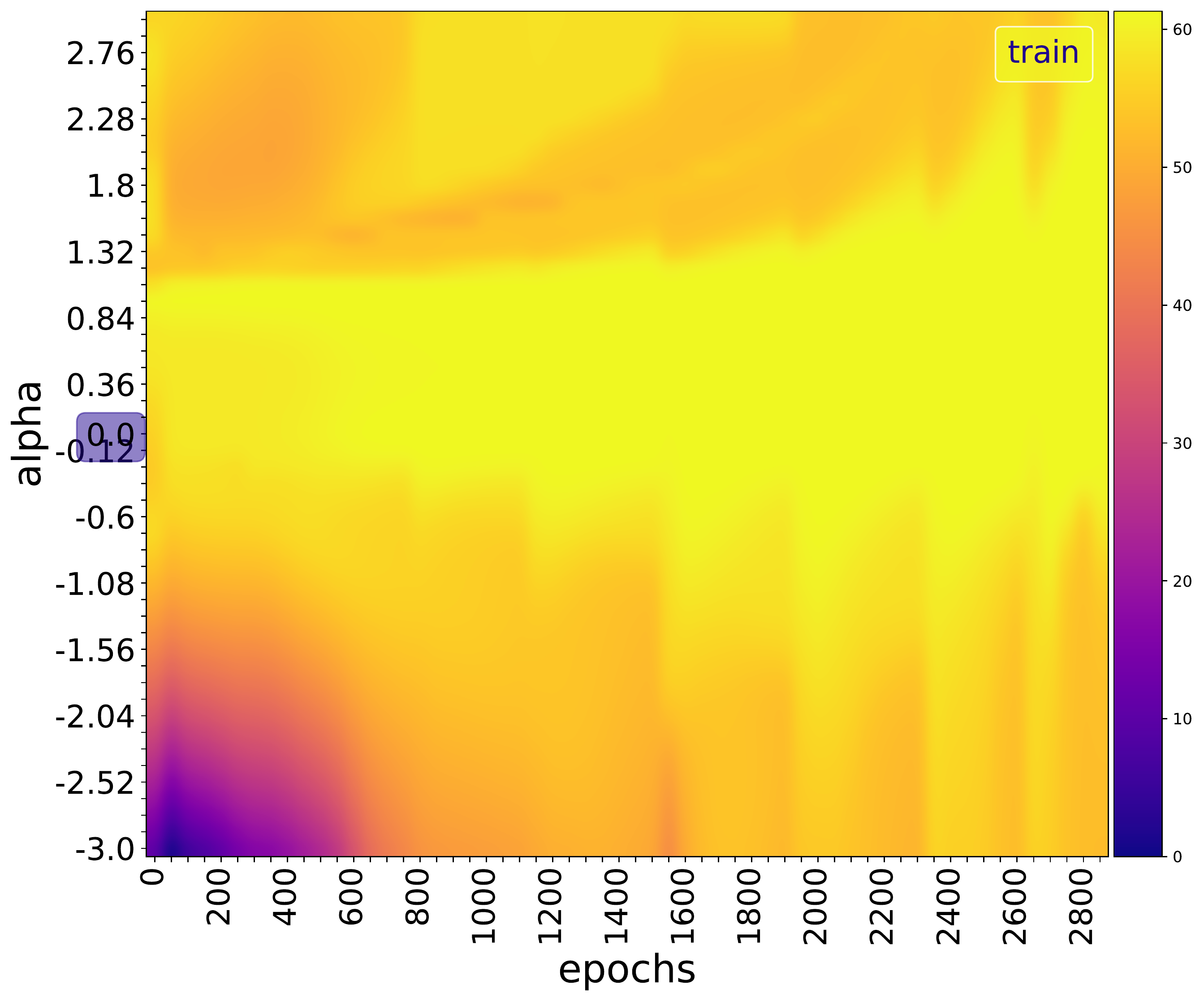}
\includegraphics[width=\sizefig\linewidth]{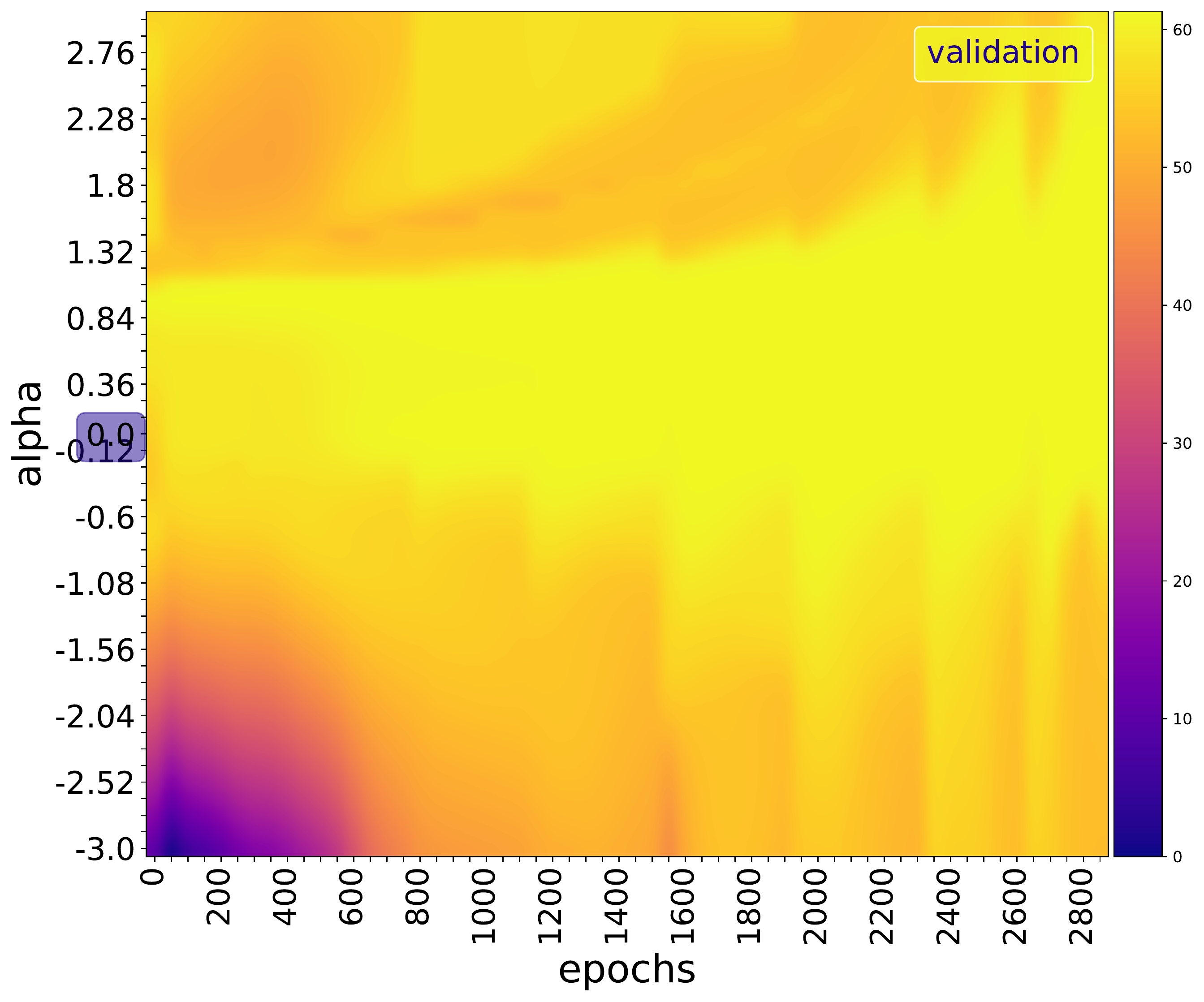}
}
\hfill
\subfigure[Accuracy surface : $f_t(\alpha) = Acc(\theta_t + \alpha \vec{\delta}_t)$ for each epoch $t$]{
\includegraphics[width=\sizefig\linewidth]{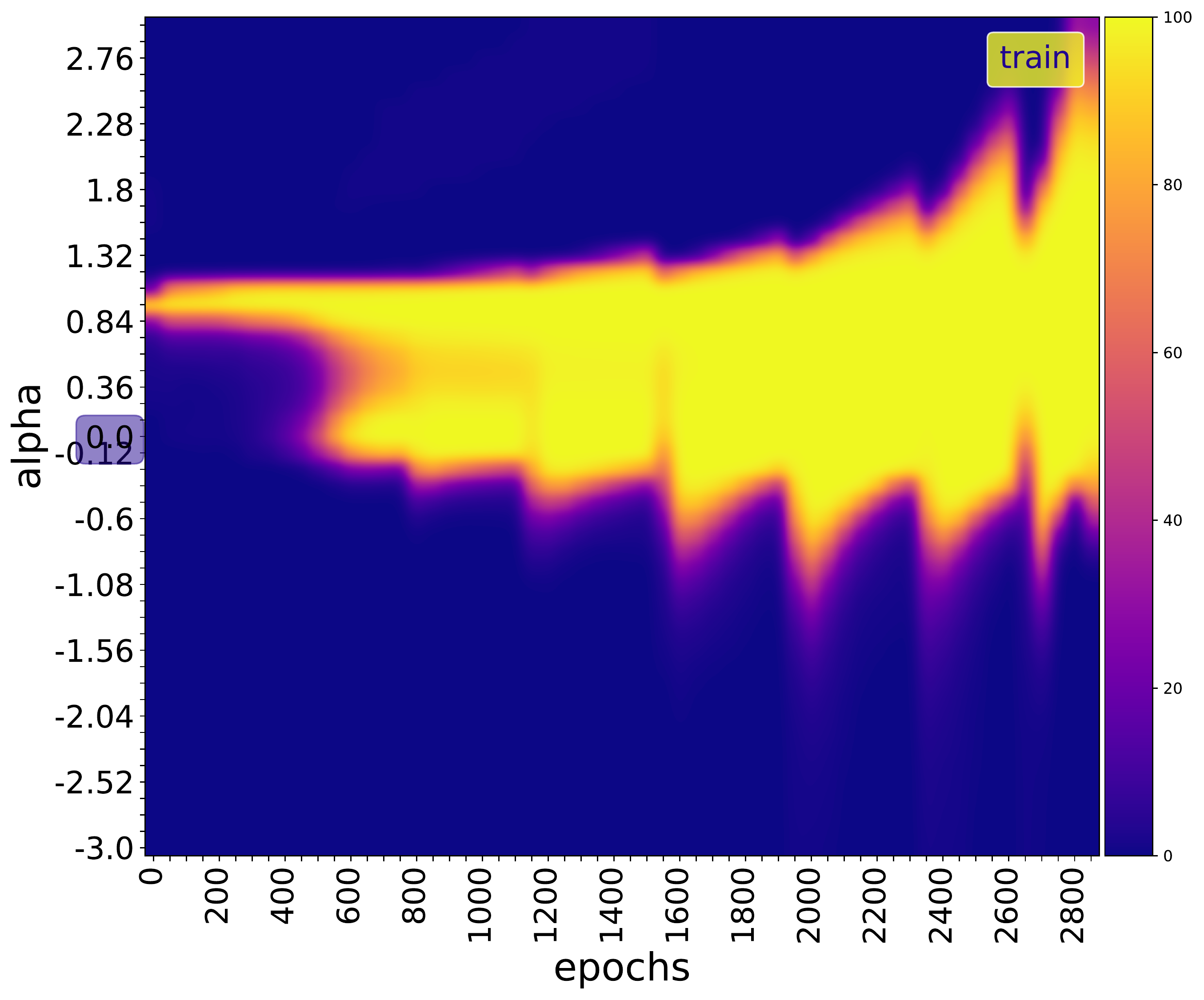}
\includegraphics[width=\sizefig\linewidth]{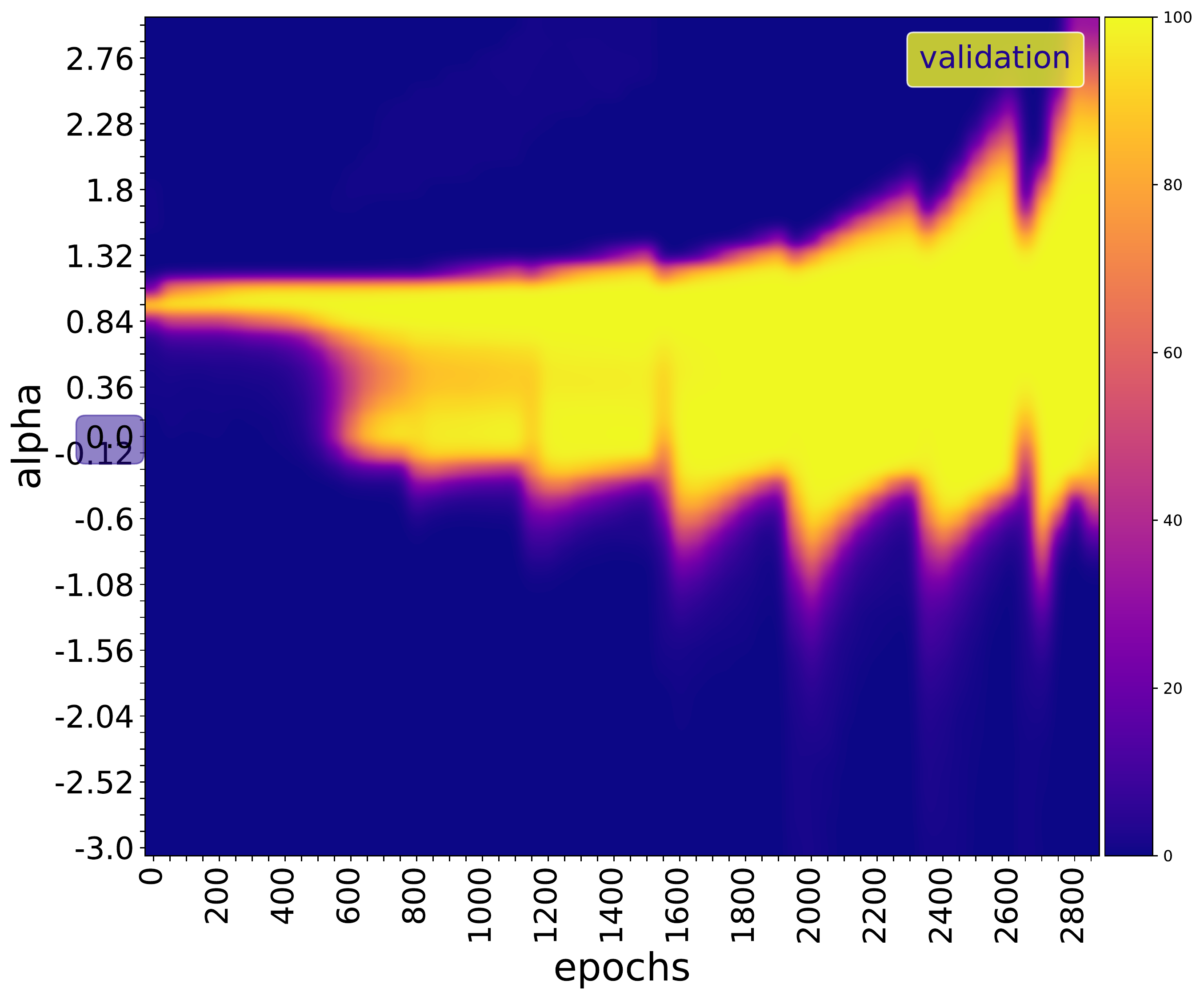}
}
\hfill
\caption{Similar to figure \ref{fig:30_t_T}, 
but with $r=0.8$. What changes between the two figures is the time it takes for the model to reach the global minimum. This time decreases as $\Theta \left( 1/ r ^{\gamma} \right)$ with $r$, $\gamma > 0$ 
(section \ref{subsec:grokking_appendix}).}
\vspace{-0.01in}
\label{fig:80_t_T}
\end{figure}

\def\sizefig{.49}

\begin{figure}[h]
\hfill
\subfigure[Loss surface : $f_t(\alpha) = Loss(\theta_t + \alpha \vec{\delta}_t)$ for each epoch $t$]{
\includegraphics[width=\sizefig\linewidth]{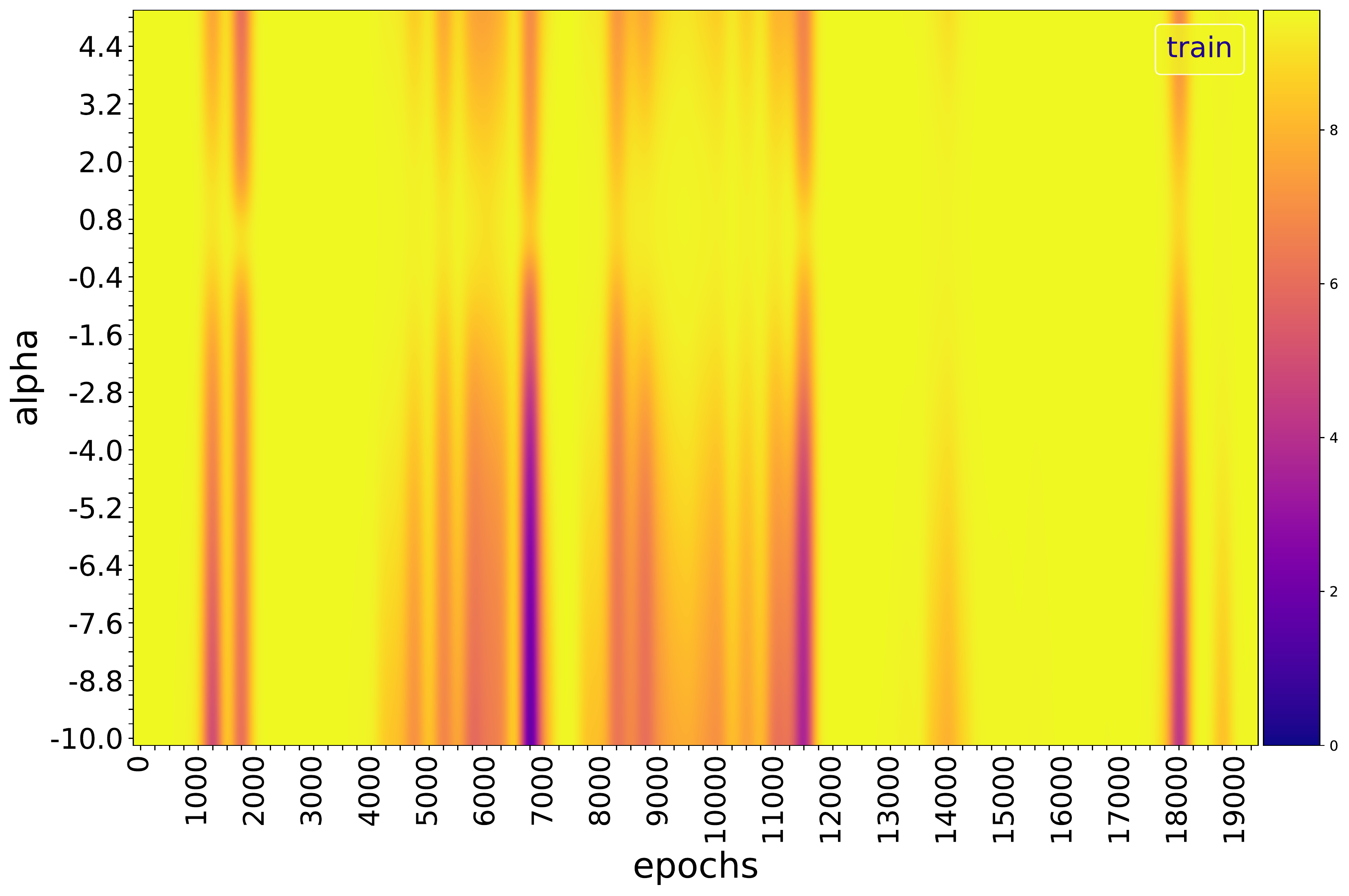}
\includegraphics[width=\sizefig\linewidth]{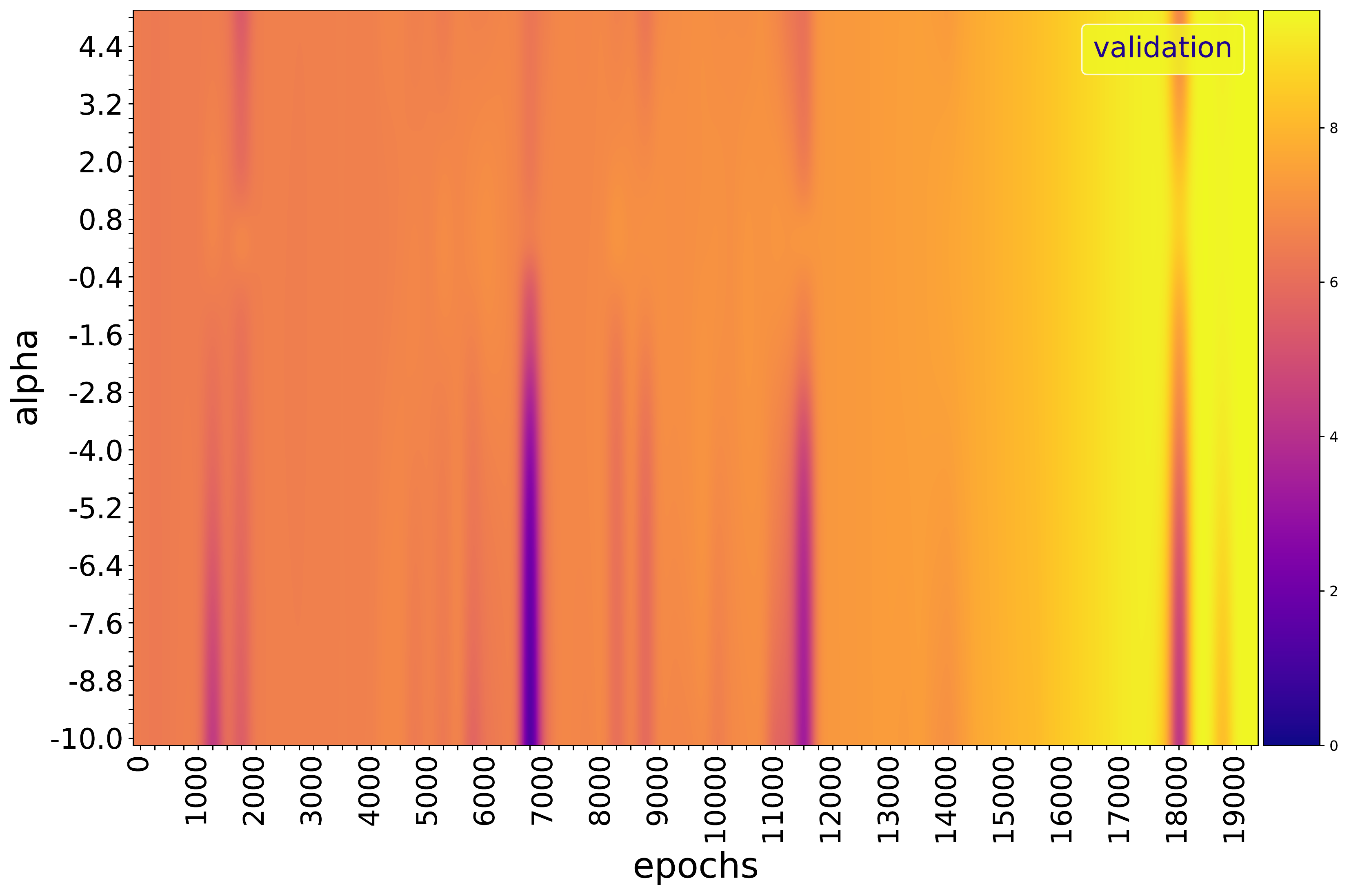}
}
\hfill
\subfigure[Accuracy surface : $f_t(\alpha) = Acc(\theta_t + \alpha \vec{\delta}_t)$ for each epoch $t$]{
\includegraphics[width=\sizefig\linewidth]{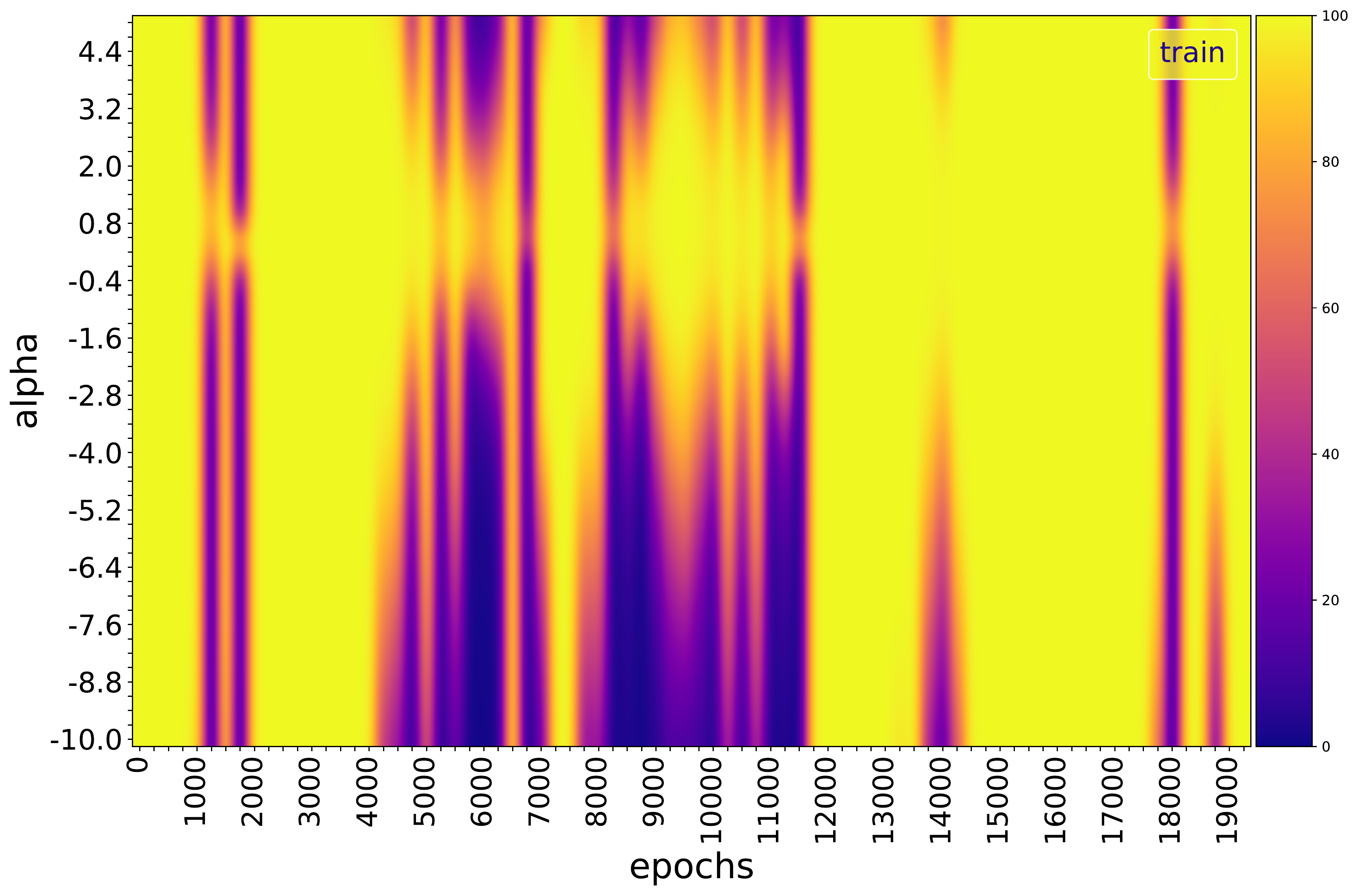}
\includegraphics[width=\sizefig\linewidth]{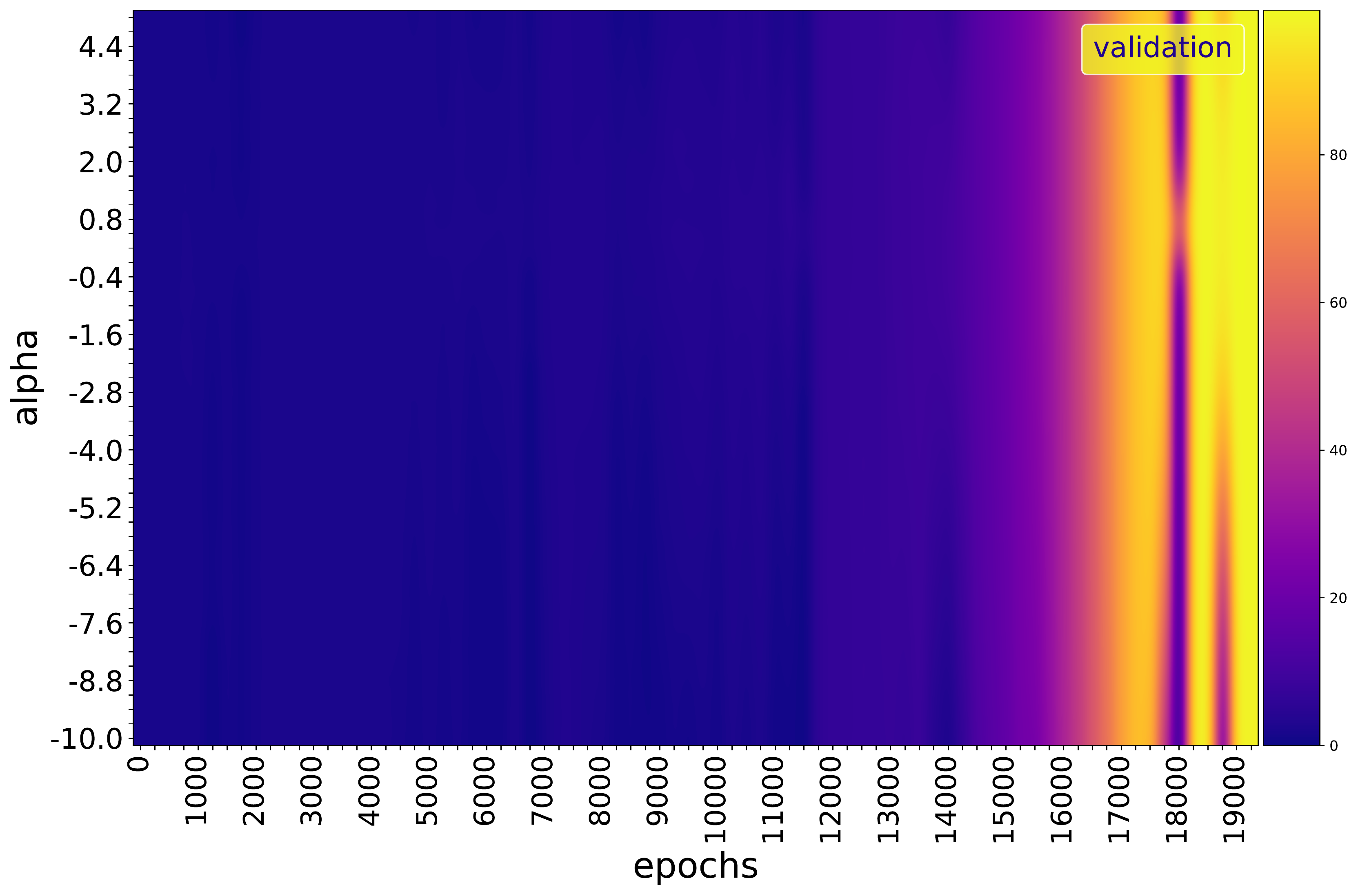}
}
\hfill
\caption{1D projection of the grokking loss/accuracy surface for epoch to epoch, $\vec{\delta}_t \propto \theta_{t+1} - \theta_t \propto - \nabla L (\theta_t)$.
This allows a closer look at the behaviour of the landscape in the (local) direction of the gradient, at each training step $t$. For some $t$, the loss, although disturbed, remains locally flat (up to a resolution used for visualization). These are the instants during which the weight progresses along the directions of least curvature, resulting in a stagnation of the loss. These points are usually followed by a slingshot.
}
\vspace{-0.10in}
\label{fig:30_t_t+1}
\end{figure}


To measure the level of curvature of the loss function, we compute the maximum (${\lambda}_{max}$) and minimum (${\lambda}_{min}$) eigenvalue of its Hessian (more details in the Appendix \ref{sec:curvature}). We observe that there is no significant negative curvature in the trajectory. The curvature remains generally positive and is greatly disturbed at the slingshot points (${\lambda}_{min}$ remains in general close to 0 while ${\lambda}_{max}$ is large, but during slingshot ${\lambda}_{min}$ becomes negative). Our loss is ${\lambda}_{max}$-smooth, i.e. its gradient is ${\lambda}_{max}$-Lipschitz. It is known from optimization literature that a function with bounded Hessian eigen-values has the gradient that tend to decay when the parameter gets closer to the minimum, in constrast to non-smooth one that generaly have abrupt bends at the minimum, which cause significant oscillations for gradient descent \citep{bubeck2015convex}. We also observed that more than 98\% of the total variance in the parameter space occurs in the first 2 PCA modes much smaller than the total number of weights, suggesting that the optimization dynamics are embedded in a low-dimensional space \citep{DBLP:conf/nips/Li0TSG18,doi:10.1073/pnas.2015617118}. Moreover, the model remains in a lazy training regime \citep{chizat2019lazy,berner2021modern} most of the time, as the measure of cosine distance between the model weights from one training step to the next remains almost constant, except at the slingshot location.
These observations seem to support the hypothesis that the model crosses a perturbed valley of bad solutions and experiences disturbances along the direction of the valley. The valley here is just a surface with a large curvature in most directions and no (low) curvature in the rest of the directions. Hence the learning activity of SGD is governed by these directions of high curvature.  When the iterates fall in the valley, we are at the minimum for the training objective (figure \ref{fig:30_t_T}), so that the model can memorize the training data. This local minimum in the valley directions is flat, but in the high curvature directions, is sharp. With the accelerated or adaptative gradient methods, the pathological curvature is no longer a problem, but the local minimum and saddle points are. 
This interpretation of grokking was the working hypothesis of \citet{DBLP:conf/uai/DziugaiteR17}: SGD finds good solutions only if they are surrounded by a relatively large volume of solutions that are nearly as good, as we saw above with grokking optimum that is surrounded by many local minima along the principal directions of curvature. Their work focuses on showing that, under realistic hypotheses, SGD performs implicit regularization or tends to find solutions that possess some particular structural property that we already know to be connected to generalization, like widder minima, that are less difficult to reach by SGD than sharper one as they have a large basin of attraction. Moreover, the loss function here does not seem to satisfy the strict saddle property ($\lambda_{min} < 0$), which guarantees for gradient descent to not converge to saddle points on continuously differentiable functions when using a step size smaller than $1/\lambda_{max}$ \citep{lee2016gradient}. The dynamic, therefore, has components in the unstable subspace of the Hessian (a sub-space spanned by the dimensions of non-negative eigenvalues), and if $\theta_0$ is chosen out of this space (which happens with zero probability for a random initialization), there is hope that we will quickly converge to a global solution. The spectral signature reflects the dynamics in and out of this unstable sub-space. The strict saddle property is indeed verified only when there are strong oscillations.

\section{Related works 
}
\label{sec:related_works}

\paragraph{Grokking} 
\citet{power2022grokking} are the first to have studied and named the grokking phenomenon. They train a decoder-only causal transformer \citep{DBLP:journals/corr/VaswaniSPUJGKP17} to predict the response $c$ of a binary operation of the form $a \circ b=c$, where $a$, $b$, $c$ are discrete symbols without internal structure, and $\circ$ a binary operation (addition, composition of permutations, bivariate polynomials, etc). \citet{Liu2022TowardsUG} studied grokking in simple toy models, and embedding layer followed by a multilayer perceptron, for classification and regression. The authors empirically show that generalization on algorithmic datasets coincides with the emergence of structure in embeddings, that the quality of representation predicts generalization, and they defined the notion of representation quality in the toy setting and shown that it predicts generalization by developing an effective theory to describe the learning dynamics of the representations in the same toy setting. They also illustrate phase diagrams including memorization, comprehension, confusion and grokking (similar to comprehension, but with a very large number of training steps between memorization and generalization) as a function of hyperparameters. The taxonomy used throughout this paper is identical to theirs, with the difference that we refer in our work to the phases along the training trajectory. We examined in more detail the loss landscape during these different training phases, but we do not analyze the embedding structure. 
\citet{Thilak2022TheSM}'s slingshot mechanism is quite correlated with our hypotheses about oscillations. Their work focuses on characterizing periodic oscillations observed in the different phases of model training, and they were the first to empirically observe that grokking almost exclusively happens at the onset of slingshots, and is absent without it. Our work reveal that these oscillations are caused at the level of the landscape by the escape of sharp mountains surrounding local optima. Moreover, the model weights change abruptly during slingshot, and slightly outside the slingshot. Other works have also been interested in grokking recently, such as the mechanistic interpretability of grokking \citep{nanda2022grokking} and grokking as a phase transition \citep{phase2022grokking}.

\paragraph{Neural networks loss landscape}

The visualization methods use in this paper was first introduced by \citet{DBLP:journals/corr/GoodfellowV14}, and is still one the most use method in date to study the effect of loss landscapes on generalization \citep{DBLP:journals/corr/GoodfellowV14,im2016empirical,smith2017exploring,DBLP:conf/iclr/KeskarMNST17,DBLP:conf/nips/Li0TSG18,doi:10.1073/pnas.2015617118}. \citet{DBLP:conf/nips/Li0TSG18} observe that, when networks become sufficiently deep, neural loss landscapes quickly transition from being nearly convex to being highly chaotic, and that this transition coincides with a dramatic drop in generalization error, and ultimately to a lack of trainability. \citet{doi:10.1073/pnas.2015617118} investigate the connection between the the loss function landscape and the stochastic gradient descent (SGD) learning dynamics. They show that around a solution found by SGD, the loss function landscape can be characterized by its flatness in each PCA direction. As \citet{DBLP:conf/nips/Li0TSG18}, they show that optimization trajectories lie in an extremely low dimensional space.


\paragraph{Flatness, Sharpness} \citet{10.1162/neco.1997.9.1.1} defined a flat minimizer as a (large) region in weight space with the property that each weight vector from that region leads to similar small error, and propose an algorithm to search such a minimizer, called flat minimum search. Similarly, we can say that a minimizer is sharp when the loss function has a (very) high condition number near it, at any point in its basin of attraction. \citet{DBLP:conf/iclr/KeskarMNST17} characterize flatness using eigenvalues of the Hessian, and empirically show that  large-batch methods make the minimizers of training and testing functions sharp while small batch methods make them flat. \citet{10.1162/neco.1997.9.1.1} and \citet{DBLP:conf/iclr/KeskarMNST17} methods to quantify sharpness are not invariant to symmetries in the network, so, are not sufficient to determine generalization ability, as show by \citet{DBLP:conf/icml/DinhPBB17,DBLP:conf/nips/NeyshaburBMS17}. 


\paragraph{Related phenomena} Grokking can be seen as an extreme gradient starvation phenomenon \citep{NEURIPS2021_0987b8b3} or a very slow double descent \citep{DBLP:conf/iclr/NakkiranKBYBS20}, for which the model only manages to learn the features necessary for generalization in the terminal phases of training \citep{pezeshki2021multiscale}. In fact, we observe that the
validation loss exhibits a double descent behaviour with an initial decrease, then a growth before rapidly decreasing to zero when the model groks. Grokking can also be seen as a phase transition (see \citep{nanda2022grokking,phase2022grokking}). Regarding catastrophic forgetting, there is a link between it when learning many tasks and the sharpness of the optimum for each task; 
so that the slightest update to one task makes the optimum escape from its basin of attraction for the other tasks. 
\citet{mirzadeh2020understanding,mirzadeh2021wide} formalize this.

\section{Summary and Discussion}
\label{sec:conclusion}

We made the following observations:

\noindent 1. The memorization phase is characterized by a perturbed landscape, and it is separated from comprehension by a perturbed valley of bad solutions. 
Small data results in the slow progression of SGD in this region, causing a delay in generalization. During the comprehension phase, the loss and accuracy of training and validation show a periodic perturbation. \citet{Thilak2022TheSM} named a related phenomenon the \textit{slingshot mechanism}. We found that these perturbation points are characterized at the level of loss (resp. accuracy) by a sudden increase-decrease  (resp. decrease-increase), at the level of the model weights by a sudden variation of the relative cosine similarity, and at the level of the loss landscape by obstacles. This last point goes against what \citet{DBLP:journals/corr/GoodfellowV14} observed, namely that a variety of state-of-the-art neural networks never encounter any significant obstacles from initialization to solution. The slingshot mechanism also contradicts the idea that SGD spends most of its time exploring the flat region at the bottom of the valley surrounding a flat minimizer \citep{DBLP:journals/corr/GoodfellowV14}, since it goes with the model from confusion to the terminal phase of training, even after the model generalized, for many datasets.
    
\noindent 2. The Hessian of the grokking loss function is characterized by larger condition numbers, leading to a slower convergence of gradient descent. We observed that more than 98\% of the total variance in the parameter space occurs in the first 2 PCA modes much smaller than the total number of weights, suggesting that the optimization dynamics are embedded in a low-dimensional space \citep{DBLP:conf/nips/Li0TSG18,doi:10.1073/pnas.2015617118}. Moreover, the model remains in a lazy training regime \citep{chizat2019lazy,berner2021modern} most of the time, as the measure of cosine distance between the model weights from one training step to the next remains almost constant, except at the slingshot location.

From the point of view of the landscape, grokking seems a bit clearer: landscape geometry
has an effect on generalization, and can allow in the early stages of training to know if the model will generalize or not by  just looking at a microscopic quantity characteristic of that landscape like the empirical risk. 

\paragraph{Limitations} Some limitations  that would require further scrutiny include:
    
\noindent 1. We view the loss surface with a dramatic reduction in dimensionality, and we need to be careful in interpreting these graphs. 

 \noindent 2. We are only optimizing an unbiased estimate of the total loss function (added up across all training examples), whose structure may be different from the global loss function, so that the delay in generalization is an effect of each individual term in the loss function, or of the noise induced by the sampling of these terms, or of the stochasticity in the initialization of weights, generation of dropout masks, etc \citep{DBLP:journals/corr/GoodfellowV14}. More research needs to be done to find out if it is indeed the fact that the loss surface is perturbed that causes the generalization in the long run.

\paragraph{Perspectives} The future direction is to study what a feature learning-based explanation of double descent \cite{pezeshki2021multiscale} has in common with grokking and to see if it is possible to design a specific learning curve as \citet{chen2021multiple} derived for the generalization curve. Also, this work focused on the detection of grokking however ultimately pave the way to also induce grokking. Another path of development is to move from arithmetic tasks with deterministic answers 
to language models and computer vision with more probabilistic answers.

\nocite{langley00}

\section*{Acknowledgments}
We acknowledge the support from Canada CIFAR AI Chair Program and from the Canada Excellence Research Chairs Program.  
 This research used computing resources   on  Summit supercomputer at the Oak Ridge Leadership Computing Facility, which is a DOE Office of Science User Facility supported under Contract DE-AC05-00OR22725. The allocation was granted to  the INCITE project on ``Scalable Foundation Models for Transferable Generalist AI”.
\bibliography{example_paper}
\bibliographystyle{icml2023}

\appendix


\section{
Grokking
}
\label{subsec:grokking_appendix}

\subsection{Training phases}

Previous work \citep{power2022grokking,Liu2022TowardsUG} used the terms confusion, memorization and comprehension in the phase diagram based on different hyperparameters, but 
in this paper they also refer to the phases along the training trajectory. Conventional deep neural networks training is split into two phases \citep{shwartz-ziv2017opening,DBLP:conf/iclr/NakkiranKBYBS20,doi:10.1073/pnas.2015617118}. A first phase during which the network learns a function with a small generalization gap and a second phase during which the network starts to overfit the data, leading to an increase in the test error. However, \citet{DBLP:conf/iclr/NakkiranKBYBS20} show that in some regimes, the test error decreases again and can reach a lower value at the end of the training compared to the first minimum, suggesting potential training phases to exploit. \citet{doi:10.1073/pnas.2015617118} distinguish two phases, an initial fast learning phase where the loss function decreases quickly and sometimes abruptly, followed by an exploration phase 
when the training error reaches its minimum value and the overall loss still decreases, but much more slowly and gradually.  However, considering just two phases does not allow to study grokking properly, as the whole ingredient of grokking is in the memorization phase and in how the model gets in this phase and out.

\subsection{Non-grokking}

\begin{figure}[htp]
\centering
\includegraphics[width=1.\linewidth]{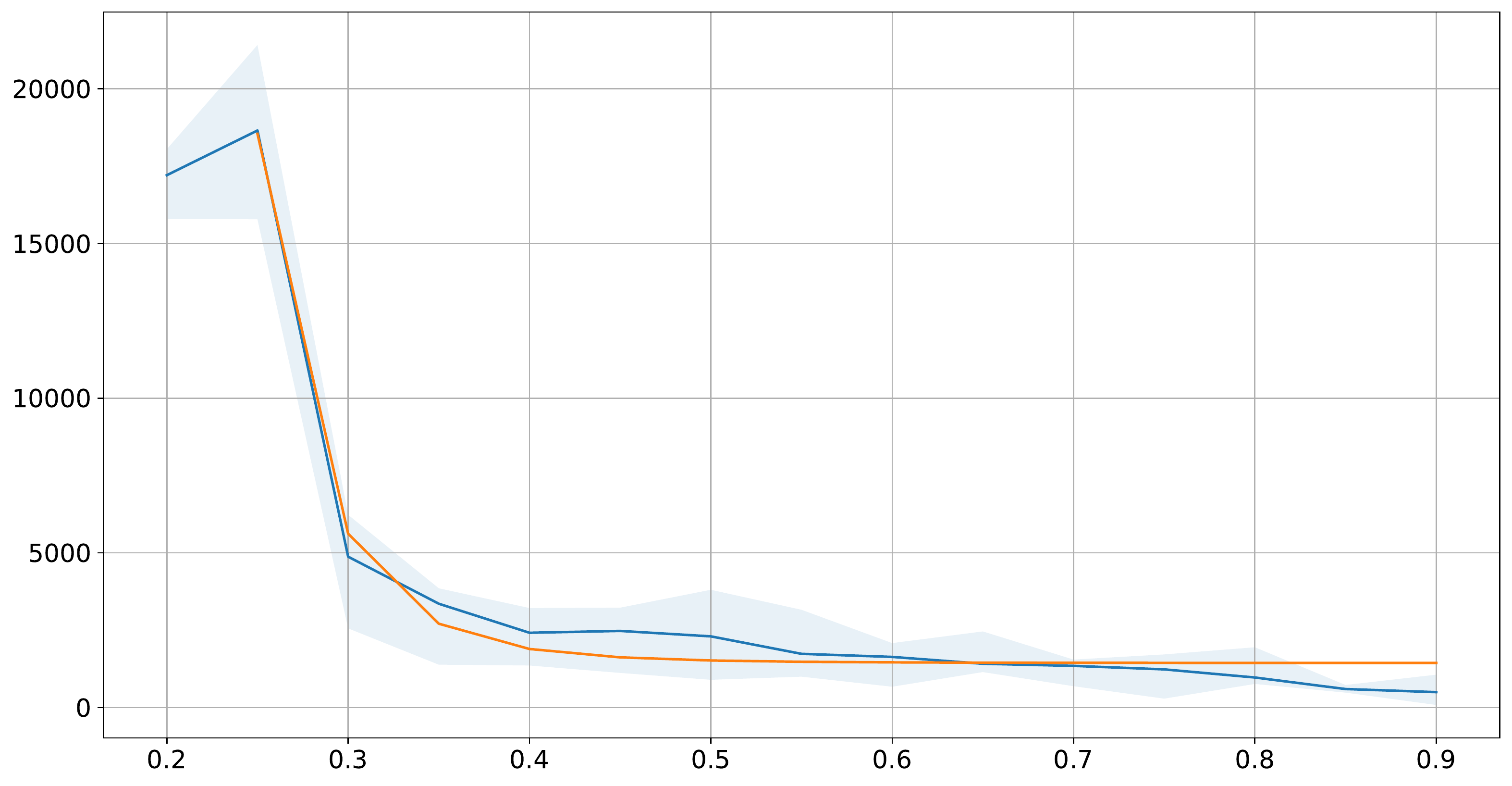} 
\hfill
\vspace{-0.02in}
\caption{ 
$t_4(r)$ is a transition step from memorization to generalization for each training data fraction $r >= 0.2$. The \textcolor{tab:blue}{blue} curve shows the empirically obtained values of $t_4(r)$ (y-axis) as a function of $r$ (x-axis), and the \textcolor{orange}{orange} curve represents an estimate of the form $t_4(r) = a r ^{-\gamma} + b$ fitted on the obtained data, with $(\gamma, a, b) = (7.73, 1.09 \times 10^{15}, 1442.63)$ in this 
case. 
}
\label{fig:phases}
\end{figure}

\begin{figure}[tbh]
\hfill
{\includegraphics[width=0.49\linewidth]{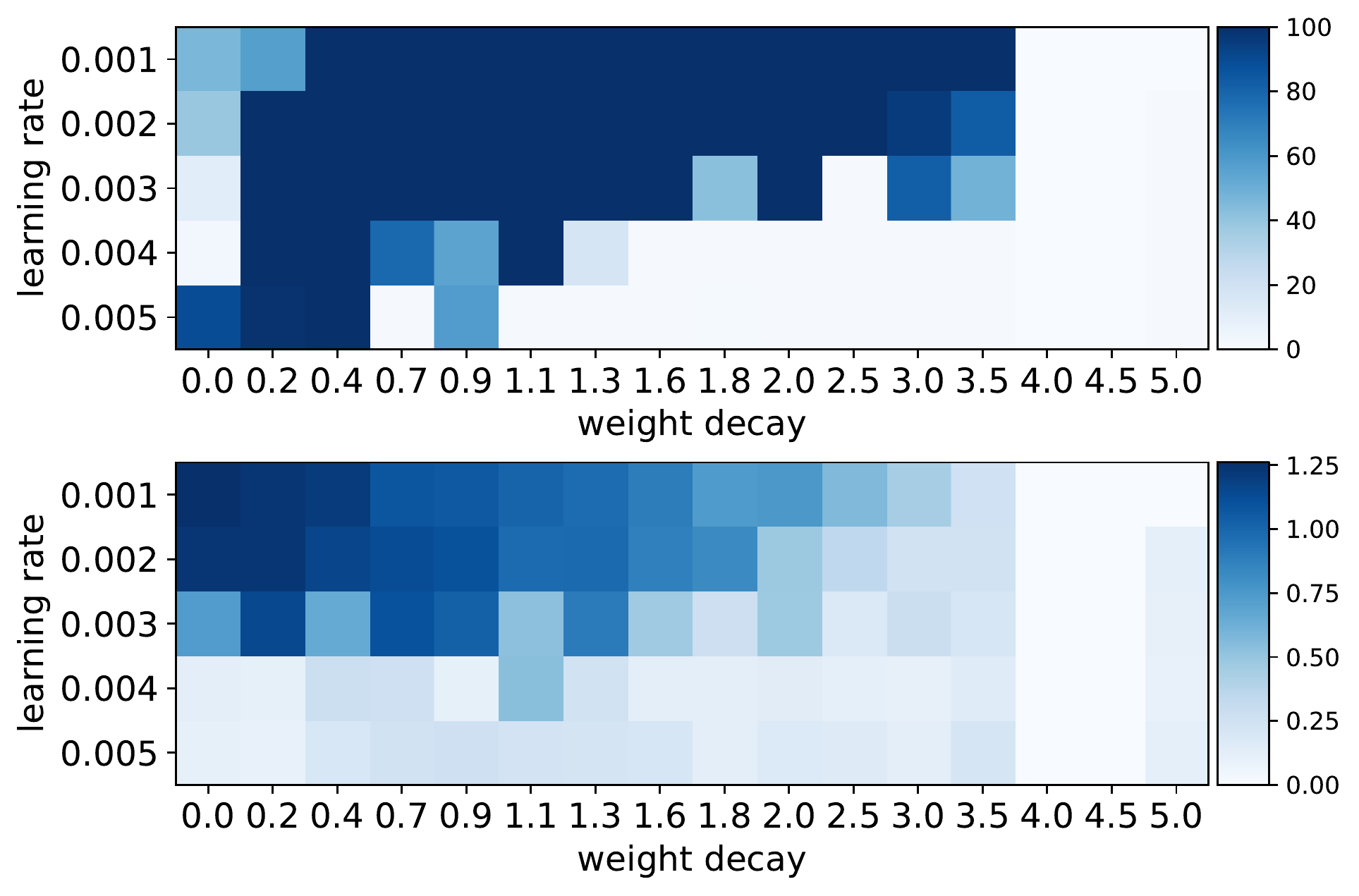}}
\hfill
\hfill
{\includegraphics[width=0.49\linewidth]{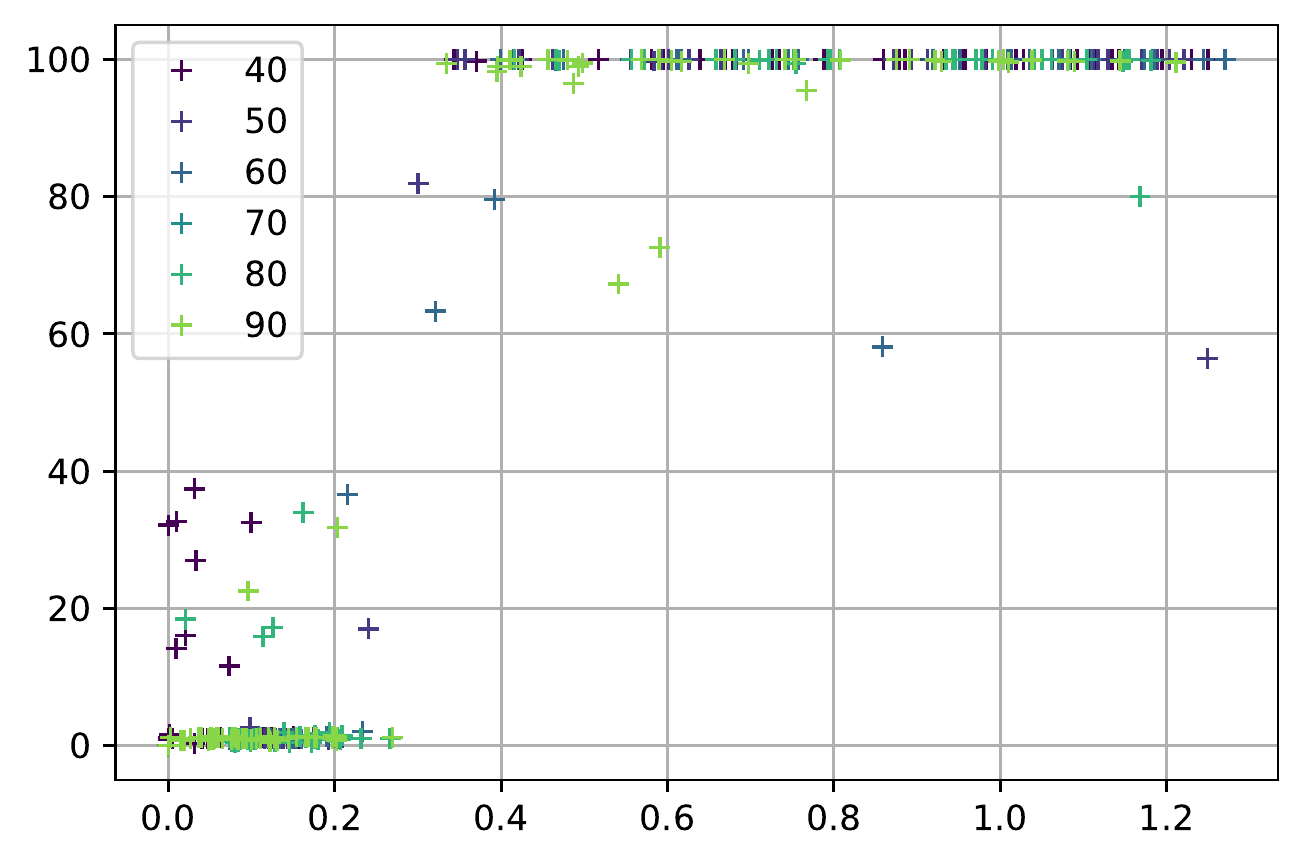}}
\hfill
\vspace{-0.02in}
\caption{Left) The first figure (top) represents the validation accuracy (\%) at the end of the training ($10k$ steps), and the second figure (bottom) represents the spectral energy (activity) in the training loss for the first 400 training steps ($r = 0.5$). On the x-axis we have the weight decay strength, and on the y-axis we have the learning rate. Right) On the x-axis we have the (normalize) activity (400 steps), and on the y-axis we have the validation accuracy (\%), for different value of $r$.
On the x-axis we have the weight decay strength, and on the y-axis we have the learning rate. A similarity is observed between the oscillation patterns in the training loss during the initial stages of training and the validation accuracy. This suggests that the spectral signature can be used as an indicator or proxy for the upcoming grokking phenomenon. 
The highest degree of generalization is typically observed when using small learning rates and small weight decay. While large learning rates may increase oscillations, this does not directly lead to grokking and is not necessarily evident in the early stages of training. Instead, such effects become more noticeable near the basin of attraction of the minimum.
}
\label{fig:spect_diagram_1}
\end{figure}

\begin{figure}[tbh]
\hfill
\subfigure[$r=0.4$]
{\includegraphics[width=0.49\linewidth]{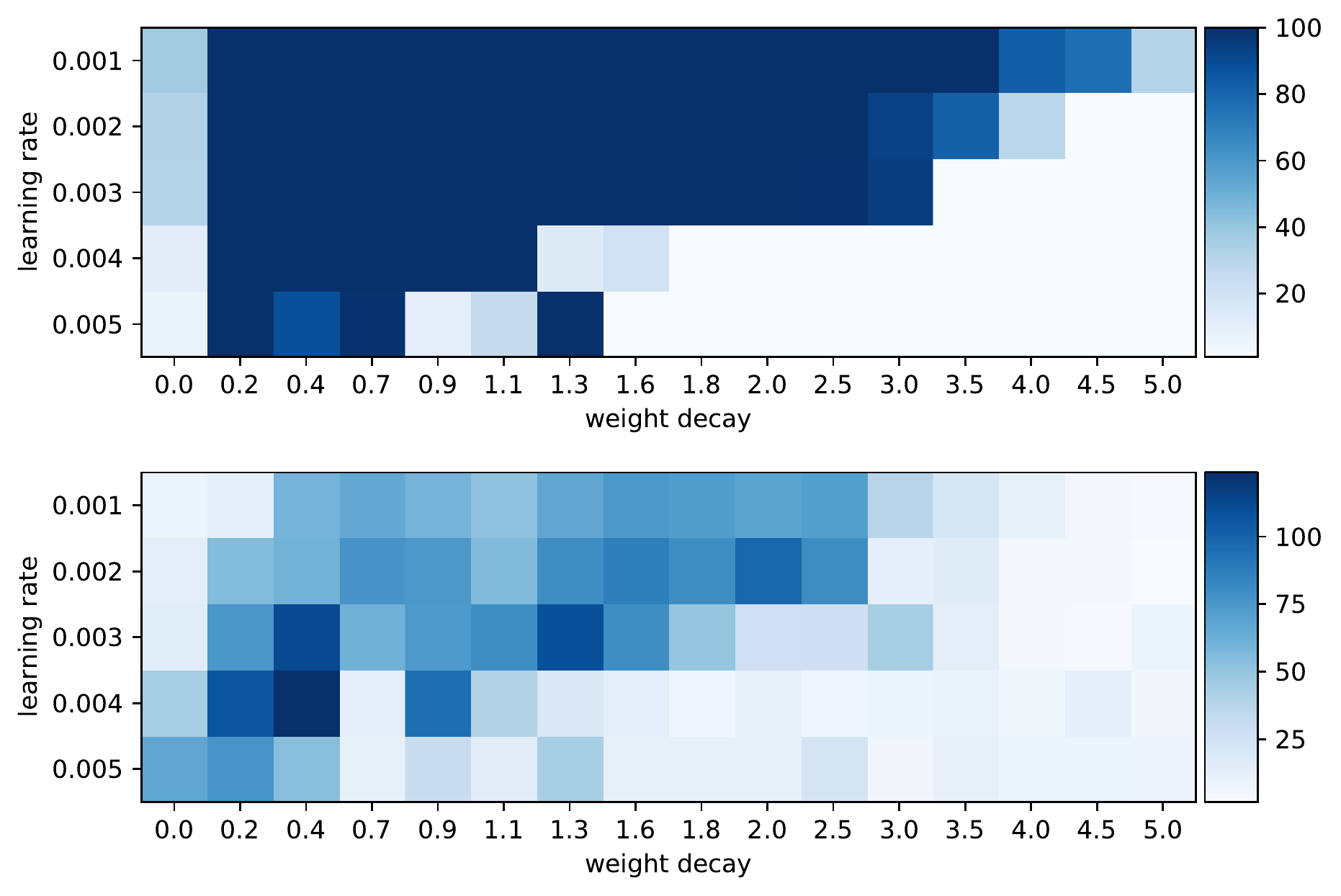}}
\hfill
\subfigure[$r=0.5$]
{\includegraphics[width=0.49\linewidth]{images/spect_diagram/spect_diagram_50.pdf}}
\hfill
\subfigure[$r=0.7$]
{\includegraphics[width=0.49\linewidth]{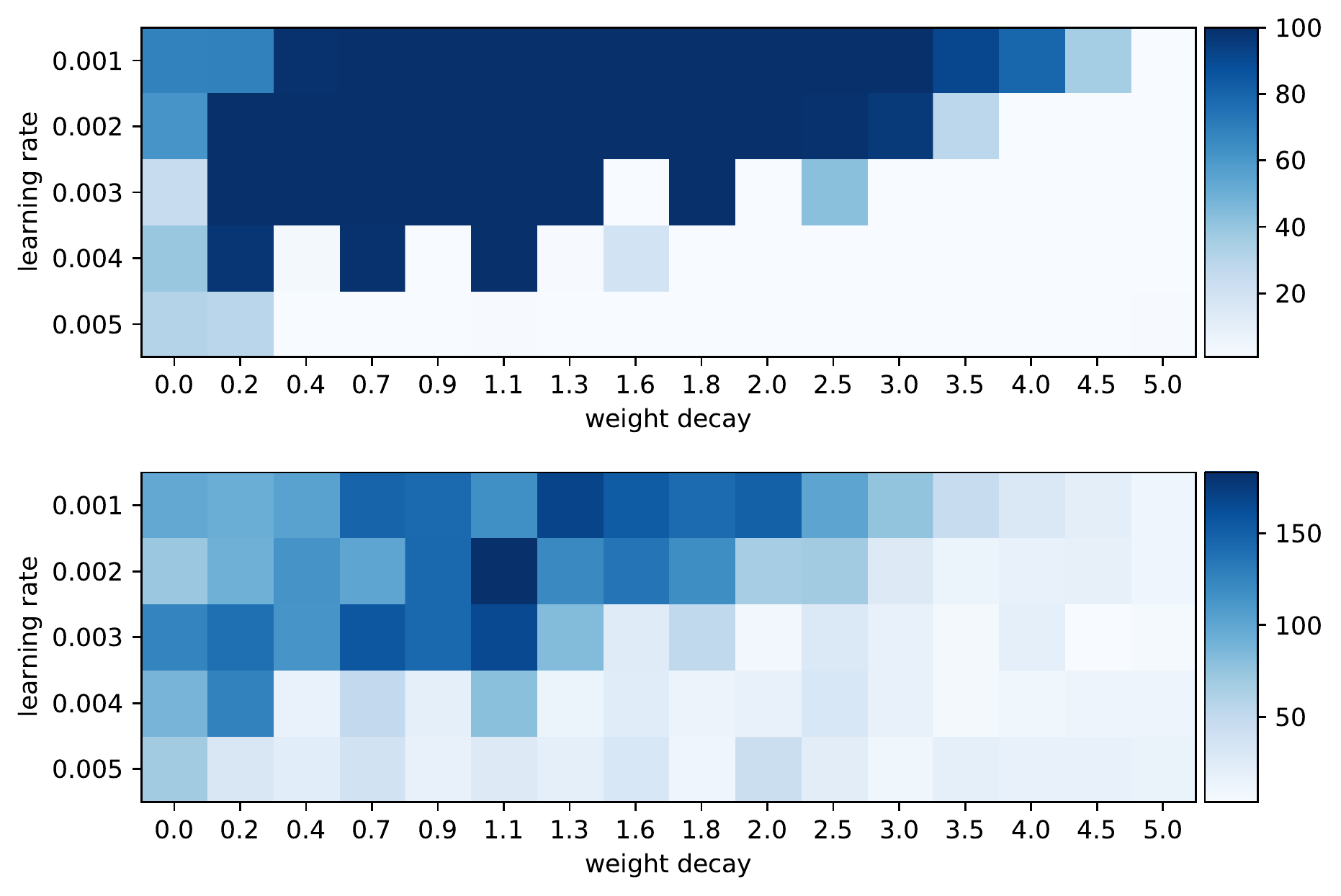}}
\hfill
\subfigure[$r=0.9$]
{\includegraphics[width=0.49\linewidth]{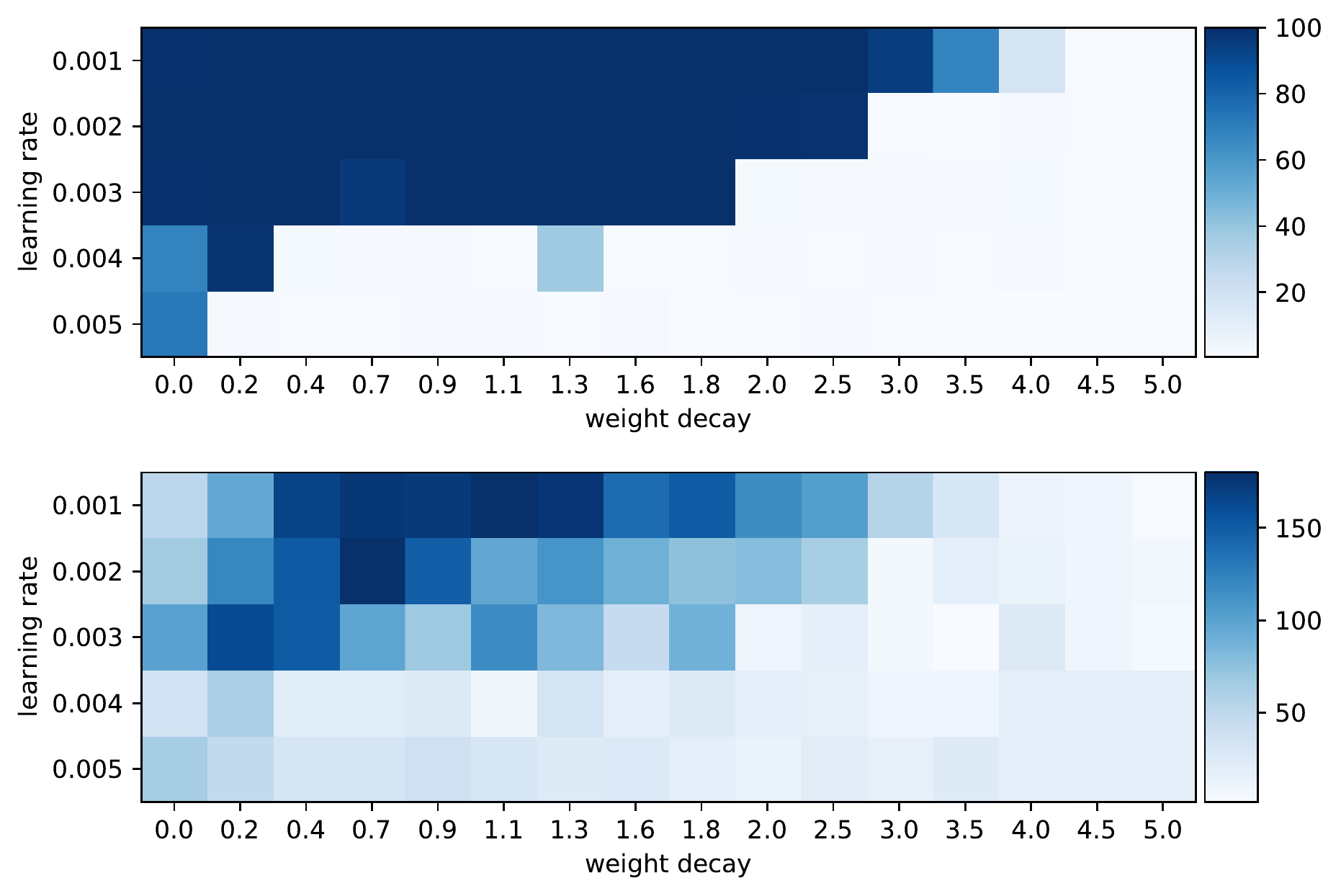}}
\hfill
\vspace{-0.02in}
\caption{
For (a)-(d), the first figure (top) represents the validation
accuracy ($\%$) at the end of the training (10k steps), and
the second figure (bottom) represents the spectral energy
(activity) in the training loss for the first 400 training steps. On the x-axis we have the weight decay strength,
and on the y-axis we have the learning rate. A similarity is observed
between the oscillation patterns in the training loss during
the initial stages of training and the validation accuracy.}
\label{fig:spect_diagram_2}
\end{figure}

Although it is easy to identify grokking, it is very difficult to give a formal definition of its opposite since nothing contradicts the fact that if the model is allowed to train indefinitely it will not eventually generalize. Knowing the hyperparameters (learning rate, weight decays strength, ...) that allow to have grokking (
let's note by $H_g$ the set of values of such hyperparameters), we use them to train our model for different data sizes. Then we fit a function that predicts $t_4$, the generalization step, for each training data fraction $r$.  We then used the expression of $t_4(r)$ estimated as a proxy of the number of training steps needed to get the generalization for each $r$. That is, if for other hyper-parameters that are not in $H_g$, and for a given training data fraction $r$, we train a model for $t_4(r)+\epsilon$ (we used $\epsilon = 1k$) steps and have no generalization, we can stop the training. This does not necessarily imply that there will be no grokking if the model is left in training for longer.  The limit of this approach is that $t_4(r)$ is just an empirical law estimated for $r \ge r_{min} > 0$, which collapses when $r \rightarrow 0$.

\subsection{The over-parameterization ratio and the effect of the optimizer parameters}

In general, more data leads to faster grokking (figure 
\ref{fig:phases}-b
), i.e. $t_2(r)$ and $t_4(r)$ are a decreasing functions of $r$. Empirically, $t_4$ follows a power law of the form $t_4(r) = a r ^{-\gamma} + b$. This was first predicted by \citet{phase2022grokking}. For $r \ge r_{min}$, we found $(\gamma, a, b) = (7.73, 1.09 \times 10^{15}, 1442.63)$ for modular addition (figure \ref{fig:phases}.b), and $(\gamma, a, b) = (1.18, 1.85 \times 10^{6}, 0.0)$ for multiplication in $S_5$, in $H_g$. 
Smaller learning rates require more training steps for convergence whereas larger learning rates result in rapid changes and require fewer training epochs. But a learning rate that is too large can cause the model to converge too quickly to a suboptimal solution, whereas a learning rate that is too small can cause the process to get stuck. This is well verified with grokking, because the more we increase the learning step, the faster we observe grokking, up to a threshold that depends on the value of the weight decay strength used. 

\subsection{Spectral signature}

Figures and \ref{fig:spect_diagram_1} and \ref{fig:spect_diagram_2} show a similarity between the oscillation in the training loss in the early phases of training and the validation accuracy for many values of $r$. 



\section{Loss landscape}
\label{sec:landscape_appendix}

\begin{figure}[h]
\hfill
\subfigure[Loss surface : $f_t(\alpha) = Loss(\theta_t + \alpha \vec{\delta}_t)$ for each epoch $t$]{
\includegraphics[width=\sizefig\linewidth]{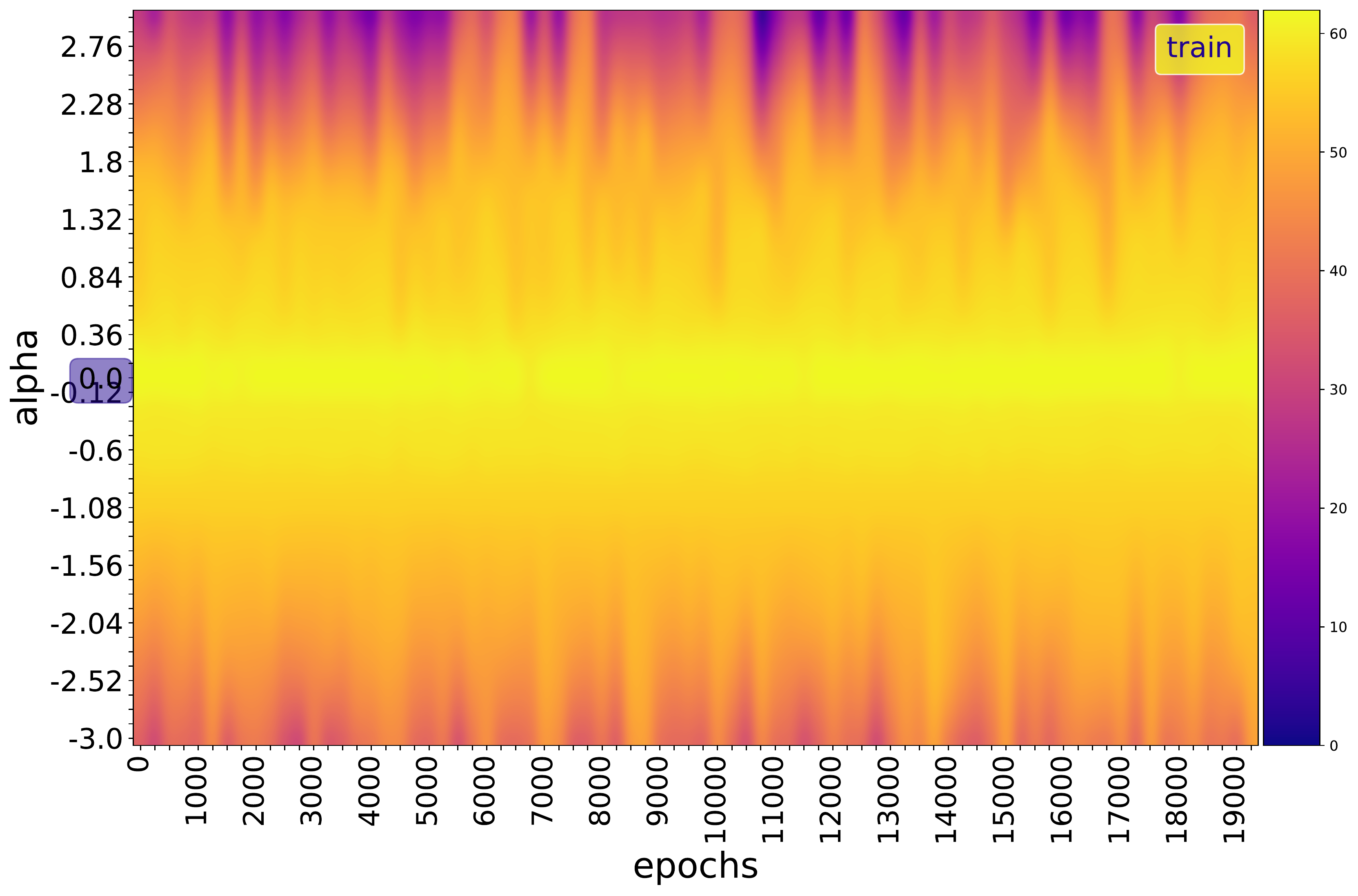}
\includegraphics[width=\sizefig\linewidth]{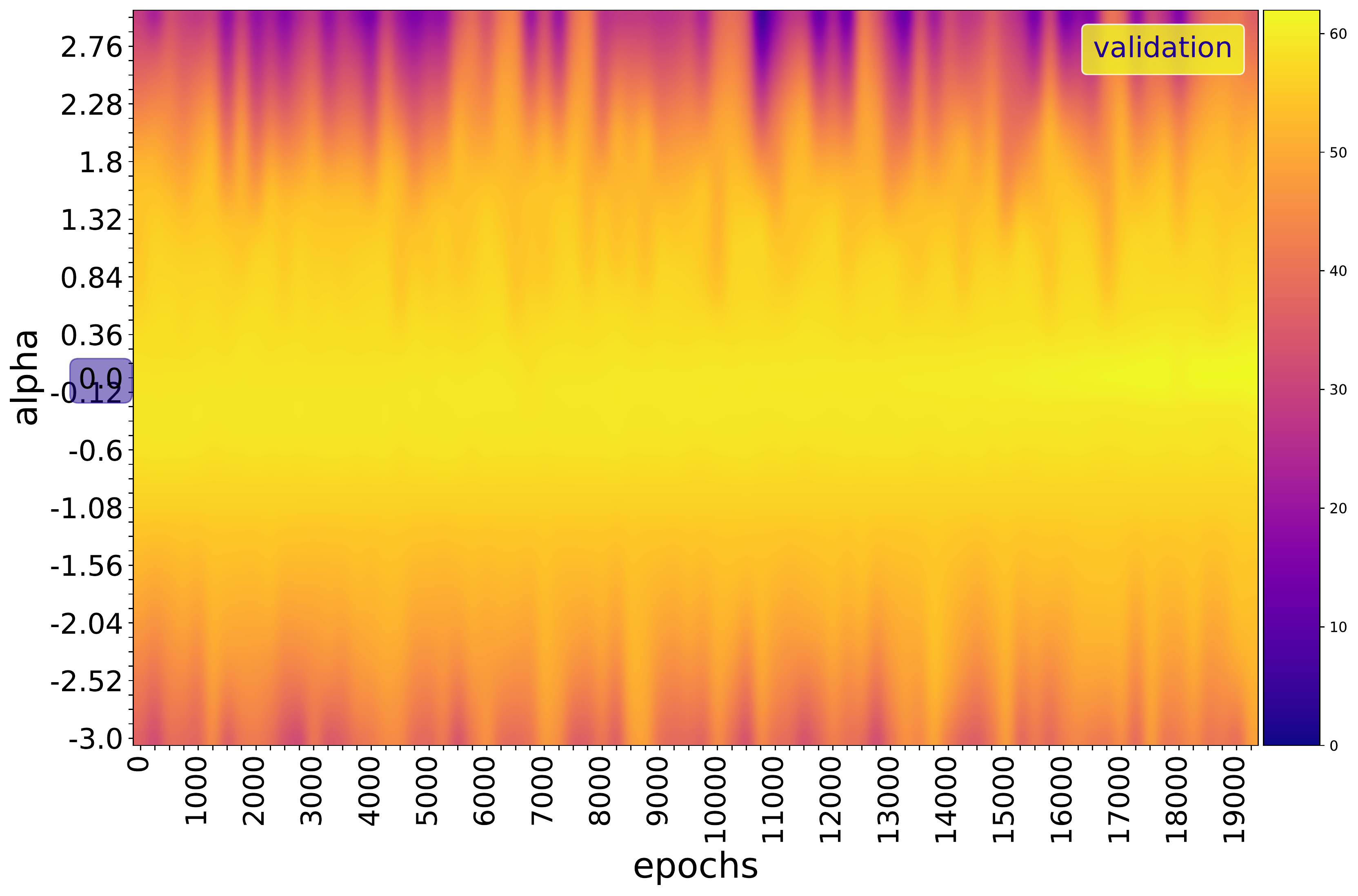}
}
\hfill
\subfigure[Accuracy surface : $f_t(\alpha) = Acc(\theta_t + \alpha \vec{\delta}_t)$ for each epoch $t$]{
\includegraphics[width=\sizefig\linewidth]{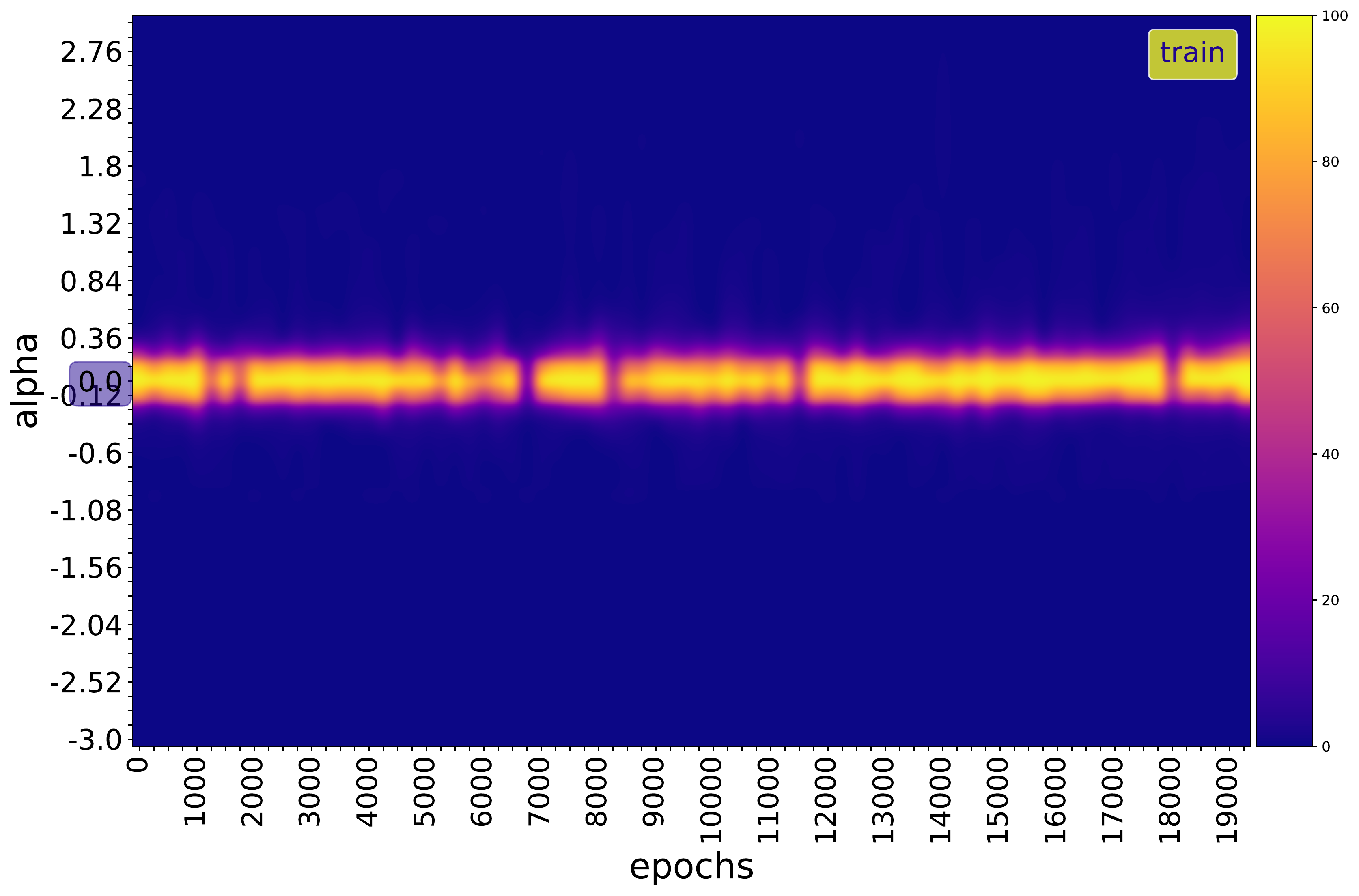}
\includegraphics[width=\sizefig\linewidth]{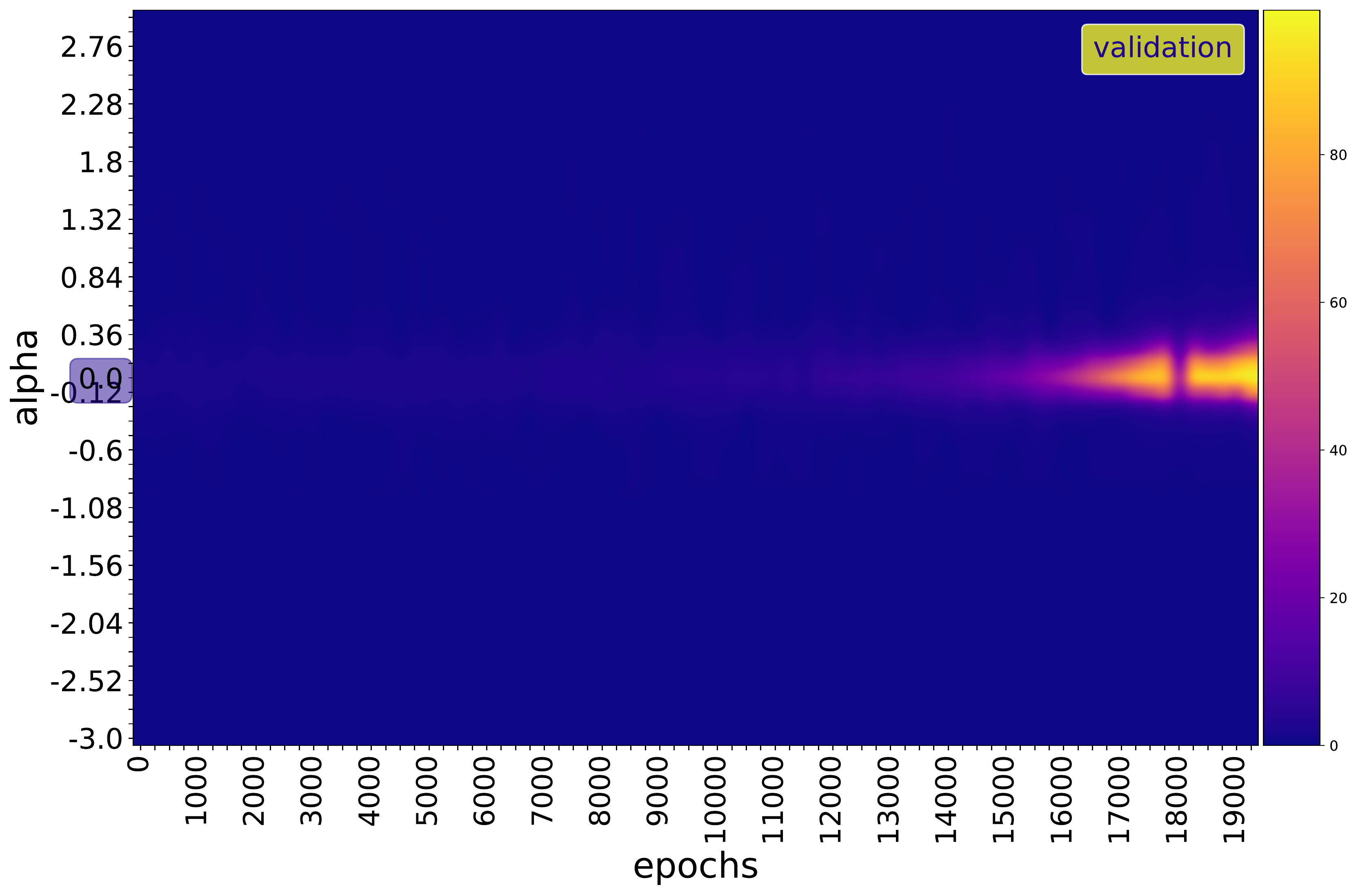}
}
\hfill
\caption{1D projection of the grokking loss/accuracy surface in a random direction $\vec{\delta}$. In this direction, no obstacles are visible along the trajectory, and the surface is rather well-conditioned.
}
\vspace{-0.10in}
\label{fig:rand_30}
\end{figure}

\begin{figure}[h]
\hfill
\subfigure[Loss surface : $f_t(\alpha) = Loss(\theta_t + \alpha \vec{\delta}_t)$ for each epoch $t$]{
\includegraphics[width=\sizefig\linewidth]{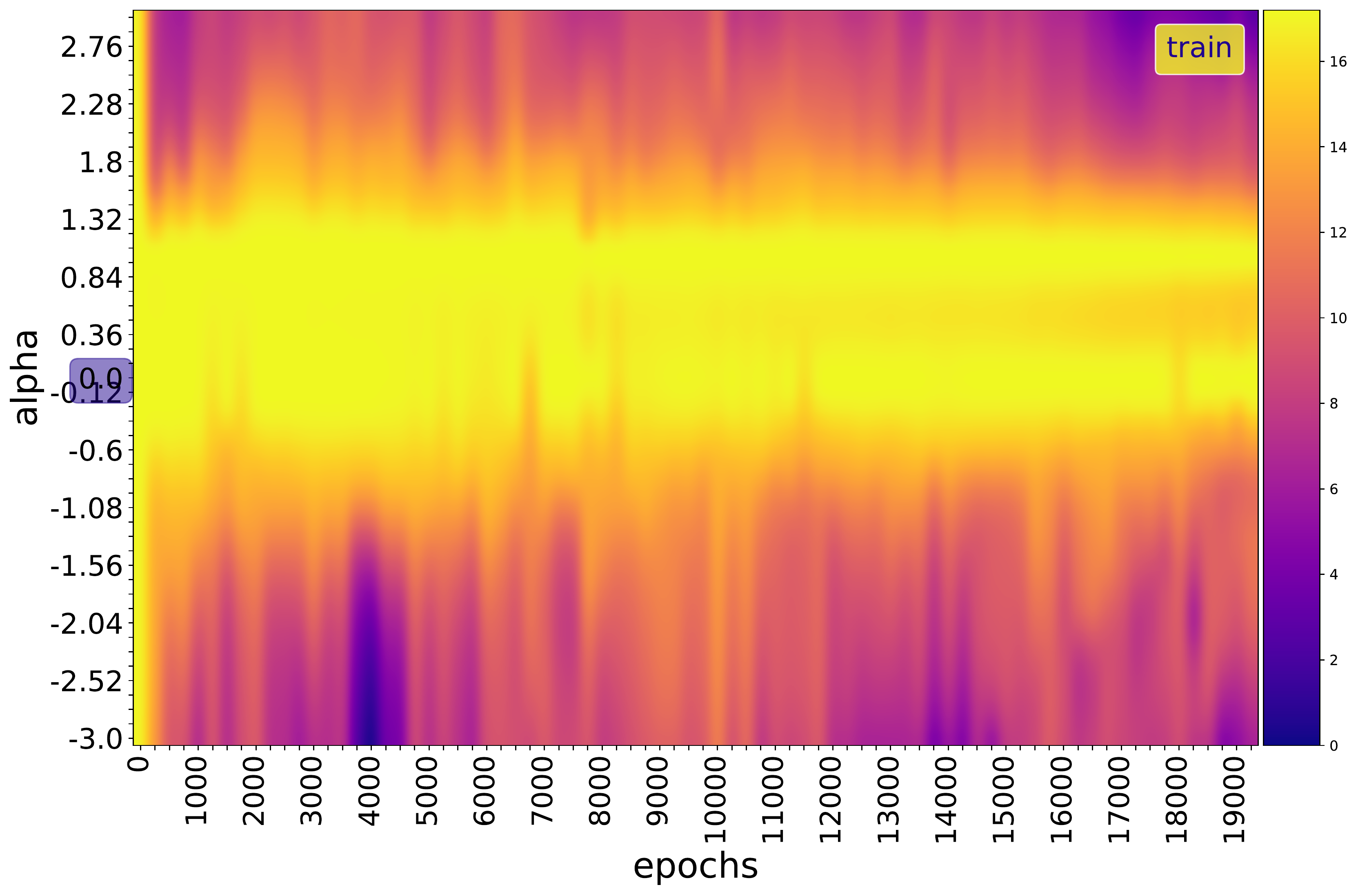}
\includegraphics[width=\sizefig\linewidth]{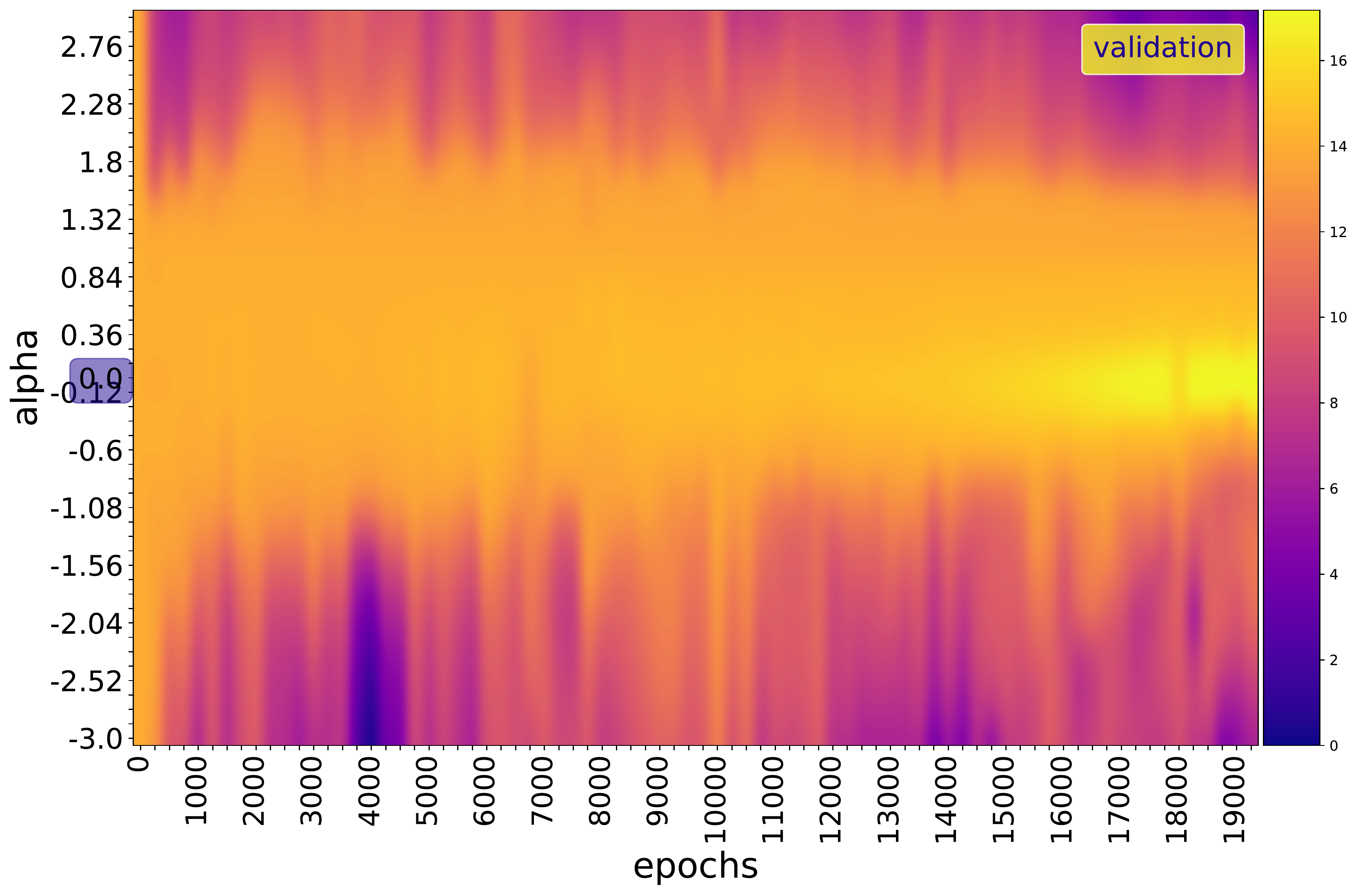}
}
\hfill
\subfigure[Accuracy surface : $f_t(\alpha) = Acc(\theta_t + \alpha \vec{\delta}_t)$ for each epoch $t$]{
\includegraphics[width=\sizefig\linewidth]{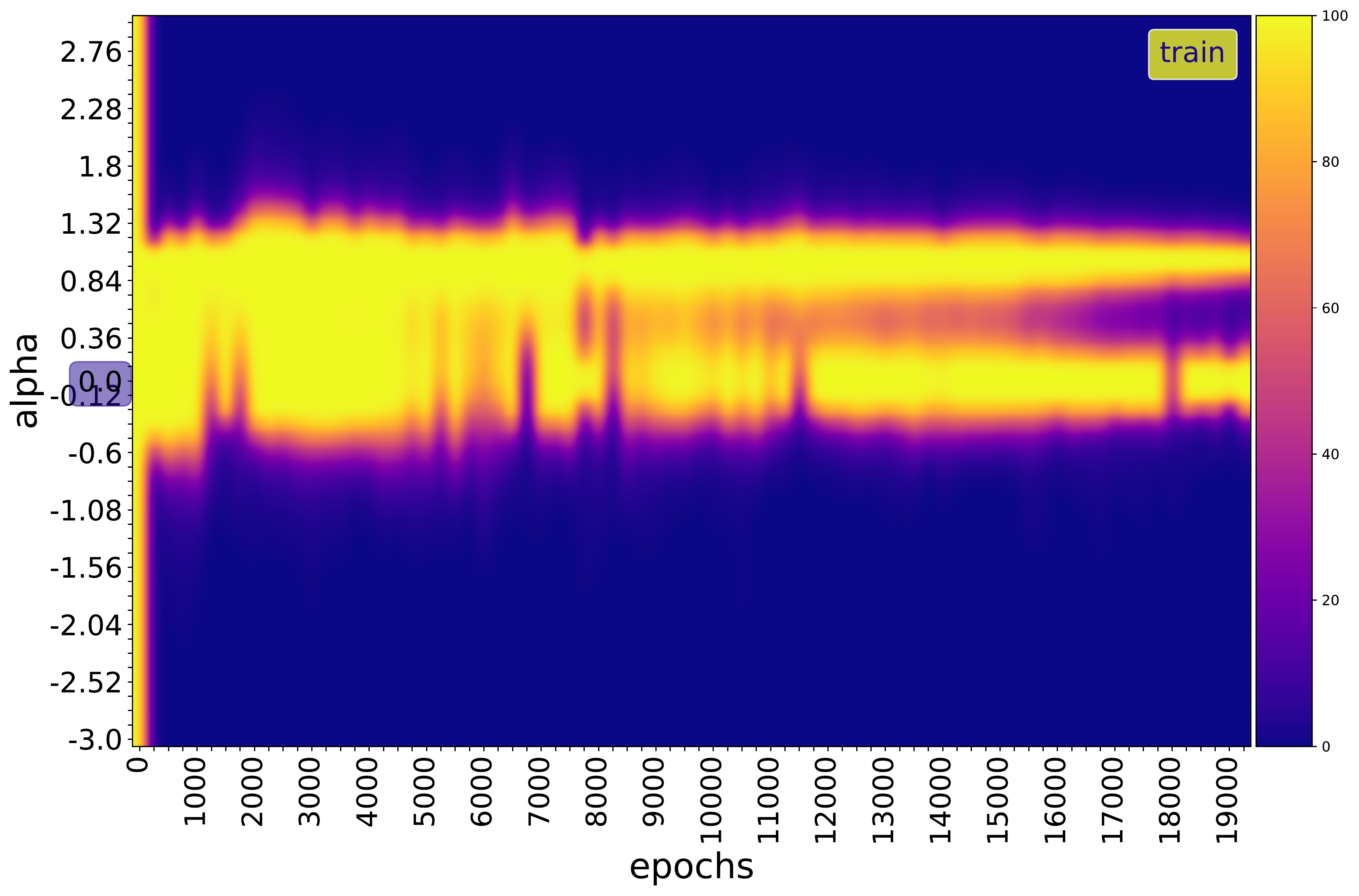}
\includegraphics[width=\sizefig\linewidth]{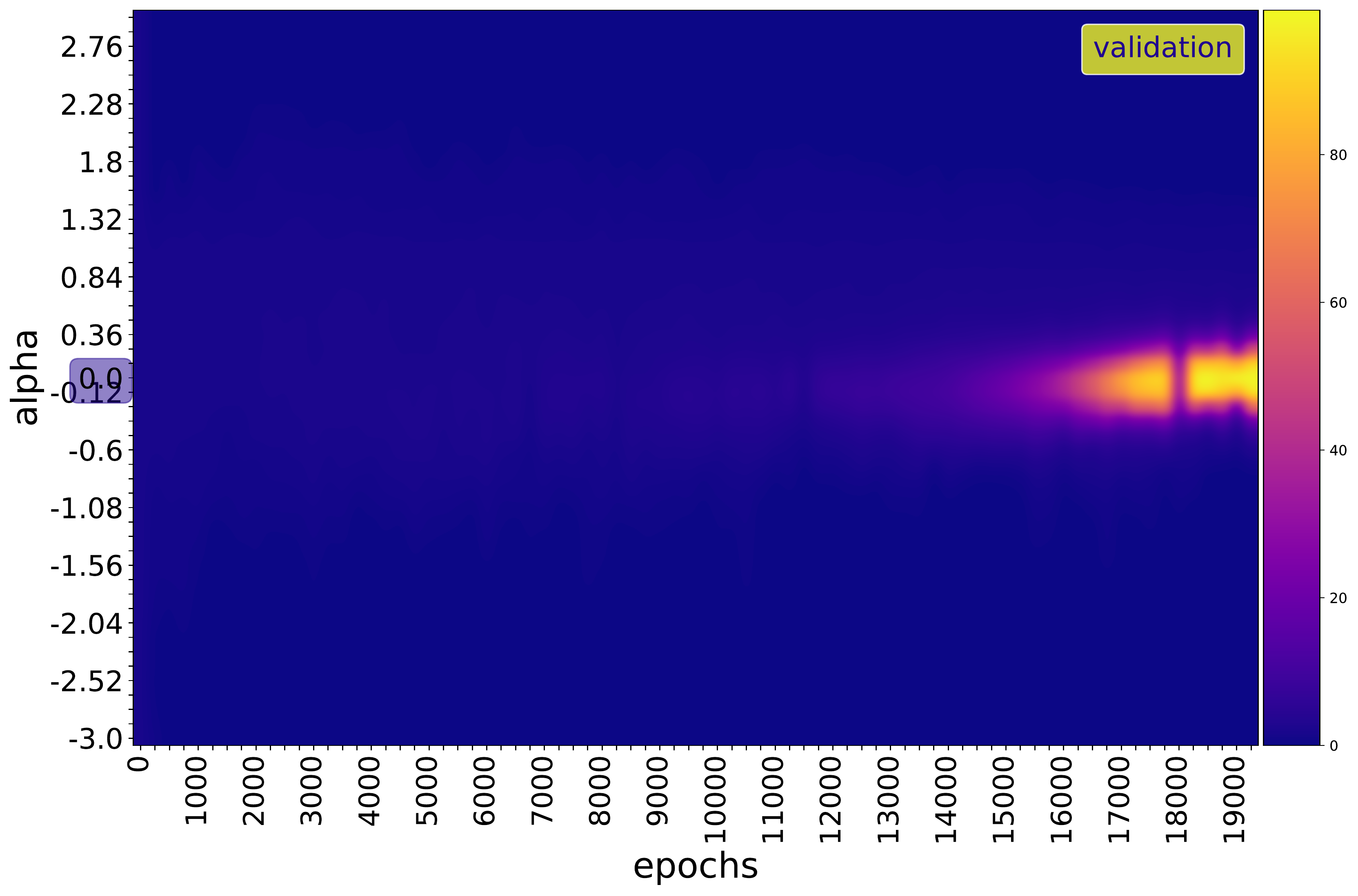}
}
\hfill
\caption{1D projection of the grokking loss/accuracy surface from the initialization point to any step $t$, $\vec{\delta}_t \propto \theta_0 - \theta_t$.
In this direction, each point considered behaves locally as a minimum, and as one approaches the optimum, local minima become preponderant. 
}
\vspace{-0.01in}
\label{fig:t0_30}
\end{figure}

For $r = 0.3$, figure \ref{fig:cover} shows the 1D projection of the grokking loss surface for a single epoch of training (just after the grokking step), while figures \ref{fig:30_t_T} ($\vec{\delta}_t \propto \theta^* - \theta_t$), \ref{fig:30_t_t+1} ($\vec{\delta}_t \propto \theta_{t+1} - \theta_t$), \ref{fig:t0_30} ($\vec{\delta}_t \propto \theta_0 - \theta_t$) and \ref{fig:rand_30} (random $\vec{\delta}_t$) show it for different training epochs.

\subsection{Loss surface curvature}
\label{sec:curvature}
The approach described in the previous section allows us to visualize the loss surface under a dramatic dimensionality reduction, and we need to be careful how we interpret the plots. One way to measure the level of convexity in a loss function is to compute the principal curvatures, which are simply eigenvalues of the Hessian. Since we are in a very high dimension, it would be impractical to compute all the eigenvalues of the Hessian, or simply to compute the Hessian itself. To avoid this dimensional problem, we estimated the curvature through the condition number of the Hessian of $L(\theta)$ at each step $t$, which is simply the ratio $\lambda_{min}(t) / \lambda_{max}(t)$, where $\lambda_{min}(t)$ and $\lambda_{max}(t)$ are, respectively, the minimum and maximum eigenvalues of the Hessian of $L_t$. 
The condition number is a direct measure of pathological curvature. Larger condition numbers imply slower convergence of gradient descent. \citet{DBLP:conf/nips/Li0TSG18} compute the above ratio 
at each point of the loss surface and observe that the convex-looking regions in the surface plots correspond to regions with insignificant negative eigenvalues, while chaotic regions contain large negative curvatures; and that for convex-looking surfaces, the negative eigenvalues remain extremely small.  

\begin{figure}[htp]
\vskip 0.2in
\begin{center}
\centerline{
\includegraphics[width=1.\linewidth]{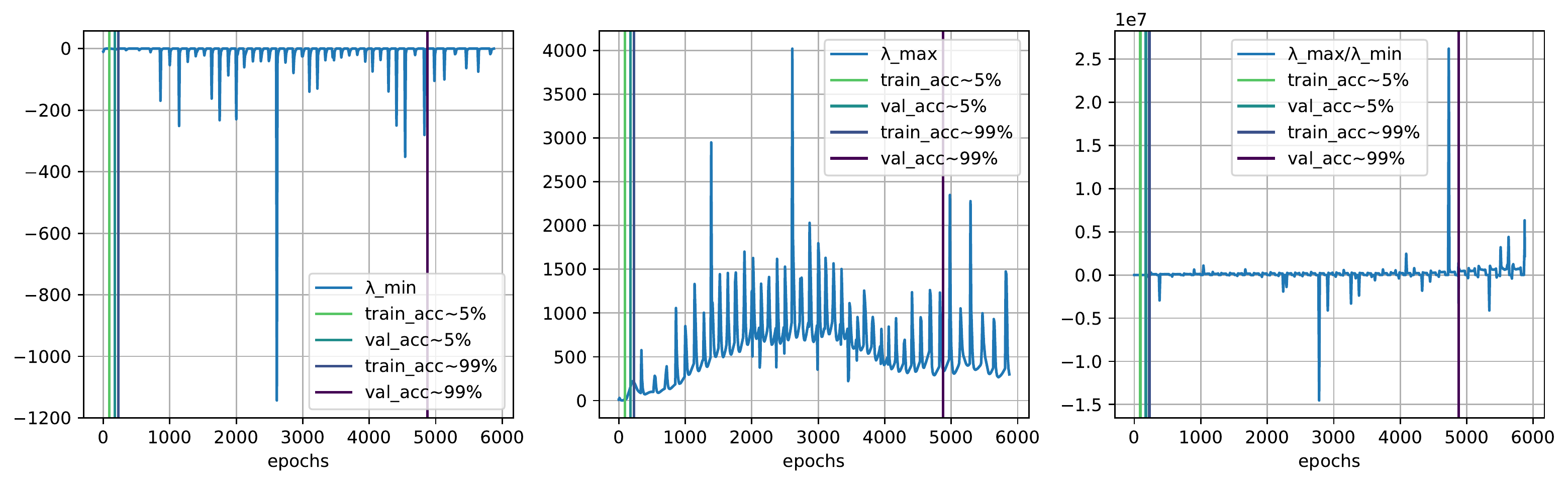}
}
\centerline{
\includegraphics[width=1.\linewidth]{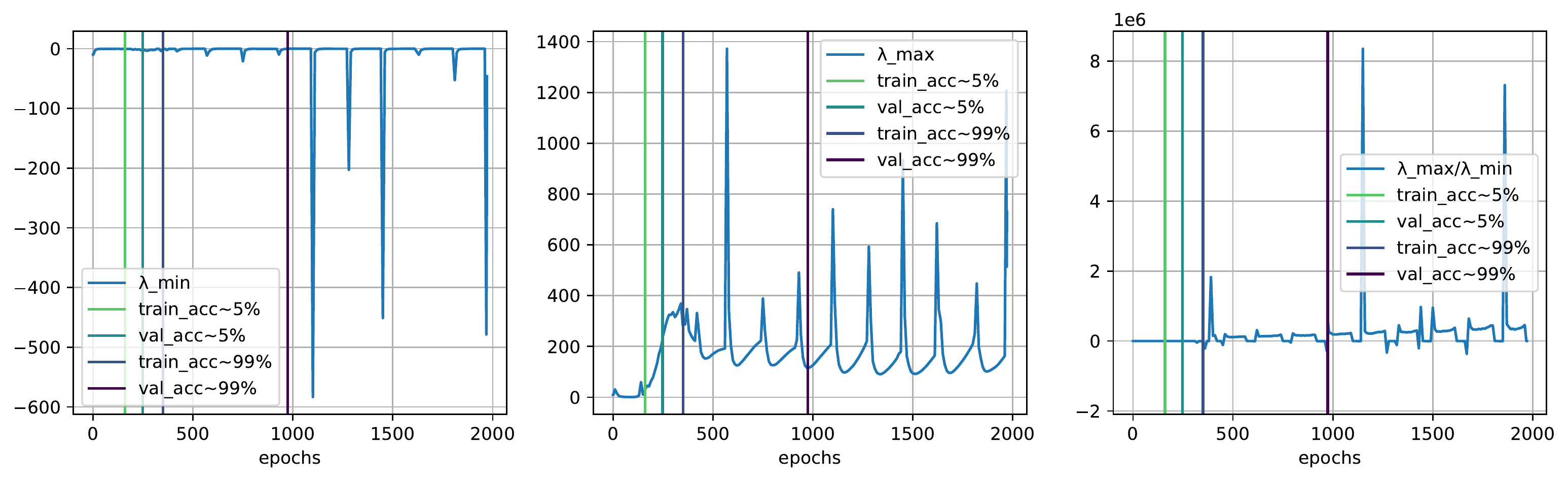}
}
\caption{
Condition number of the Hessian over training epoch for modular addition (top, $r=0.3$, bottom, $r=0.8$). 
We can see that there is no significant negative curvature in the trajectory (figure \ref{fig:hessian}). The curvature remains in general positive and is greatly disturbed at the slingshot points. 
}
\label{fig:hessian}
\end{center}
\vskip -0.2in
\end{figure}

To measure the level of curvature of the loss function, we compute the maximum (${\lambda}_{max}$) and minimum (${\lambda}_{min}$) eigenvalue of its Hessian (figure \ref{fig:hessian}). We observe that there is no significant negative curvature in the trajectory. The curvature remains generally positive and is greatly disturbed at the slingshot points (${\lambda}_{min}$ remains in general close to 0 while ${\lambda}_{max}$ is large, but during slingshot ${\lambda}_{min}$ becomes negative). Our loss is ${\lambda}_{max}$-smooth, i.e. its gradient is ${\lambda}_{max}$-Lipschitz. It is known from optimization literature that a function with bounded Hessian eigenvalues has a gradient that tends to decay when the parameter gets closer to the minimum, in contrast to a non-smooth one that generally has abrupt bends at the minimum, which causes significant oscillations for gradient descent \citep{bubeck2015convex}.

In fact, Let $\{\lambda_t(i)\}_i$ be the eigenvalues of $\mathcal{H}_t$ and $\{v_t(i)\}_i$ the associated eigenvectors. For a very small step size $\epsilon_t$ of SGD, $L_{t+1} - L_t \approx - \epsilon_t \| G_t \|^2 +  \frac{1}{2} \epsilon_t^2 G_t^T \mathcal{H}_t G_t - o(\epsilon_t^2 \|G_t\|^2)$. Futher, if $\lambda_t(i)  > 2/\epsilon_t$, we get $- \epsilon_t \| G_t \|^2 +  \frac{1}{2} \epsilon_t^2 G_t^T \mathcal{H}_t G_t > 0$ \footnote{This comes come from the fact that near $\theta_t$, $L(\theta) \approx L_t + (\theta - \theta_t)^T G_t +  \frac{1}{2} (\theta - \theta_t)^T \mathcal{H}_t (\theta - \theta_t) + o(\|\theta - \theta_t\|^2)$. Taking $\theta = \theta_t - \epsilon_t G_t$ gives the first approximation. The last inequality is obtained with $G_t^T \mathcal{H}_t G_t = \sum_{i} \lambda_t(i)
\langle G_t, v_t(i) \rangle^2$ and $\sum_{i} \langle G_t, v_t(i) \rangle ^2  = \| G_t \|^2$.
}.
When $\mathcal{H}_t$ has some large positive eigenvalues (i.e., high-curvature directions) and some eigenvalues close to 0 (i.e., low-curvature directions), gradient descent bounces back and forth in high-curvature directions and makes slow progress in low-curvature directions. In this case, the optimization problem has an ill-conditioned curvature. Furthermore, if the loss function near $\theta_t$ has a high condition number, that is, very small steps cause an increase in the cost function (for example, if $\theta_t$ is a very sharp minimum surrounded by high loss regions), 
the optimization problem becomes also ill-conditioned. During training, 
if the gradient norm does not shrink but $G_t^T \mathcal{H}_t G_t$ increases in order of magnitude, learning can become very slow despite a strong gradient. 
The above observation ($\lambda_t(i) > 2\epsilon_t^{-1}$)
is similar to what \citet{herrmann2022chaotic} [Theorem 2.1] defines for the eigenvalues of a positive-definite $\mathcal{H}_t$, as a locally chaotic training behaviour of the Local Lyapunov Exponents. They show evidence that neural network training is intrinsically locally chaotic due to the negative eigenspectrum of the Hessian, and that network training with SGD exhibits globally edge-chaotic behaviour. This observation is also linked to the \textit{progressive sharpening} phenomenon \citep{cohen2021gradient} in which $max_{i} \lambda_t(i)$ increases and reaches a value that is equal to or slightly larger than $2\epsilon_t^{-1}$, leading the model to enter an Edge of Stability regime where loss shows non-monotonic training behaviour over short time spans \citep{Thilak2022TheSM}. 

\subsection{Other landscape measures}

\begin{figure}[h]
\hfill
\foreach \tdp in {
30, 35, 40, 45, 50, 55, 60, 65, 70, 75, 80, 
90}{ 
\subfigure[$ r = \tdp \%$]{\includegraphics[width=.232\linewidth]{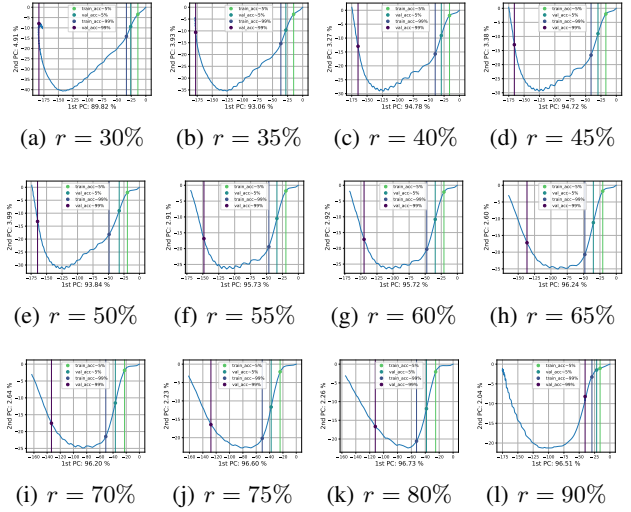}}
\hfill
}
\caption{Projection along the axis from initialization to solution, $\alpha(t)$ and $\beta(t)$. We can see that more than 98\% of the total variance in the parameter space occurs in the first 2 PCA modes much smaller than the total number of weights, suggesting that the optimization dynamics are embedded in a low-dimensional space.}
\label{fig:PCAs}
\end{figure}

To see how far optimization strays from the primary linear subspace, we plot the norm of the residual of the parameter value after projecting the parameters at each epoch into the 1-D subspace using the following two approaches. Let $M = [\theta_t - \theta_T]_{1 \le t \le T - 1} \in \mathbb{R}^{(T-1) \times d}$ where $T$ is the total number of training steps. We applied PCA to the matrix $M$, select the $2$ most explanatory directions, then project each parameter $\theta_t$ on this these two directions to have $\alpha(t)$ and $\beta(t)$.
On figure \ref{fig:PCAs}, we plot the projection along the 2 first PCA axis from initialization to solution, $\alpha(t)$ and $\beta(t)$. More than 98\% of the total variance in the parameter space occurs in the first 2 PCA modes much smaller than the total number of weights, suggesting that the optimization dynamics are embedded in a low-dimensional space \citep{DBLP:conf/nips/Li0TSG18,doi:10.1073/pnas.2015617118}.

\begin{figure}[tbh]
\hfill
\subfigure[$r=0.3$]{\includegraphics[width=.49\linewidth]{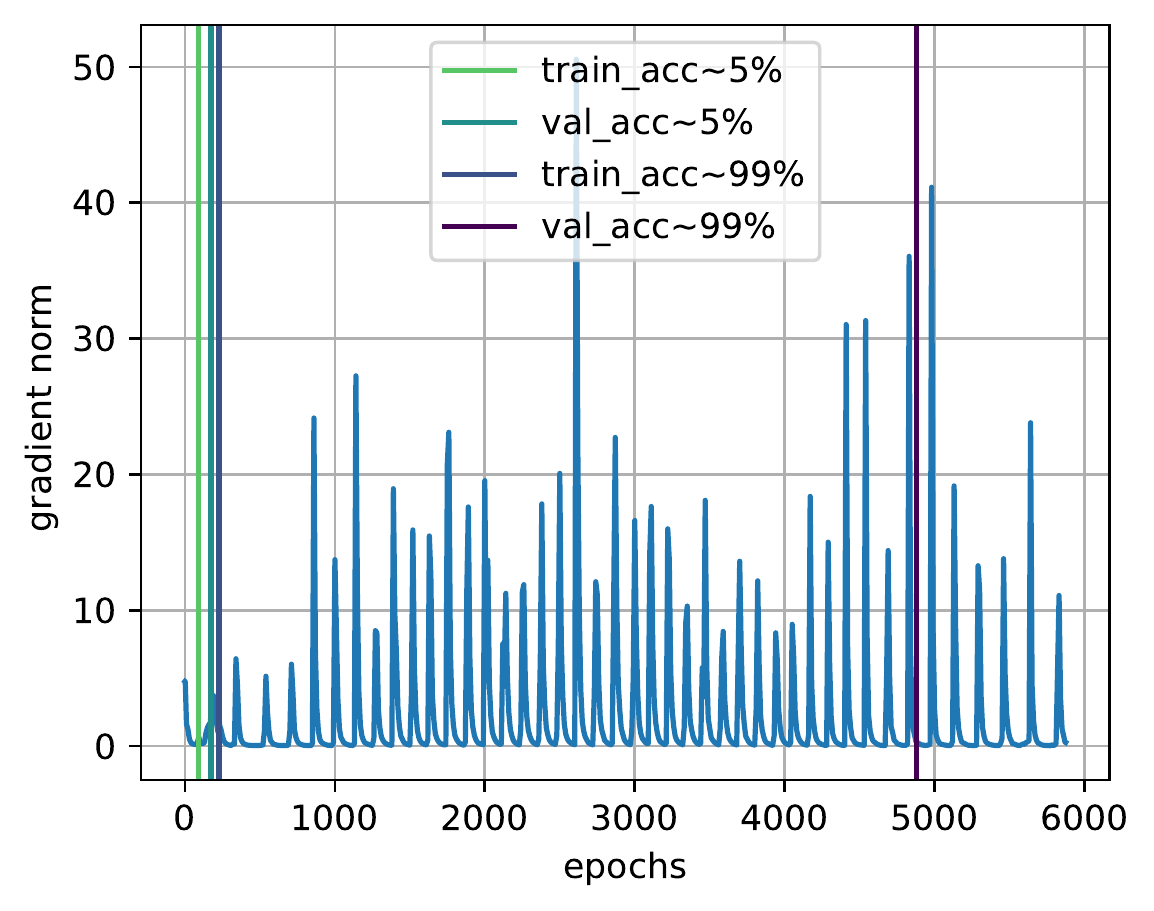}}
\hfill
\subfigure[$r=0.8$]{\includegraphics[width=.49\linewidth]{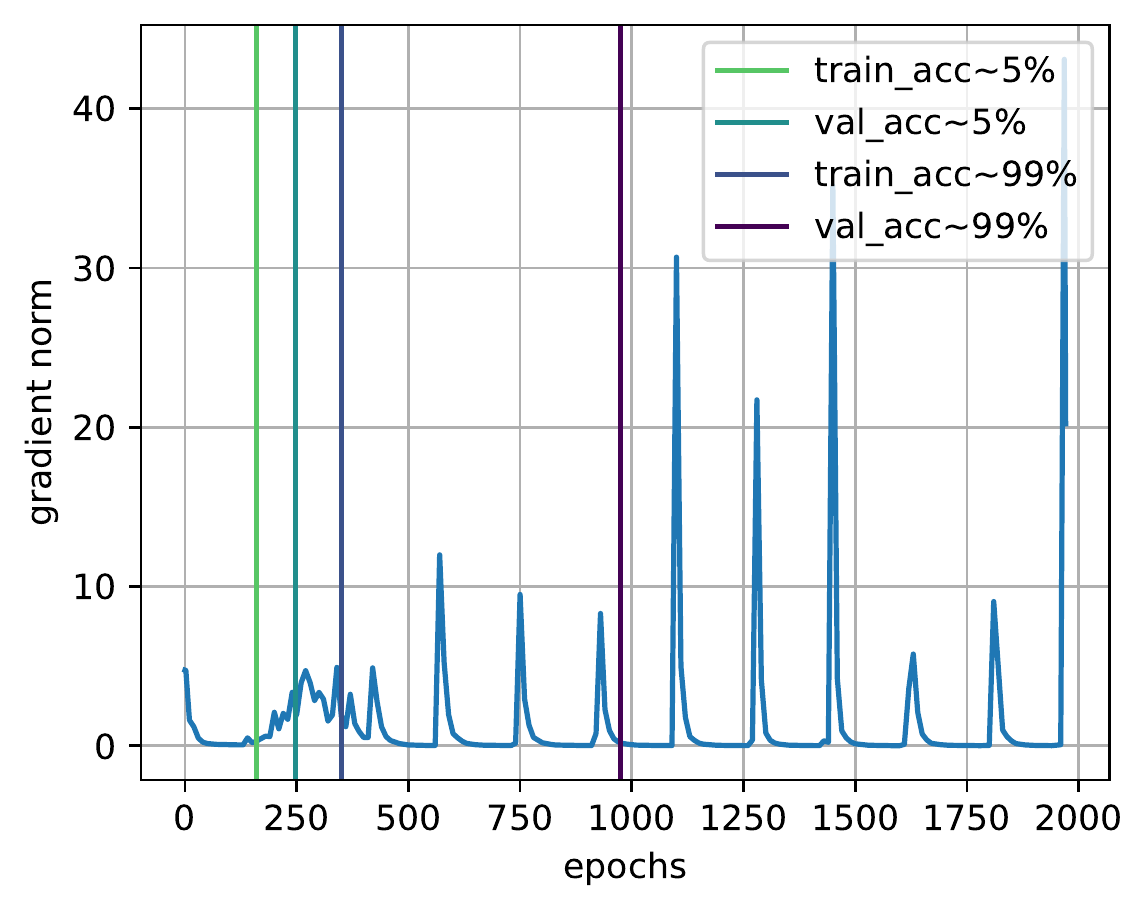}}
\hfill
\vspace{-0.02in}
\caption{Gradient Norm $\| G_t \|^2$. Most of the time there is no significant gradient, which results in a weak progression of gradient descent. 
}
\label{fig:grad_norm}
\end{figure}

\begin{figure}[tbh]
\centering
\includegraphics[width=1.\linewidth]{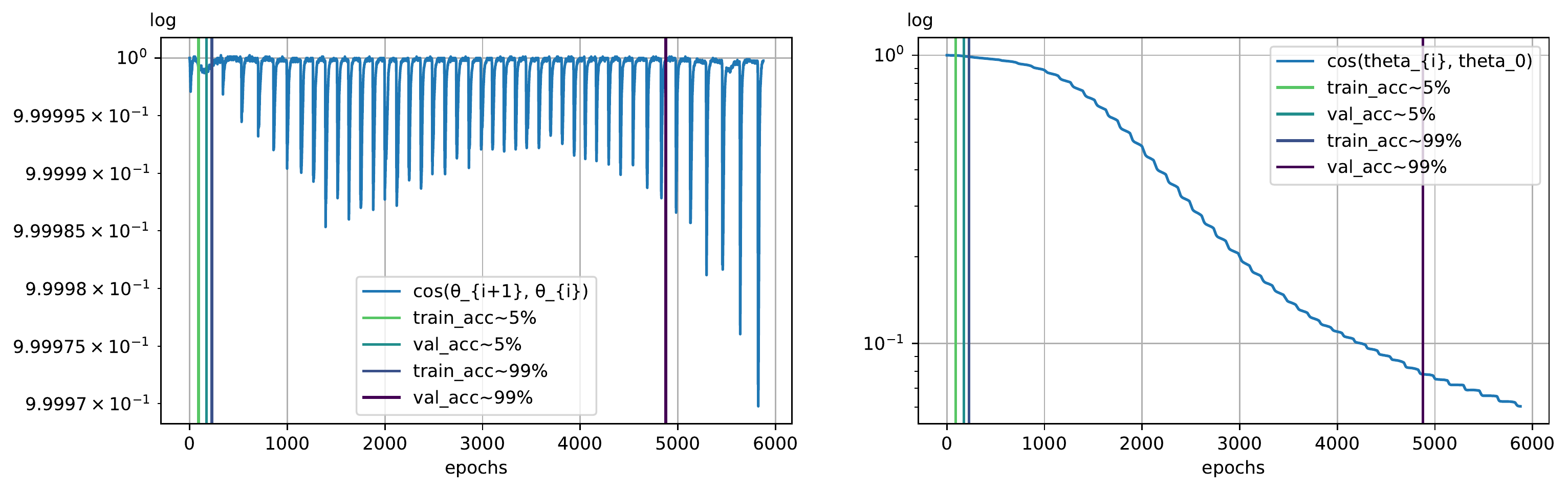}
\includegraphics[width=1.\linewidth]{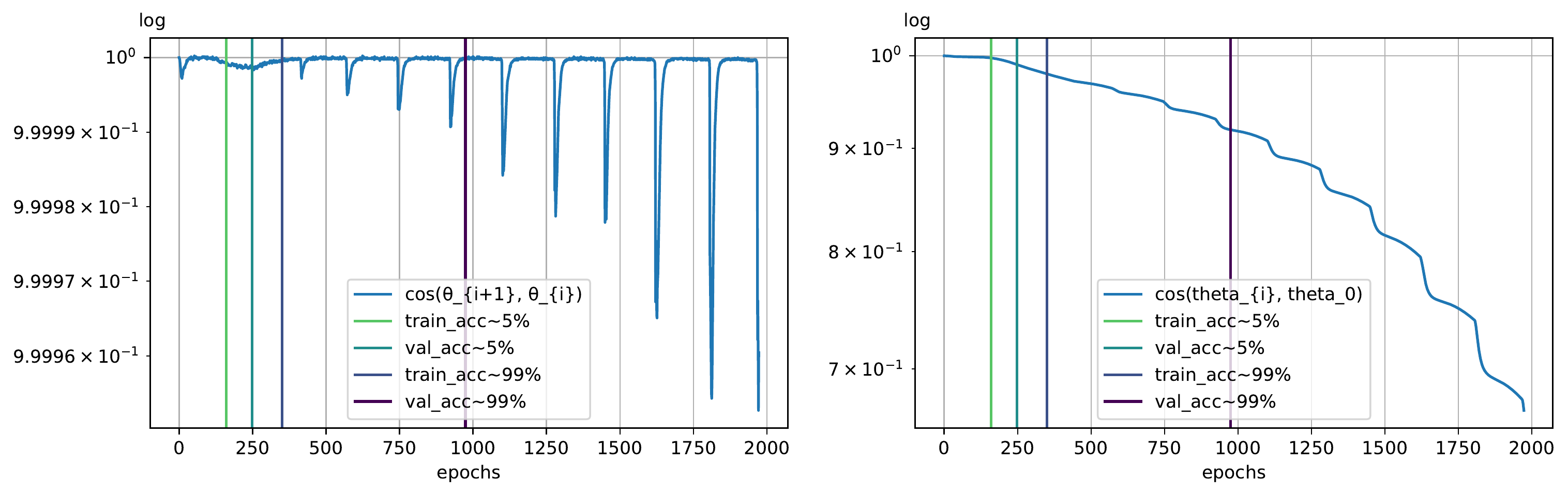}
\caption{Relative consine similarity ($r=0.3$ and $0.8$). Left) $cos(\theta_t$, $\theta_{t+1})$ Right) $cos(\theta_t$, $\theta_{0})$. Most of the time the weights progress very little, jumping when the gradient explodes, with a slingshot effect.
}
\label{fig:angles}
\end{figure}

Also, the cosine similarity measure between the model weights from one epoch to the next remains almost constant, except at the slingshot location (figure \ref{fig:angles}). This allows us to see that our model passes through the anomalies observed above. It has been shown that with high probability over the initialization; the iterates of the gradient descent algorithm even stay in a small fixed neighbourhood of the initialization during training. Because the parameters only move very little, this type of training has also been coined lazy training \citep{chizat2019lazy, berner20modern}. These steps where the distances change abruptly are caused by spikes in the training gradient (figure \ref{fig:grad_norm}).

\subsection{The Slingshot mechanism}

Above, we derive $\dot{L}(t) = - \| G(t) \|^2$ for a small enough step size. This relaxation property of gradient descent reflects the fact that the loss function cannot increase. However, we lose this property with certain accelerated or adaptative methods like Adam, or on ill-conditioned problems as we saw above with $L_{t+1} > L_t$ for $\epsilon_t \lambda_t(i) > 2$. In the case of grokking, the loss, even when it becomes zero, presents a sudden growth, followed by decay, and this is in a periodic way. 
\citet{Thilak2022TheSM} have recently shown that this phenomenon is general to the optimization of deep neural networks. Between two slingshots, the gradient is almost zero, all eigenvalues of the Hessian are nonnegative, and there is one direction of very large curvature that dominates the others and the weights update, resulting in a small change in weight from one training step to another. The model thus seems to traverse a flat valley.

To reduce this phenomenon (i.e. the amplitude of the spikes), we clipped the gradient norm during training (using a threshold $\eta > 0$).  This had the effect of slowing down generalization, but not preventing it.

$$
\dot{\theta} = \left\{
    \begin{array}{ll}
        - \frac{\eta}{\| G(t) \|}     G(t) & \mbox{if } \| G(t) \| \ge \eta \\
        - G(t) & \mbox{otherwise.}
    \end{array}
\right. (\eta > 0)
$$

$$
\Longrightarrow \dot{L} = \left\{
    \begin{array}{ll}
         - \eta \| G(t) \| & \mbox{if } \| G(t) \| \ge \eta \\
        - \| G(t) \|^2 & \mbox{otherwise.}
    \end{array}
\right. (\eta > 0)
$$

The spikes are reduced as $\eta \rightarrow 0$, but remain visible as $\theta \rightarrow \theta^*$.  By reducing the learning rate extremely, the spikes also become less visible, but we also pay the same cost because the model takes more steps to generalize.

\subsection{Three-dimensional visualizations}

\begin{figure}[H]
\FPeval{\width}{clip(0.45)}
\hfill
\includegraphics[width=\width\linewidth]{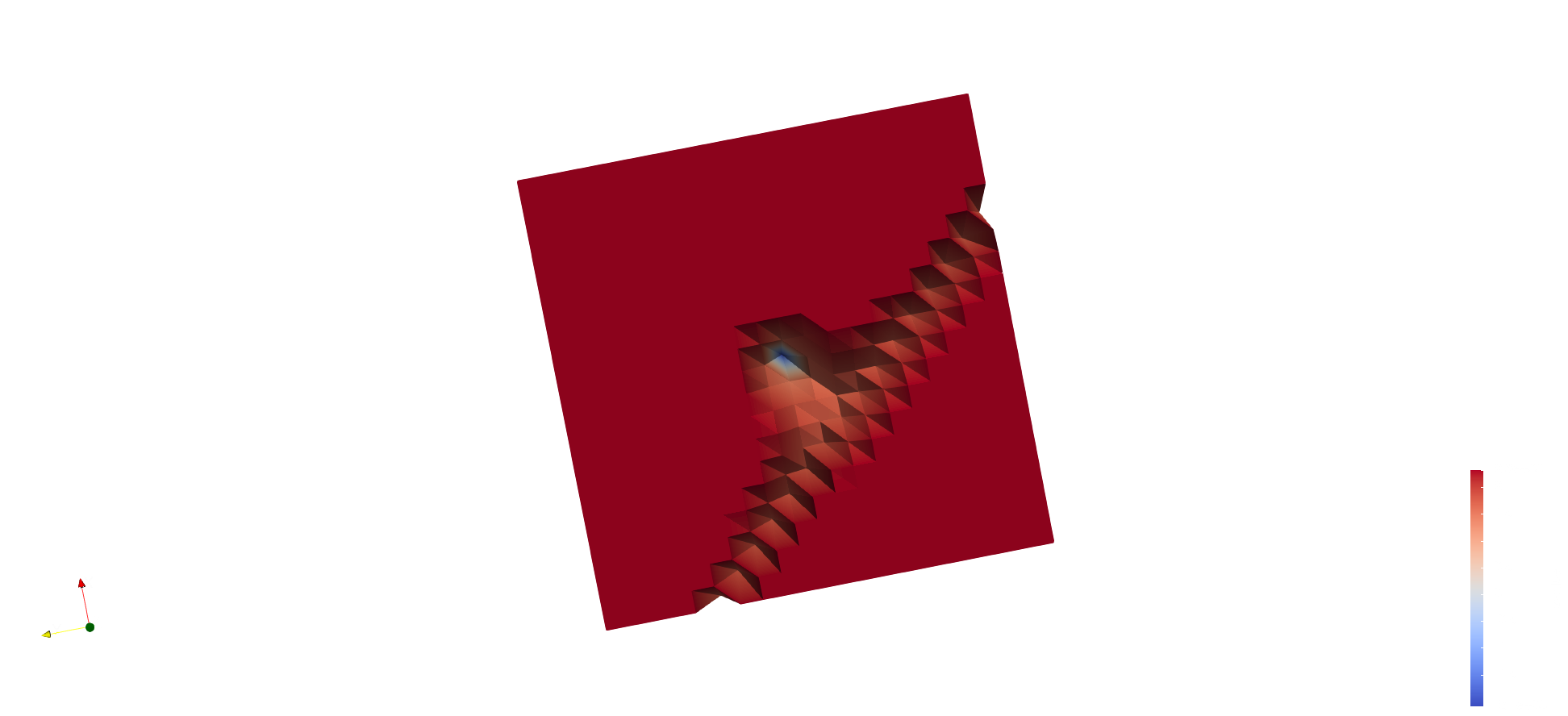}
\includegraphics[width=\width\linewidth]{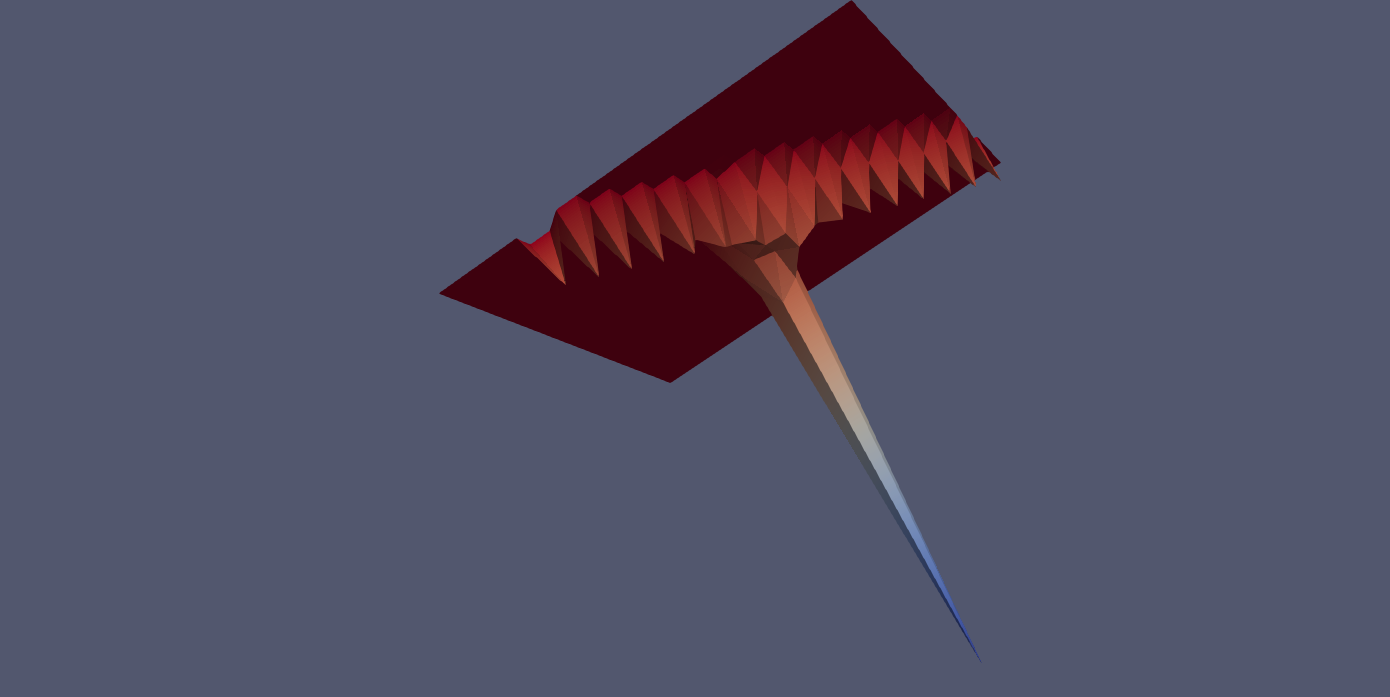}
\hfill
\caption{
3D ($r=0.5$)
}
\label{fig:3D}
\end{figure}

We can accomplish this by viewing a heatmap of the cost function in 2D (figure \ref{fig:3D}).

\onecolumn

\begin{figure}[htp]
\vskip 0.2in
\begin{center}
\foreach \tdp in {
90, 85, 80, 75, 70, 65, 60, 55, 50, 
40, 35, 
25}{ 
\subfigure
[$r=0.\tdp$]
{
\includegraphics[width=.30\linewidth]{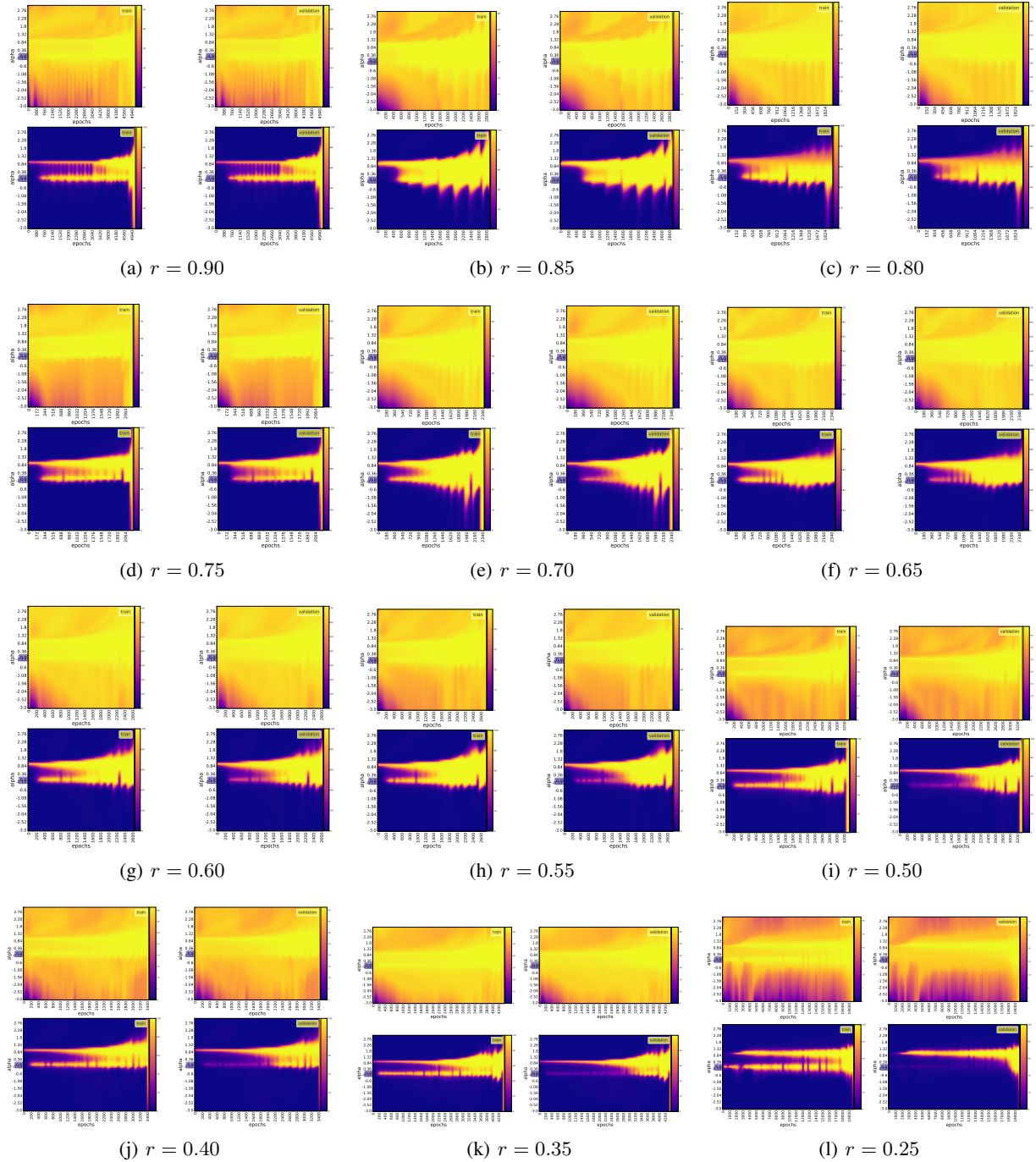}
}
}
\caption{
1D projection of the grokking loss (top) and accuracy (bottom) surface for different value of the training data fraction $r$, for \textbf{modular addition}. The direction used for each training epoch $t$ is $\vec{\delta}_t = \theta^* - \theta_t$.}
\label{fig:+_t_T_all}
\end{center}
\vskip -0.2in
\end{figure}

\begin{figure}[htp]
\vskip 0.2in
\begin{center}
\foreach \tdp in {
90, 85, 80, 75, 70, 65, 60, 55, 50, 
45, 
40, 35}{ 
\subfigure
[$r=0.\tdp$]
{
\includegraphics[width=.30\linewidth]{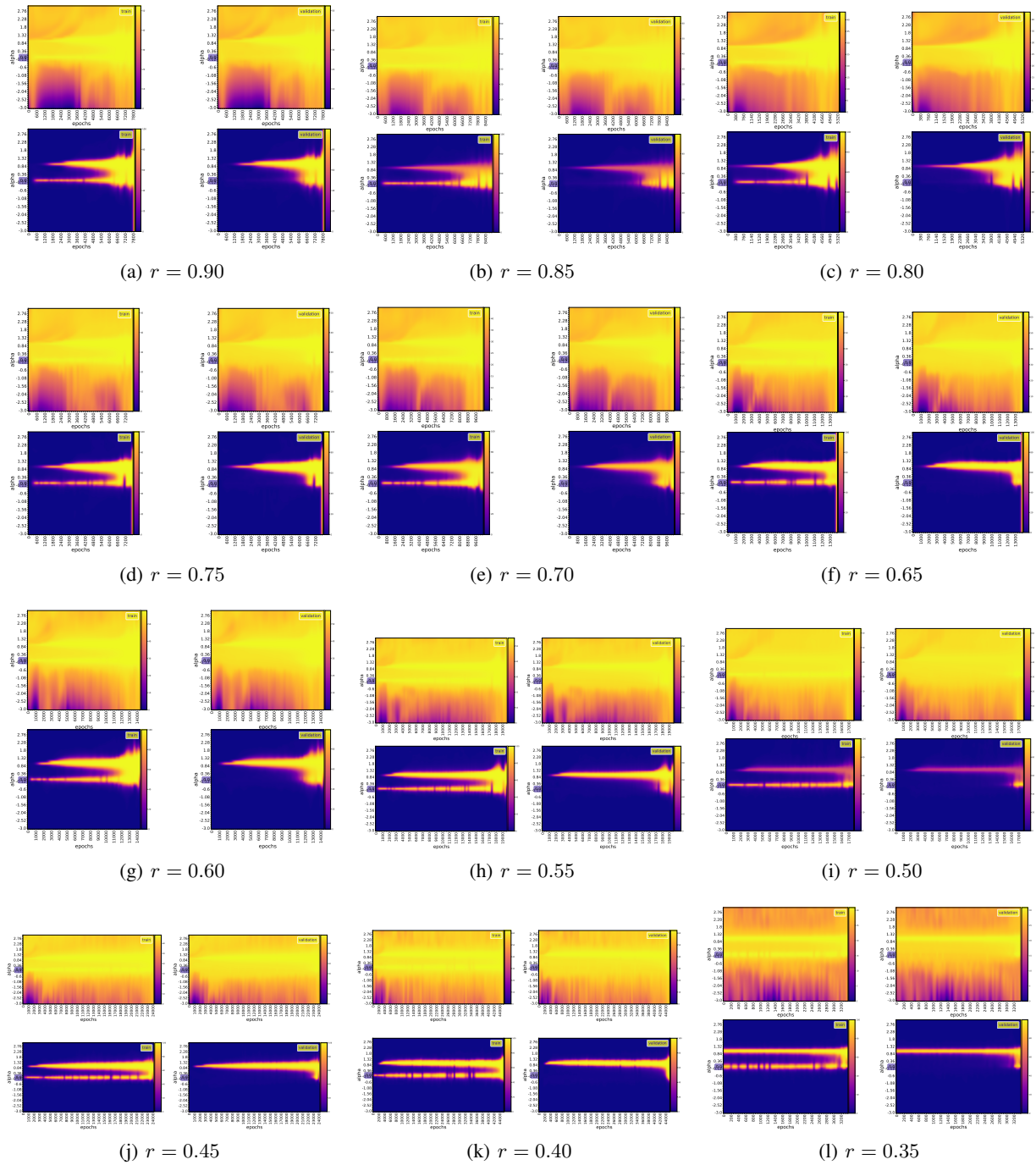}
}
}
\caption{
1D projection of the grokking loss (top) and accuracy (bottom) surface for different value of the training data fraction $r$, for \textbf{multiplication in $S_5$}. The direction used for each training epoch $t$ is $\vec{\delta}_t = \theta^* - \theta_t$.
}
\label{fig:s5_t_T_all}
\end{center}
\vskip -0.2in
\end{figure}

\begin{figure}[htp]
\vskip 0.2in
\begin{center}
\foreach \tdp in {
25, 
35, 40, 45, 50, 55, 60,  65, 70, 75, 
85, 90}{ 
\subfigure
[$r=0.\tdp$]
{
\includegraphics[width=.45\linewidth]{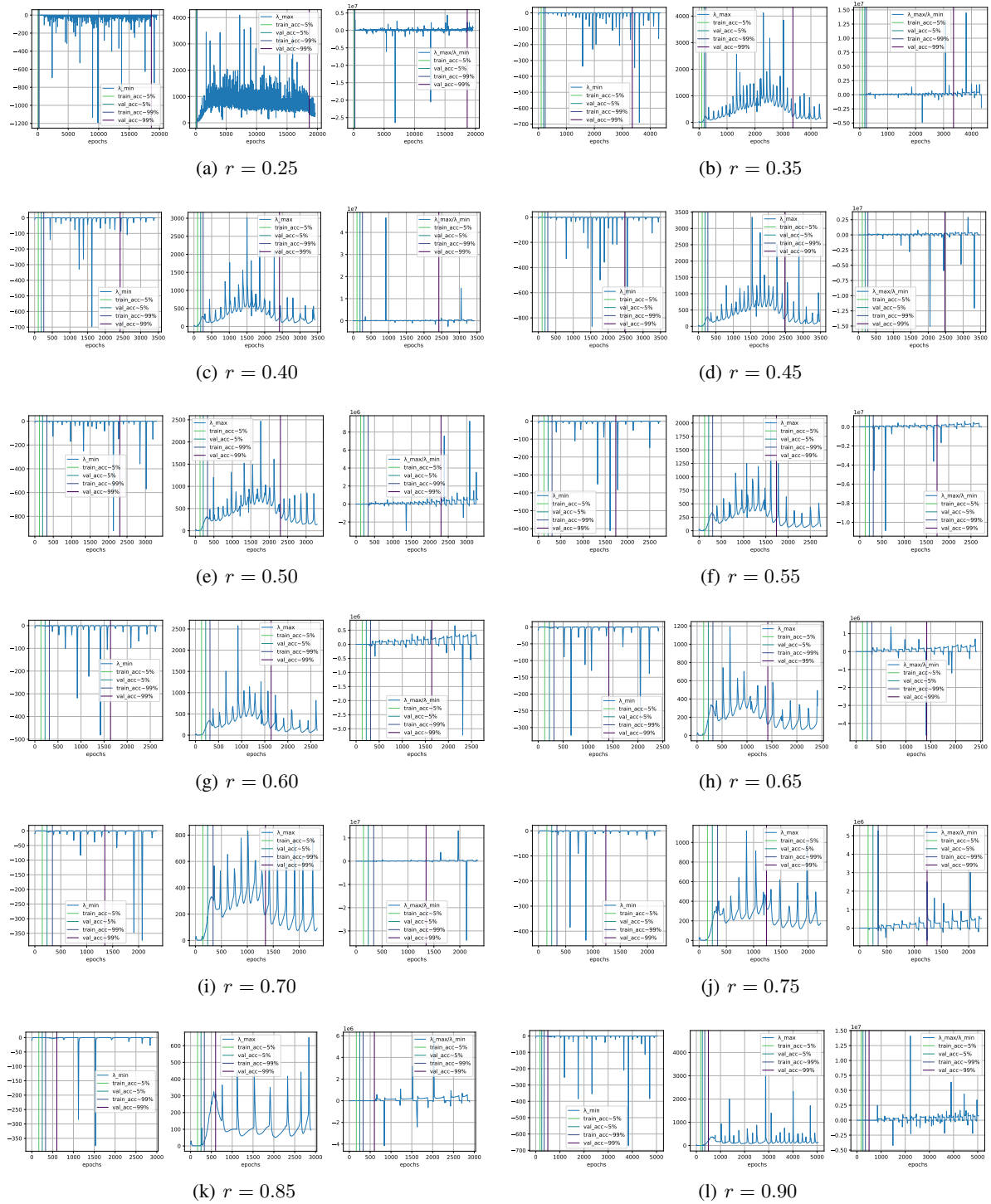}
}
}
\caption{
Condition number of the Hessian over training epoch for modular addition. 
}
\label{fig:hessian_all}
\end{center}
\vskip -0.2in
\end{figure}


\begin{figure}[htp]
\vskip 0.2in
\begin{center}
\foreach \tdp in {
25, 
35, 40, 45, 50, 55, 60,  65, 70, 75, 
85, 90}{ 
\subfigure
[$r=0.\tdp$]
{
\includegraphics[width=.14\linewidth]{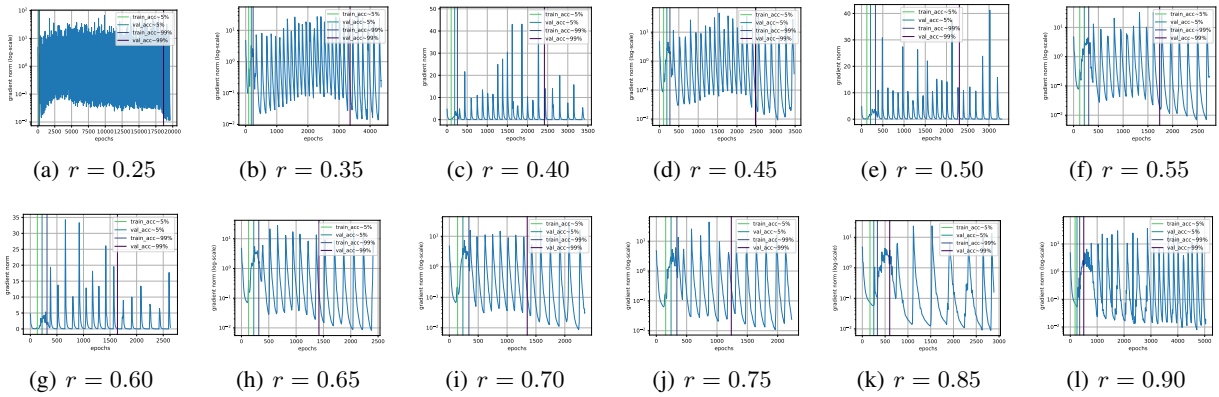}
}
}
\caption{
Gradient Norm for modular addition.
}
\label{fig:grad_norm_all}
\end{center}
\vskip -0.2in
\end{figure}


\begin{figure}[htp]
\vskip 0.2in
\begin{center}
\foreach \tdp in {
25, 
35, 40, 45, 50, 55, 60,  65, 70, 75, 
85, 90}{ 
\subfigure
[$r=0.\tdp$]
{
\includegraphics[width=.3\linewidth]{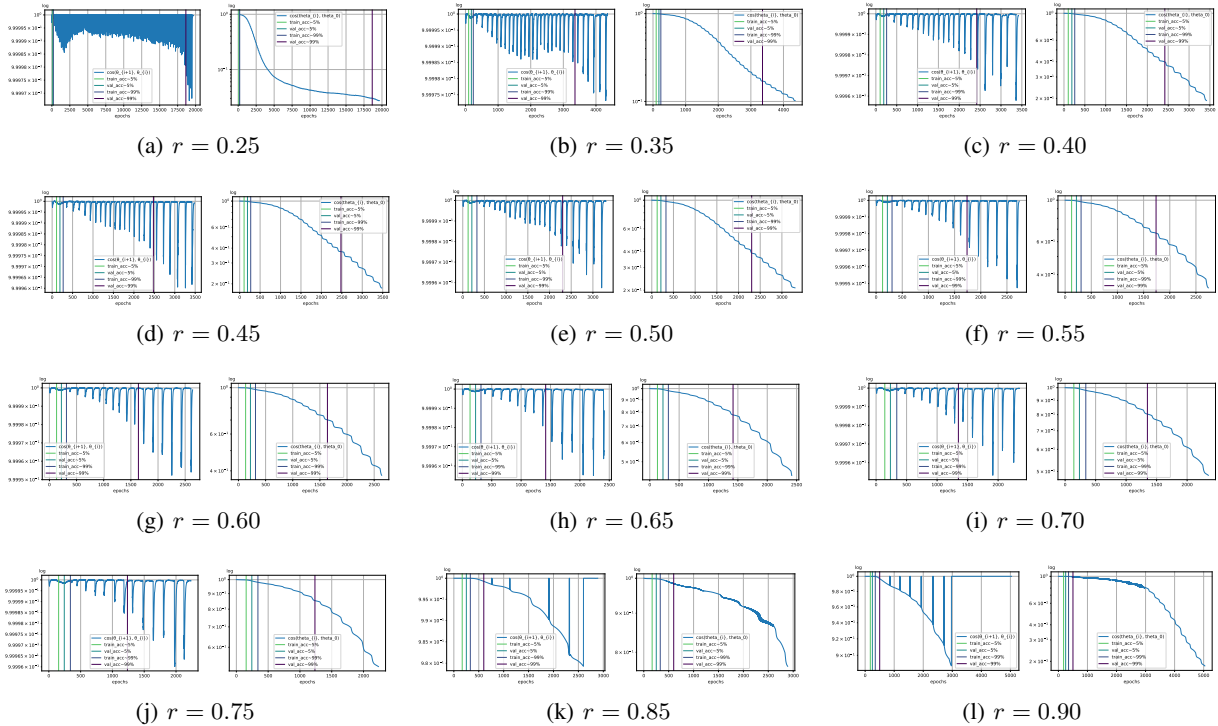}
}
}
\caption{
Relative cosine distance for modular addition. Left) $cos(\theta_t$, $\theta_{t+1})$ Right) $cos(\theta_t$, $\theta_{0})$
}
\label{fig:angles_all}
\end{center}
\vskip -0.2in
\end{figure}


\include{ID}

\include{theorical_models}

\end{document}

%% file: ID.tex
\section{Intrinsic Dimension Estimation}

We also empirically investigate the relationship between the evolution of the objective function of a neural network during training and the evolution of the intrinsic dimension (ID) of the activation manifolds of its layers.  We found that the dimensionality of the network layers (mainly the last layer) correlates with oscillations in training and validation performances (loss and accuracy) both in toy models \citep{Liu2022TowardsUG} and transformer \citep{power2022grokking}.
The intrinsic dimensionality grows at the very beginning of the training, then decreases (oscillating very strongly) until it stabilizes at a certain level of the training. 

\subsection{Methods}

Data are usually represented by high-dimensional feature vectors, but in many cases they could be in principle embedded in lower-dimensional spaces without any loss of information \citep{Facco_2017}. In other words, any set of points of a high-dimensional ambient space $\mathbb{R}^D$ usually actually lies on (or close to) another space of dimension $d$ smaller than $D$ : $d$ is then called the intrinsic dimension of the ambient space $\mathbb{R}^D$. Several methods have long been developed to estimate this intrinsic dimension \citep{doi:10.1126/science.290.5500.2323,doi:10.1126/science.290.5500.2319,NIPS2001_f106b7f9,doi:10.1073/pnas.1031596100,NIPS2004_74934548,David2005}, among which Two Nearest Neighbors (TWONN) \citep{Facco_2017} and Maximum Likelihood Estimation (MLE) approaches \citep{NIPS2004_74934548,David2005} that examine a neighborhood around each point and compute the Euclidean distance to the nearest neighbor $k^{th}$ \citep{pope2021intrinsic}. 

Assuming that the density is constant within small neighborhoods, the MLE \citep{NIPS2004_74934548}  uses a Poisson process to model the number of points found by random sampling within a given radius around each sample point \citep{pope2021intrinsic}. By relating the rate of this process to the surface area of the sphere, the likelihood equations yield an estimate of the intrinsic dimension at a given point $x$ as $$\hat{m}_k(x) = \bigg[ \frac{1}{k-1} \sum_{j=1}^{k-1} log \frac{T_k(x)}{T_j(x)} \bigg]^{-1}$$ where $T_j(x)$ is the Euclidean distance from $x$ to its $j^{th}$ nearest neighbor. \citet{NIPS2004_74934548} propose to average the local estimates at each point to obtain a global estimate ($n$ is the number of sample) :
$$\hat{m}_k = \frac{1}{n} \sum_{i=1}^{n} \hat{m}_k(x_i)$$

\citet{David2005} suggest a correction based on the averaging of the inverses : $$\hat{m}_k = \bigg[ \frac{1}{n} \sum_{i=1}^{n} \hat{m}_k(x_i)^{-1} \bigg]^{-1}$$

TWONN \citep{Facco_2017} uses just the two nearest neighbors of each point to make the estimate. The distribution of $R=\frac{\Delta v_2}{\Delta v_1}$ has as probability density function $g(R)=\frac{1}{(1+R)^2}$ \citep{Facco_2017}, where $\Delta v_l = \omega_d (r_l^d - r_{l-1}^d)$ is the volume of the hypersferical shell enclosed between two successive neighbors $l-1$ and $l$ of a given point $i$ of the dataset ($r_l$ being the distance between $i$ and its $l^{th}$ nearest neighbor), $d$ the dimensionality of the space in which the points are embedded and 
$\omega_d = \frac{\pi^{d/2}}{\Gamma(d/2 + 1)}$ 
the volume of the d-sphere with unitary radius ($\Gamma$ is the Euler gamma function). Let $\mu = \frac{r_2}{r_1} \ge 1$, then $R=\mu^d - 1$ since $r_0=0$, which allows to find an explicit formula for the distribution of $\mu$, $f(\mu) = d \mu^{-d-1} \mathbb{1}_{[1,+\infty]}(\mu)$, and thus its cumulative distribution $F(\mu) = (1-\mu^{-d}) \mathbb{1}_{[1,+\infty]}(\mu) \Rightarrow d = - \frac{log(1-F(\mu))}{log(\mu)}, \mu \ge 1$. This yields the algorithm \ref{alg:2nn_algo}. 

Note that the last step of this algorithm \ref{alg:2nn_algo} is to solve the linear equation $-\log(1-F(\mu_i)) = d*log(\mu_i)$ for any point $i$ of the data set, which by the least squares reduces to $d^* = argmin_{d \in \mathbb{R}} \sum_{i=1}^{n} \big( \log(1-F(\mu_i)) + d*log(\mu_i) \big)^2$, that is $d^* = - \frac{\sum_{i=1}^{n} log(\mu_i) \log(1-F(\mu_i))}{\sum_{i=1}^{n} log(\mu_i)^2} = -\frac{\sum_{i=1}^{n} log(1-i/n) * log(\mu_{\sigma(i)}) }{\sum_{i=1}^{n} log(\mu_i)^2}$. 
For $k=2$, we have $\hat{m}_{ki} = log(\mu_i)^{-1}$ for any point $i$ in the data set. This implies, if $-\log(1-F(\mu_i)) = d^* *log(\mu_i) \ \forall i$, $\hat{m}_k^{-1} d^* = \frac{1}{n} \sum_{i=1}^{n} \hat{m}_{ki}^{-1} d^* = - \frac{1}{n} \sum_{i=1}^{n} \log(1-i/n)$, that is, $\hat{m}_k \propto d^*$ up to the least squares error $ \sum_{i=1}^{n} \big( \log(1-F(\mu_i)) \big)^2 - (d^*)^2 \sum_{i=1}^{n} log(\mu_i)^2$. We empirically obtained a Pearson correlation coefficient of $\sim 99.60\%$ between MLE (with $k=2$) and TWONN (with a \textit{p-value} in the order of $\sim 10^{-7}$). Since we will only be interested in the evolution of the intrinsic dimension (and not its value itself), we could use any of the two methods in our work.

\begin{algorithm}[tb]
   \caption{TWONN estimator for intrinsic dimension}
   \label{alg:2nn_algo}
\begin{algorithmic}
   \STATE 1. Compute the pairwise distances for each point in the dataset $i = 1, \dots, n$
   \STATE 2. For each point $i$ find the two shortest distances $r_{i} (i) $ and $r_{i}(2)$.
   \STATE 3. For each point $i$ compute $\mu_i = \frac{r_{i} (2)}{r_{i}(1)}$
   \STATE 4. Compute the empirical cumulate $F(\mu_i)$ by sorting the values of $\{\mu_i\}_{i=1}^{n}$ in an ascending order through a permutation $\sigma$, then define $F(\mu_{\sigma(i)}) = \frac{i}{n}$
   \STATE 5. Fit the points of the plane given by coordinates $\{(log(\mu_i), -log(1-F(\mu_i))) \ | \ i=1, \dots, n\}$ with a straight line passing through the origin.
\end{algorithmic}
\end{algorithm}

\subsection{Results}

We use \citet{David2005}'s Maximum Likelihood Estimation approach with $k=2$ neighbours to estimate the intrinsic dimensionality of every layer of our model during training, on training data and validation data.  For modular addition, we varied the percentage of training data from 0.35 to 0.9 in steps of 0.5, that is, in $\{0.35, 0.40, \dots, 0.85, 0.90 \}$. We have repeated each experiment for each dataset size with 2/3/5 random seeds. We fixed the learning rate to $10^{-4}$. See figures \ref{fig:IDs_35} and \ref{fig:IDs_35_2}  below. 


\begin{figure}[h]
\centering
\includegraphics[scale=0.35]{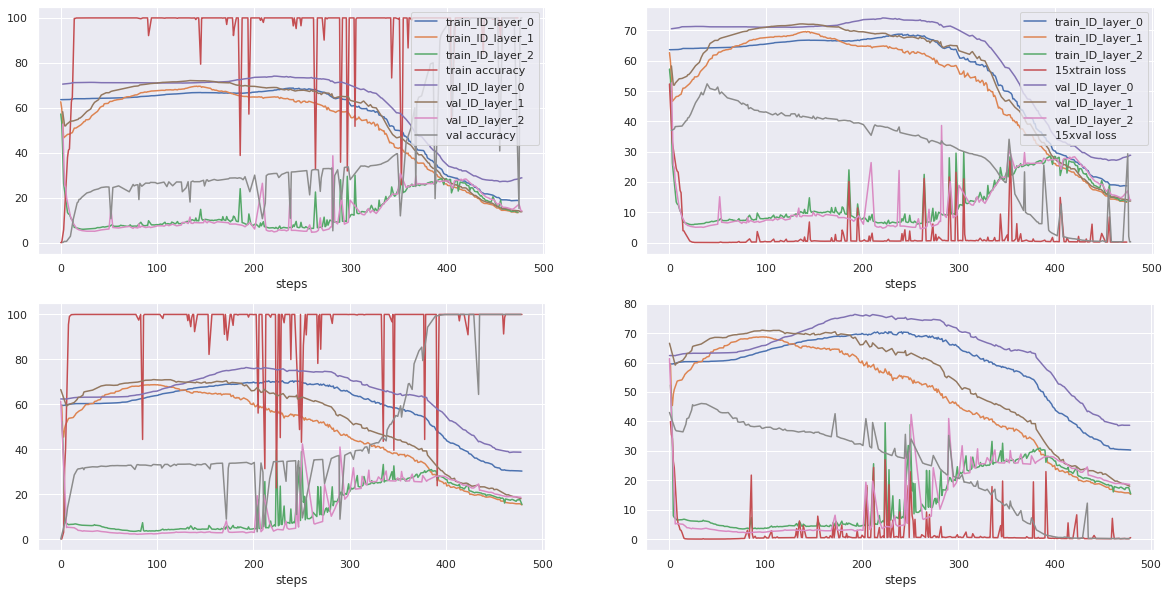}
\caption{Modular addition,
2 different initial conditions (one per row).
Steps is training steps $\times 100$.
}
\label{fig:IDs_35}
\end{figure}
    
\begin{figure}[h]
\centering
\hfill
\subfigure[
]{
\includegraphics[width=.49\linewidth]{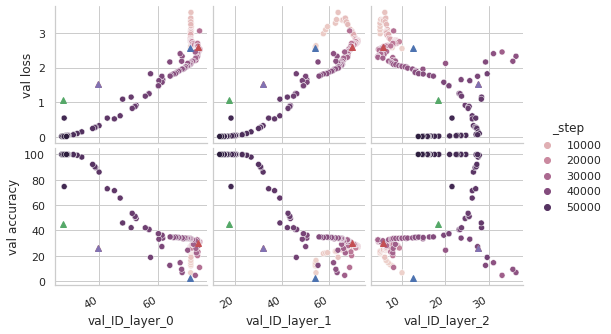}
\includegraphics[width=.49\linewidth]{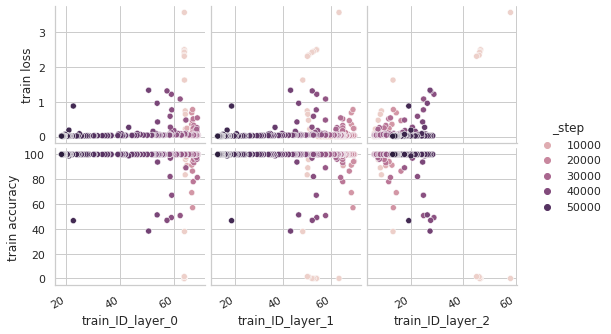}
}
\hfill
\subfigure[
]{
\includegraphics[width=.49\linewidth]{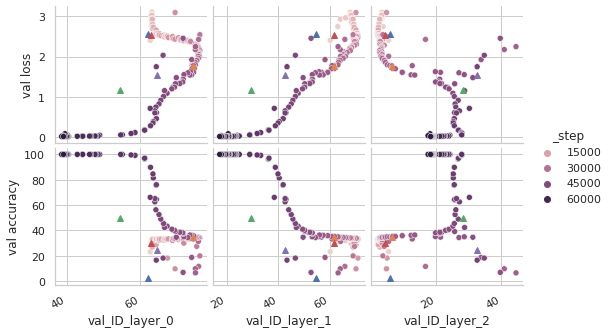}
\includegraphics[width=.49\linewidth]{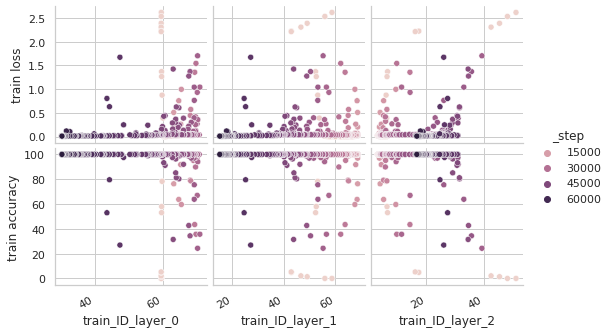}
}
\hfill
\caption{Training \& validation accuracy \& loss versus estimated intrinsic dimension in the case of modular addition, 
$2$ different initial conditions (one per column)}
\label{fig:IDs_35_2}
\end{figure}

\begin{figure}[h]
\centering
\hfill
\subfigure[
Accuracy
]{
\includegraphics[width=0.45\linewidth]{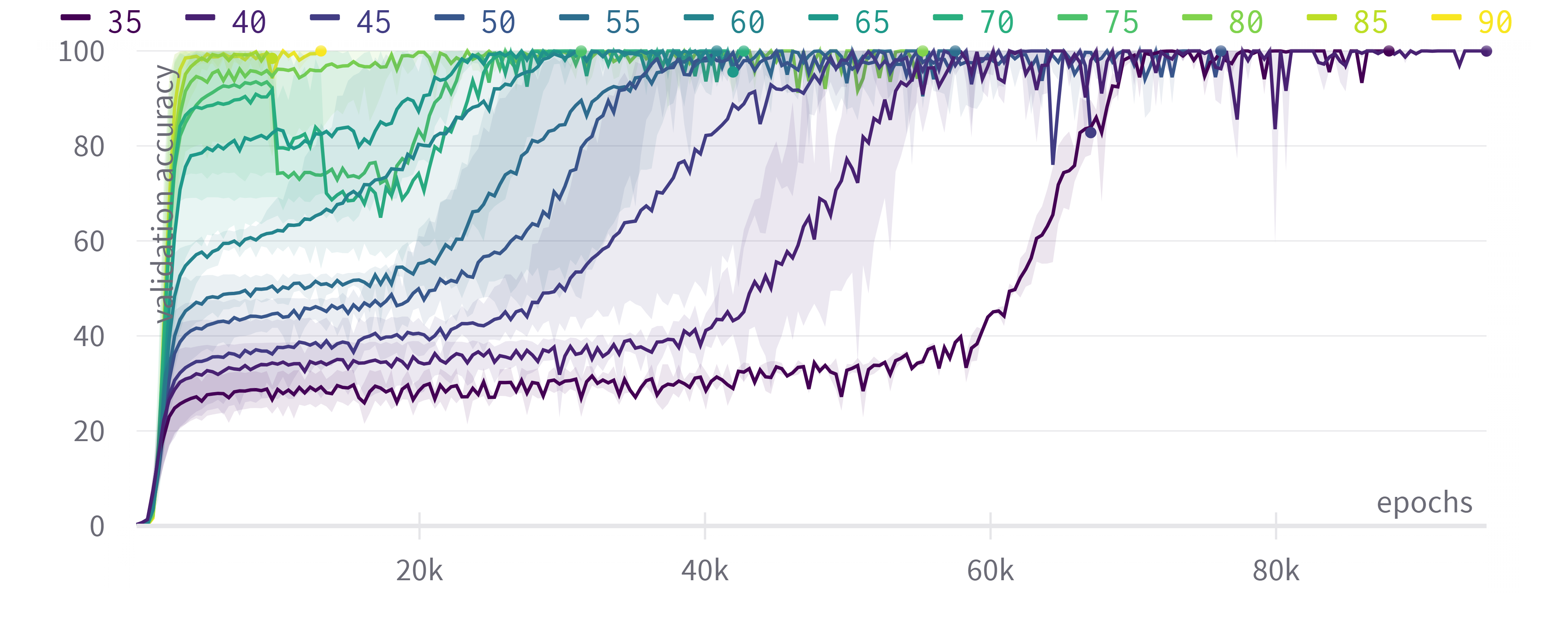}
}
\hfill
\subfigure[
Loss
]{
\includegraphics[width=0.45\linewidth]{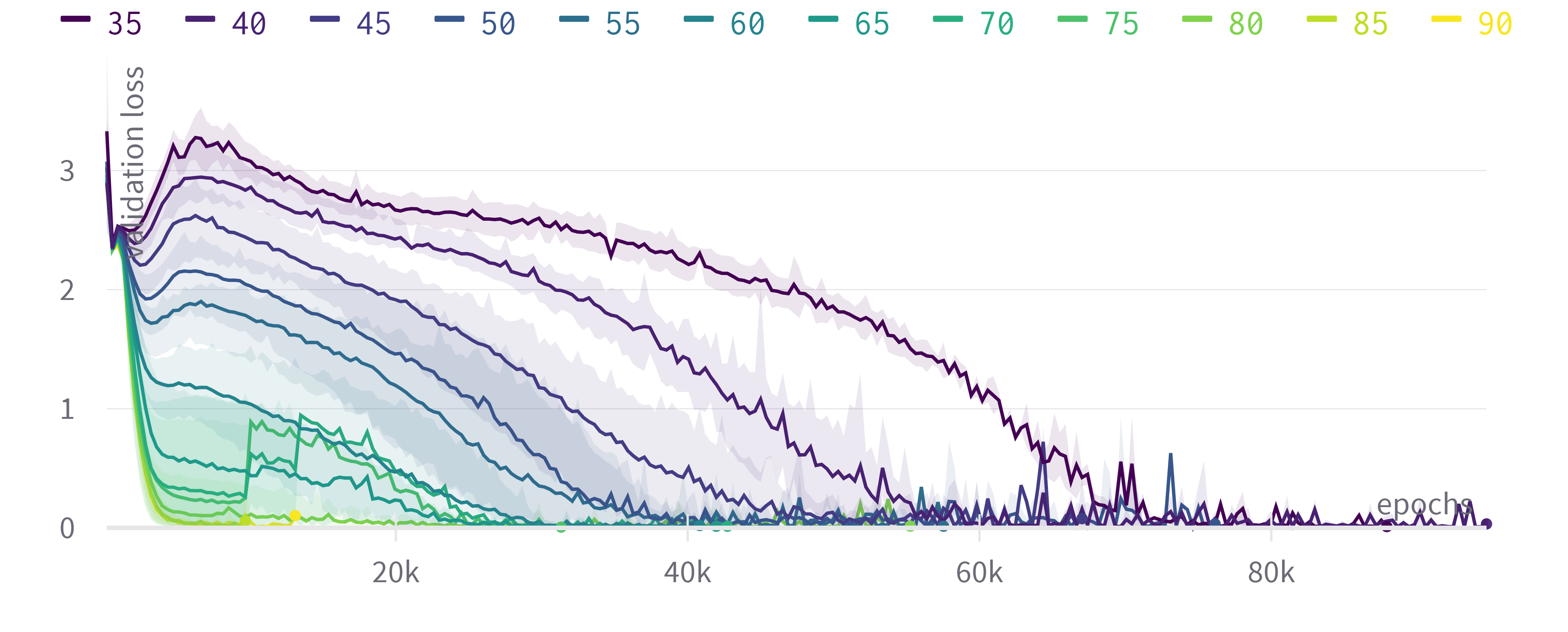}
}
\caption{(Grokking) Validation accuracy and loss for modular addition, with different datasize. 
For other operators, the validation accuracy stays at 0 before starting to grow. For addition, it goes up to the percentage of training data at the beginning of training, stays there, before grokking later.}
\label{fig:datasizeVSgrok2}
\end{figure}

%% file: theorical_models.tex
\section{Theoretical models of grokking} 
\label{sec:theo_grok}

We will use the following acronyms to designate our algorithms :
sgd for vanilla SGD, momentum for SGD with momentum  \citep{POLYAK19641}, rmsprop for the RMSProp algorithm \citep{HintonG2012}, rprop for the resilient backpropagation algorithm \citep{298623}, adam for Adam \citep{kingma2014adam} and adamax for Adamax \citep{kingma2014adam}.

\subsection{Rosenbrock function}
\label{sec:rosenbrock}

The vanilla rosenbrok function is given by $g_n(x) = \sum_{i=1}^{n/2} \big[ 100 (x_{2i} - x_{2i-1}^2)^2 + (x_{2i-1} - 1)^2 \big]$, with the gradient $\nabla_i g_n(x) =200 (x_i - x_{i-1}^2) \cdot \mathbb{1}_{i \in 2 \mathbb{N}} - \big[ 400 x_i (x_{i+1} - x_i^2) - 2(x_i - 1) \big] \cdot \mathbb{1}_{i \in  2 \mathbb{N} - 1}$, and $x^* \in \{ (1, \dots, 1), (-1, 1, \dots, 1) \} \subset \{ x, \nabla g_n(x) = 0 \}$
\footnote{
When the coordinates range from $0$ to $n-1$, $g_n(x) = \sum_{i=0}^{n/2-1} \big[ 100 (x_{2i+1} - x_{2i}^2)^2 + (x_{2i} - 1)^2 \big]$ and $\nabla_i g_n(x) = 200(x_{i} - x_{i-1}^2) \cdot \mathbb{1}_{i \in 2 \mathbb{N} + 1} - \big[ 400 x_i (x_{i+1} - x_i^2) - 2(x_i - 1) \big] \cdot \mathbb{1}_{i \in 2 \mathbb{N}}$.}. 
A more involved variant is given by $g_n(x) = \sum_{i=1}^{n-1} \big[ 100 (x_{i+1} - x_i^2)^2 + (x_i - 1)^2 \big]$, with the gradient $\nabla_i g_n(x) = 200 (x_i - x_{i-1}^2) \cdot \mathbb{1}_{i>1} - \big[ 400 x_i (x_{i+1} - x_i^2) - 2(x_i - 1) \big] \cdot \mathbb{1}_{i<n}$, and $x^* = \{1, \dots, 1) \} \subset \{ x, \nabla g_n(x) = 0 \}$
\footnote{
When the coordinates range from $0$ to $n-1$, $g_n(x) = \sum_{i=0}^{n-2} \big[ 100 (x_{i+1} - x_i^2)^2 + (x_i - 1)^2 \big]$ and $\nabla_i g_n(x) = 200 (x_i - x_{i-1}^2) \cdot \mathbb{1}_{i>0} - \big[ 400 x_i (x_{i+1} - x_i^2) - 2(x_i - 1) \big] \cdot \mathbb{1}_{i<n-1}$.}. 
The number of stationary points of this function grows exponentially with dimensionality $n$, most of which are unstable saddle points \citep{10.1162/evco.2009.17.3.437}.

\begin{figure}[h!]
\centering
\includegraphics[width=1.\linewidth]{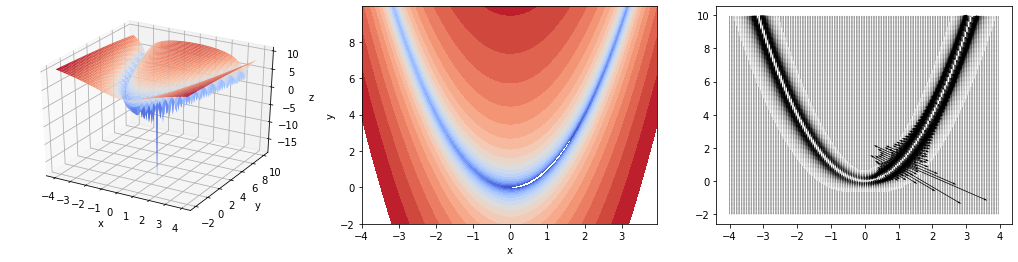}
\caption{Left) Rosenbrock function in log scale for $n=2$, Center) Contours Right) Gradient field: note how these vectors are pronounced in norm near the global minimum; that is important to understand how even near this global optimum many optimizers can fail to reach it.}
\label{fig:rosenbrock}
\end{figure}

\begin{figure}[h!]
\centering
\includegraphics[width=.9\linewidth]{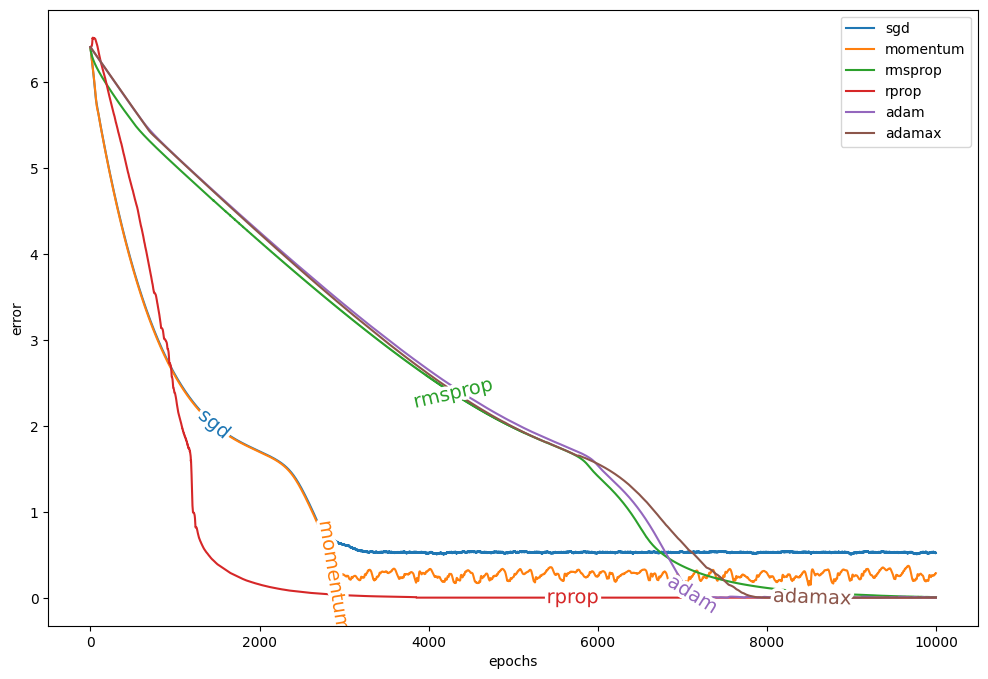}
\caption{Comparative visualization of the progression error of each algorithm on the rosenbrock function}
\label{fig:rosenbrock_loss}
\end{figure}

\begin{figure}[h!]
\centering
\includegraphics[width=.9\linewidth]{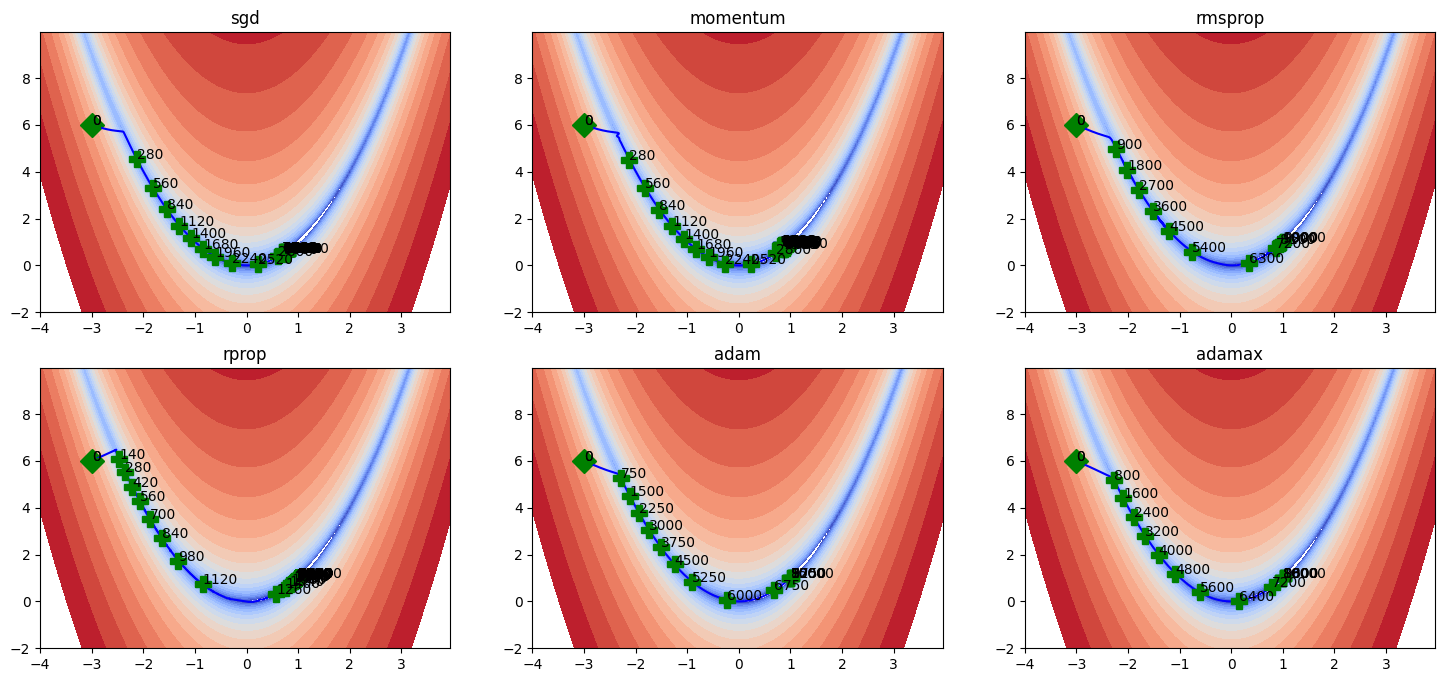}
\caption{Comparative visualization of the progression of each algorithm on the rosenbrock function}
\label{fig:rosenbrock_progression}
\end{figure}

We optimized the Rosenbrock function in a logarithmic scale (to create a ravine, figure \ref{fig:rosenbrock}), for $n=2$. The function is unimodal, and the global minimum is very sharp and surrounded in the direction of the ravine by many local minima. We fall very quickly into the ravine at the beginning of optimization because the surface is well-conditioned. Then, depending on the learning rate and the optimizer used (as well as the associated hyperparameters), we go down the ravine very slowly. Indeed, without momentum, we do not go directly down to the minimum since the gradient is almost zero along the ravine direction but very large in the perpendicular directions: we go from left to right (perpendicular to the ravine), while going down a little, but very slowly. Moreover,  once we are near the minimum, we turn there almost indefinitely. With adaptive gradient, we go down to the minimum very quickly because this direction problem is corrected (due to momentum, left-right ravine perpendicular directions cancel out): if the learning rate is too small, we will also go down very slowly (small gradient in the flat ravine direction). Unlike SGD, here, we always reach the minimum (and stay there). Also, for some learning rates and initializations, there is a double descent \citep{DBLP:conf/iclr/NakkiranKBYBS20} in error (euclidean distance between the global minimum and the current position at a given time) when landing in the ravine.  The methods that succeed in reaching the minimum are rmsprop, rprop, adam,
adamax (figures \ref{fig:rosenbrock_loss} and \ref{fig:rosenbrock_progression}). The method that comes close to it without reaching it is momentum.

\subsection{Rastrigin function}

The rastrigin function is given by $g_n(x) = na + \sum_{i=1}^{n} \big[ x_i^2 - a \cos(2 \pi x_i)  \big] = n a + x^T x -  a 1_n^T \cos(2 \pi x)$ with $a \in \mathbb{R}$. Its gradient is $\nabla g_n(x) = 2x + 2 \pi a \sin (2\pi x)$, and $x^* = 
\{0, \dots, 0) \} \subset \{ x, \nabla g_n(x) = 0 \}$.

\begin{figure}[h!]
\centering
\includegraphics[width=1.\linewidth]{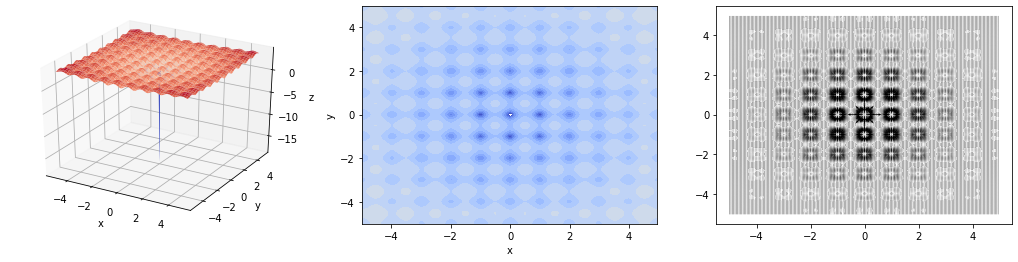}
\caption{Left) Rastrigin function in log scale ($a=10, n=2$), Center) Contours, Right) Gradient field}
\label{fig:rastrigin}
\end{figure}

\begin{figure}[h!]
\centering
\includegraphics[width=.9\linewidth]{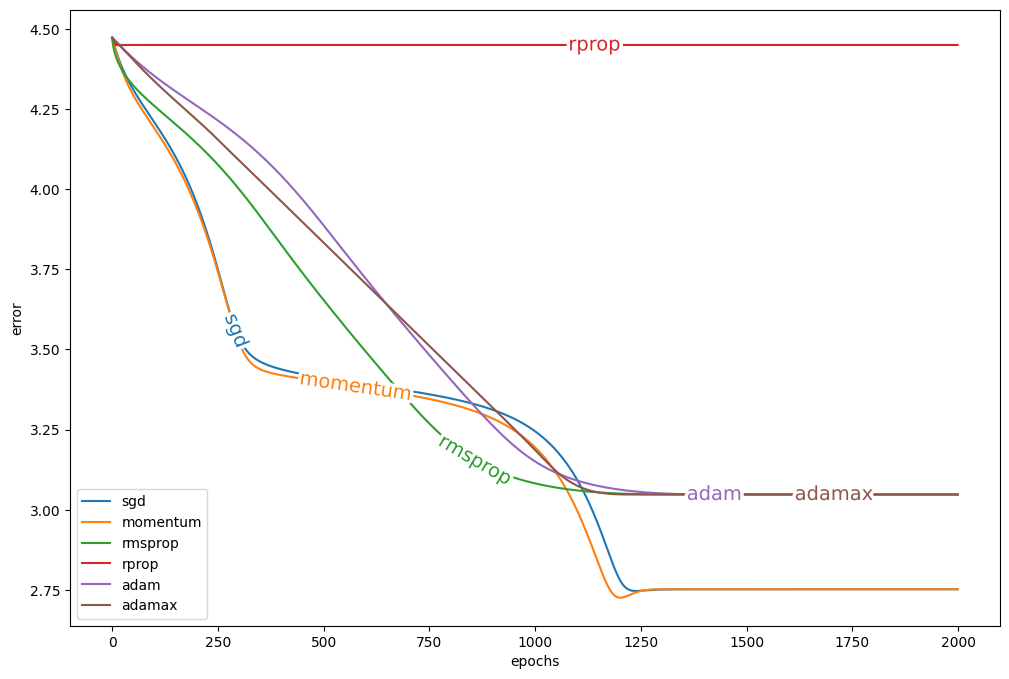}
\caption{Comparative visualization of the progression error of each algorithm on the Rastrigin function}
\label{fig:rastrigin_loss}
\end{figure}

\begin{figure}[h!]
\centering
\includegraphics[width=.9\linewidth]{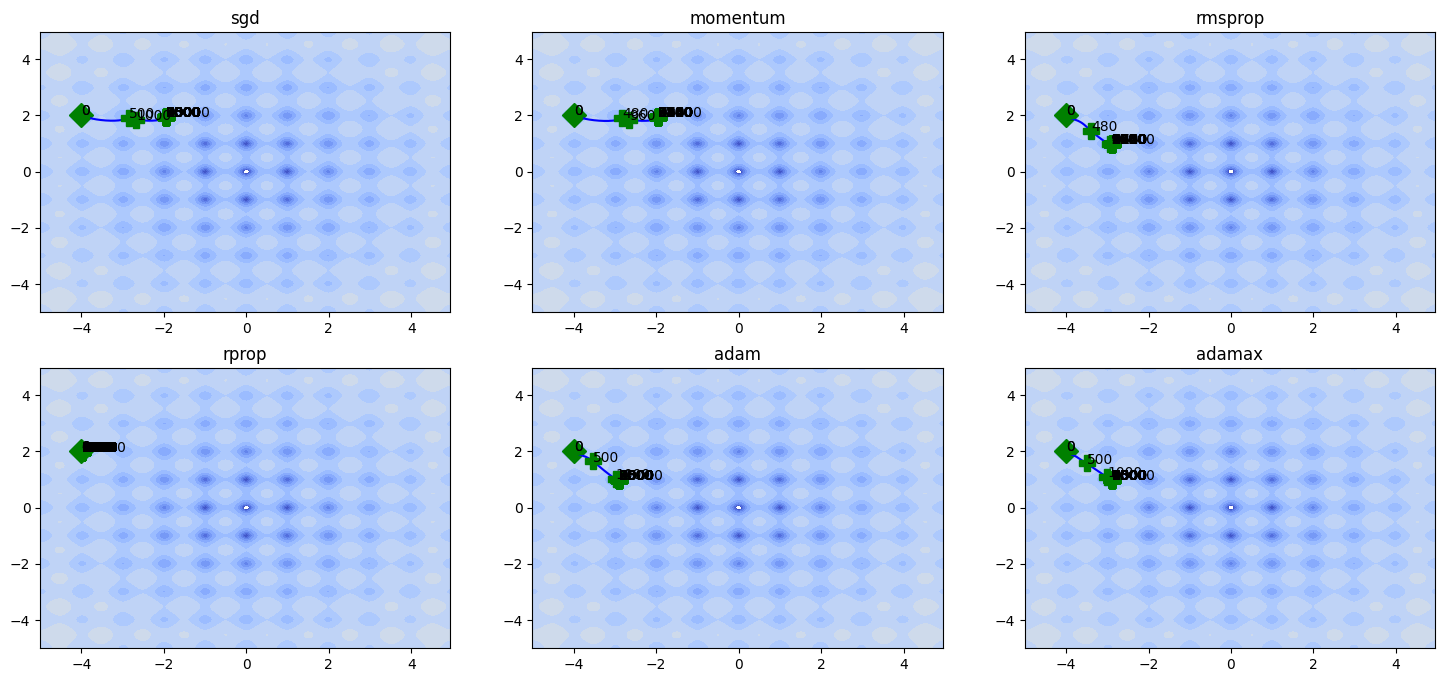}
\caption{Comparative visualization of the progression of each algorithm on the Rastrigin function}
\label{fig:rastrigin_progression}
\end{figure}

We also optimized the Rastrigin function in a logarithmic scale (to create many local minimums and make the global minimum sharp, figure \ref{fig:rastrigin}). The function is unimodal, and the global minimum is sharp and surrounded symmetrically by many local minima. At the beginning of optimization, we fall very quickly into the one local minimum. Then, depending on the learning rate and the optimizer used (and the associated hyperparameters), we can move successively from one minimum to another until we reach the global minimum. No method has succeeded in reaching the global minimum (figures \ref{fig:rastrigin_loss} and \ref{fig:rastrigin_progression}). 

